\documentclass[10pt,twocolumn,letterpaper]{article}

\usepackage[pagenumbers]{cvpr} 

\usepackage{graphicx}
\usepackage{amsmath}
\usepackage{amssymb}
\usepackage{booktabs}

\usepackage{amsfonts}
\usepackage{bbm}
\usepackage{commath}
\usepackage{paralist}
\usepackage{mathtools}
\usepackage{nccmath}
\usepackage{pifont}
\usepackage{tabularx}
\usepackage{makecell} 

\DeclarePairedDelimiter{\nint}\lfloor\rceil
\newcommand{\cmark}{\ding{51}}%
\newcommand{\xmark}{\ding{55}}%

\usepackage[pagebackref,breaklinks,colorlinks]{hyperref}

\usepackage[capitalize]{cleveref}
\crefname{section}{Sec.}{Secs.}
\Crefname{section}{Section}{Sections}
\Crefname{table}{Table}{Tables}
\crefname{table}{Tab.}{Tabs.}

\def\cvprPaperID{***} 
\def\confName{CVPR}
\def\confYear{2022}

\begin{document}

\title{Sequential Voting with Relational Box Fields for Active Object Detection}

\author{
Qichen Fu \qquad Xingyu Liu \qquad Kris M. Kitani\\
\\
Carnegie Mellon University\\
}
\maketitle

\begin{abstract}
\label{abstract}
    A key component of understanding hand-object interactions is the ability to identify the active object -- the object that is being manipulated by the human hand.
    In order to accurately localize the active object, any method must reason using information encoded by each image pixel, such as whether it belongs to the hand, the object, or the background.
    To leverage each pixel as evidence to determine the bounding box of the active object, we propose a pixel-wise voting function.
    Our pixel-wise voting function takes an initial bounding box as input and produces an improved bounding box of the active object as output.
    The voting function is designed so that each pixel inside of the input bounding box votes for an improved bounding box, and the box with the majority vote is selected as the output.
    We call the collection of bounding boxes generated inside of the voting function, the Relational Box Field, as it characterizes a field of bounding boxes defined in relationship to the current bounding box.
    While our voting function is able to improve the bounding box of the active object, one round of voting is typically not enough to accurately localize the active object. 
    Therefore, we repeatedly apply the voting function to sequentially improve the location of the bounding box. However, since it is known that repeatedly applying a one-step predictor (i.e., auto-regressive processing with our voting function) can cause a data distribution shift, we mitigate this issue using reinforcement learning (RL). 
    We adopt standard RL to learn the voting function parameters and show that it provides a meaningful improvement over a standard supervised learning approach.
    We perform experiments on two large-scale datasets: 100DOH and MECCANO, improving AP50 performance by 8\% and 30\%, respectively, over the state of the art. The project page with code and visualizations can be found at \href{https://fuqichen1998.github.io/SequentialVotingDet/}{https://fuqichen1998.github.io/SequentialVotingDet/}.
\end{abstract}

\section{Introduction}
\label{sec:intro}

\begin{figure}[t]
	\centering
	\includegraphics[width=0.49\linewidth]{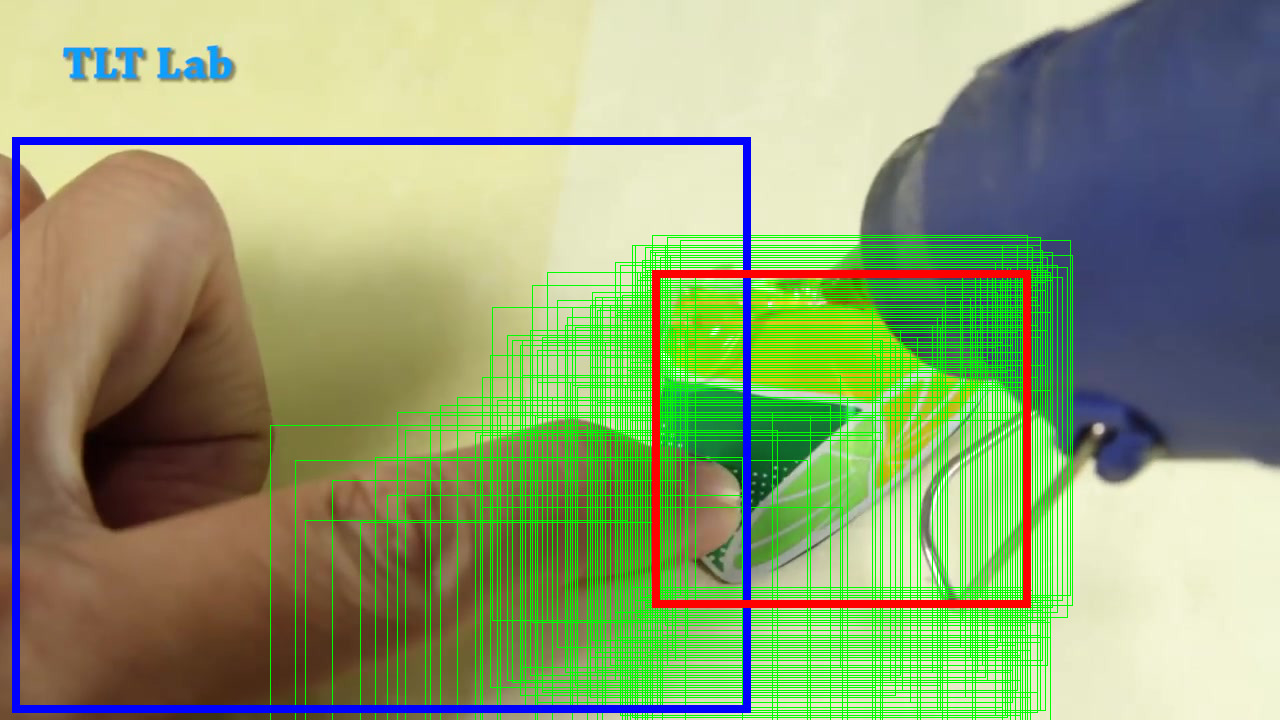}\hfill
	\includegraphics[width=0.49\linewidth]{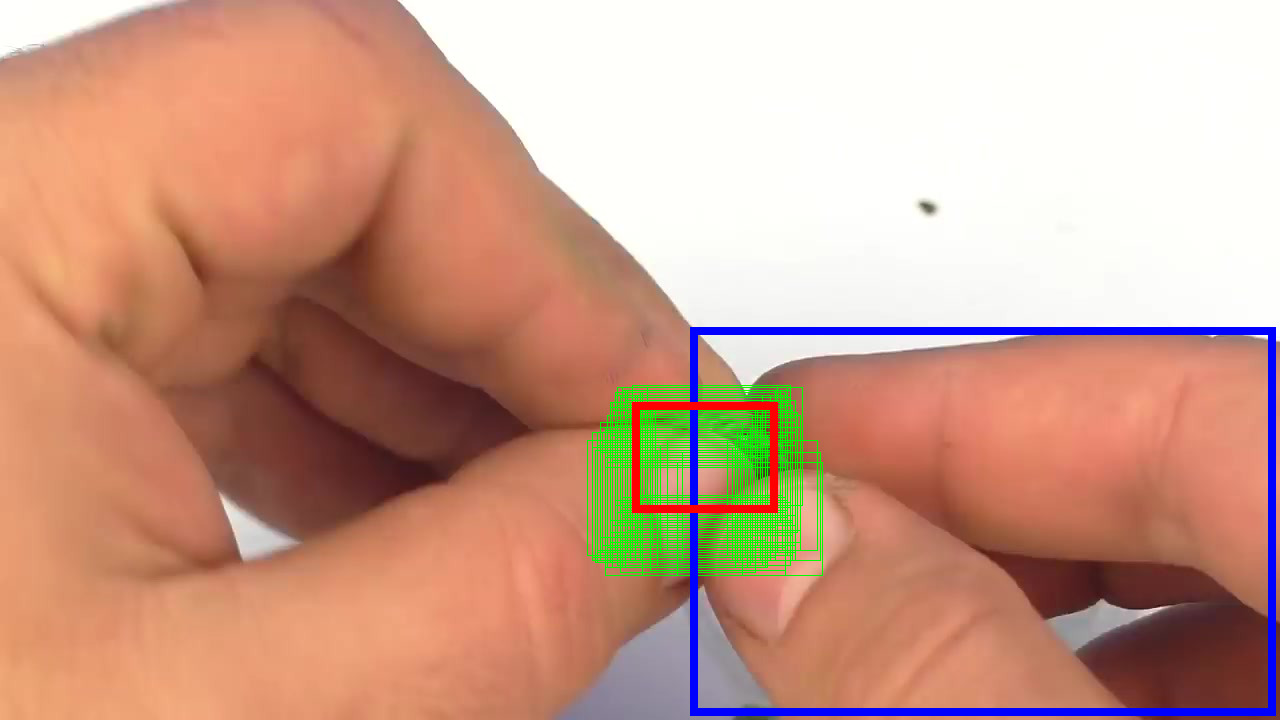}\\
	\includegraphics[width=0.49\linewidth]{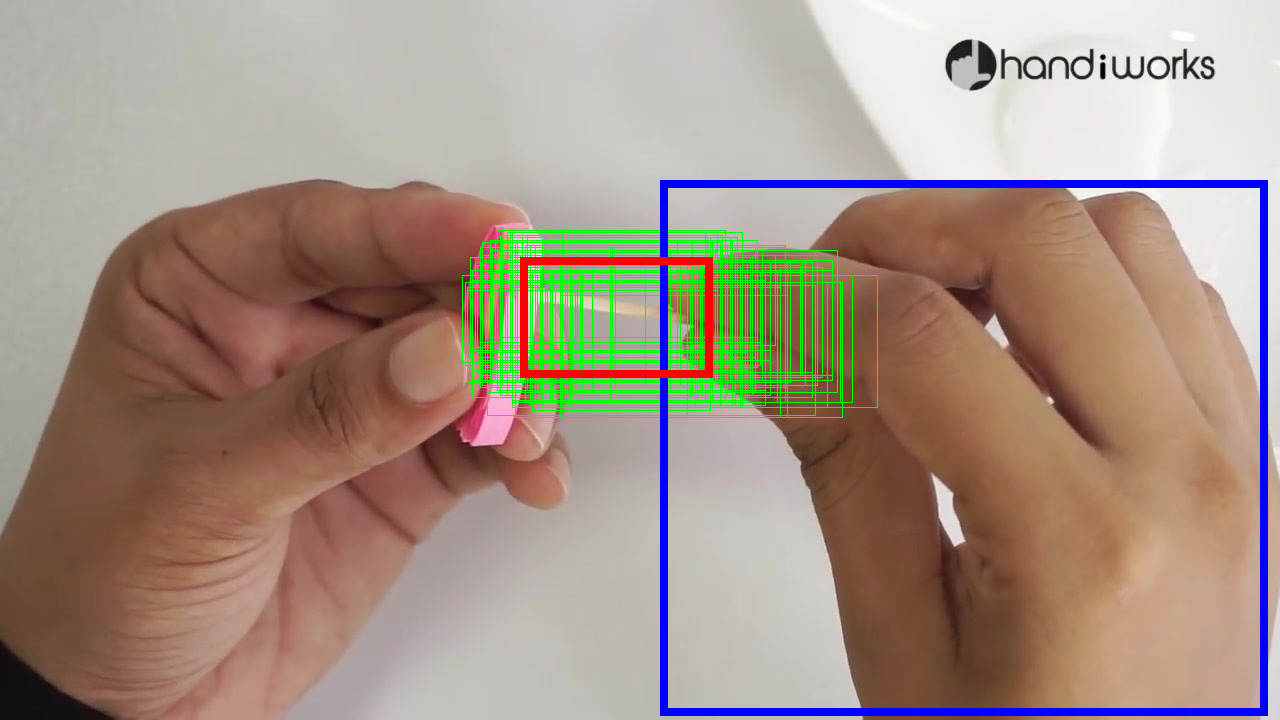}\hfill
	\includegraphics[width=0.49\linewidth]{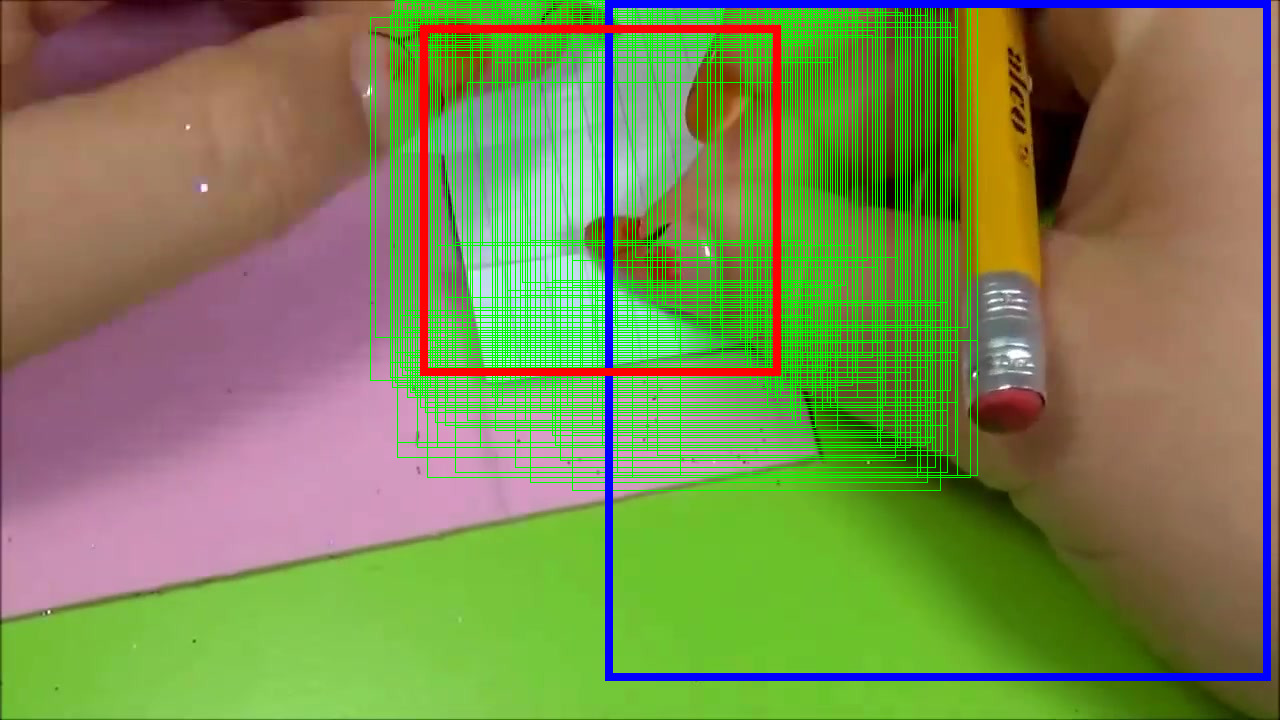}\\
	\includegraphics[width=0.49\linewidth]{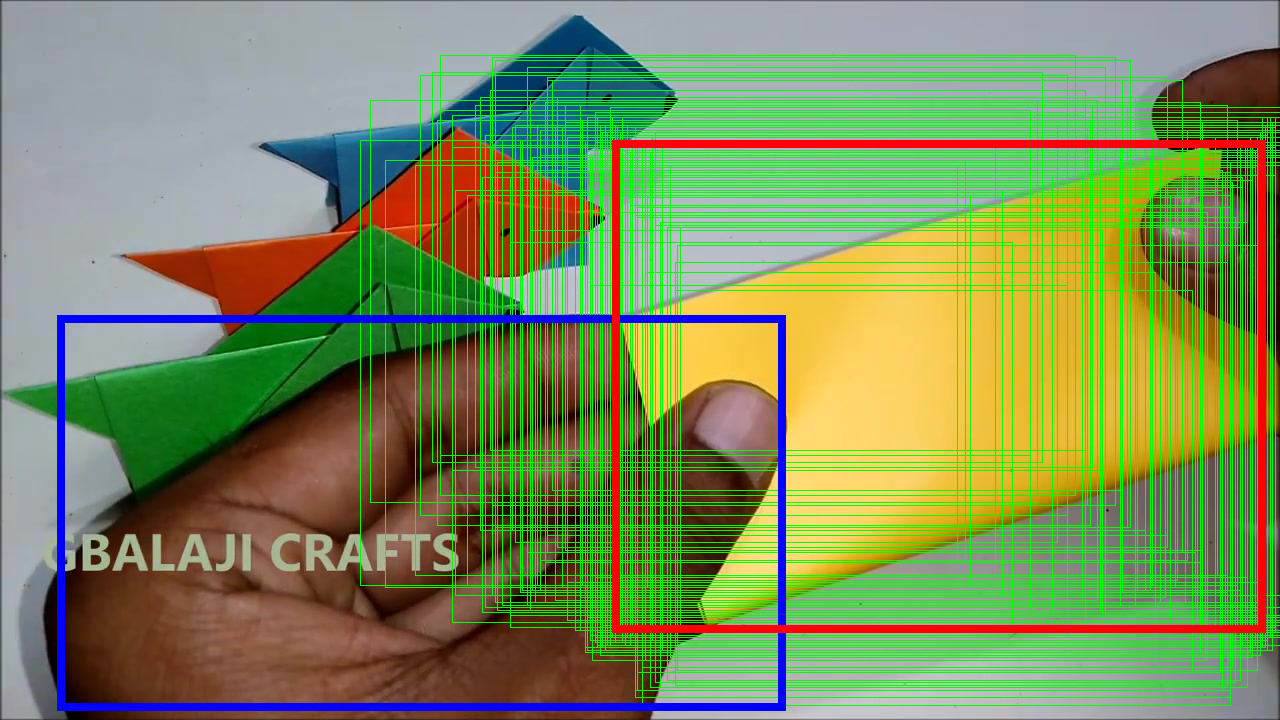}\hfill
	\includegraphics[width=0.49\linewidth]{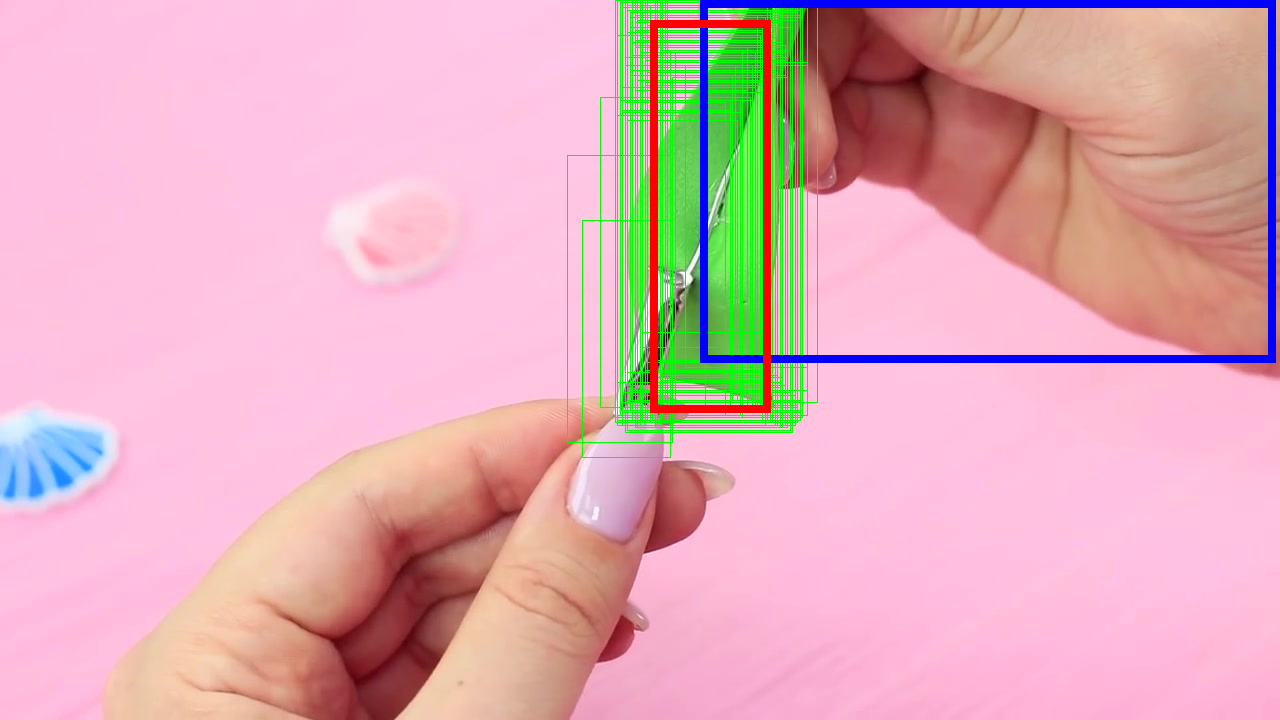}\\
		\centering
	\caption{\label{fig:h2o_votes} Relational Box Field and Pixel-wise Voting visualization. Each green bounding box is an estimated active object bounding box for a pixel inside the blue input bounding box (initialized with a detected hand box). The voting function selects the majority vote prediction (red) as the improved active object bounding box estimate. To ensure visibility, we only show 200 sampled predictions.}
\end{figure}

Finding the active object -- the object that is being manipulated by the human hand -- is a crucial task towards understanding human-object interactions, especially in egocentric videos where hands are the only visible human parts. It is also an essential step to a variety of downstream tasks including joint hand-object pose estimation \cite{hopose:hopenet, hopose:semi, hopose:h+o, hopose:interactionunkown, hopose:fulldof}, reconstruction \cite{horecon:hasson, horecon:feature_fusion}, activity recognition \cite{ego:understandactivity, ego:attention, ego:deeper}, and imitation learning \cite{imilrn}.  However, an accurate localization of the active object can be challenging due to natural occlusions caused by the hands during interactions. During a hand-object interaction, it is common for the hand to occlude most of the object in order to grasp the object. On one hand, this makes it hard to detect active objects. On the other hand, the appearance of the hand actually contains important information about the location, shape, size, and pose of the active object. It is important to develop computer vision algorithms that can leverage each pixel of the image, especially the hands and objects, to accurately estimate a bounding box around the active object.

To advance the state of art in active object detection, we introduce a pixel-wise voting function to improve the active object bounding box estimate while being robust to occlusion. The voting function takes as input an initial bounding box estimate of the active object (typically seeded by bounding box around the hand region), and then predicts an improved bounding box, where the improved bounding box is tighter and more centered around the active object. Inside the voting function, we predict a large set of improved active object bounding boxes, by allowing each pixel inside the input box to regress a new bounding box. We call the collection of active object bounding boxes as the \emph{Relational Box Field} (RBF), as they represent a field of bounding boxes related to pixels inside the input bounding box. As we can observe in \cref{fig:h2o_votes}, the pixel-wise predictions can be quite diverse because they depend on features from different locations in the input image. Our method overcomes inconsistencies across the RBF by using a technique similar to \cite{pvnet, posecnn, dehv, hoi:dirv}, where our voting function finds the consensus from pixel-wise predictions by selecting the bounding box with a majority vote. 
Similar to the Hough transform, our voting scheme is able to minimize the influence of outliers and generate stable predictions through the power of aggregation. We also show later in our experiments that our voting scheme is more robust when compared to standard regression methods.

While the voting function can provide a better active object bounding box estimation, one round of voting is typically not enough to accurately localize the active object. Similar to the idea of boosting \cite{boosting} where one uses a sequence of computational units to iteratively improve prediction performance, we apply multiple rounds of the voting function to progressively obtain a more accurate active object bounding box. However, repeatedly applying a function trained for one-step prediction (\textit{i.e.,} supervised learning or behavior cloning) can result in a data distribution shift, also known as covariate shift \cite{covariate_shift}. For example, each time a one-step predictor is used in an auto-regressive manner (\textit{i.e.}, the output is passed as input for the next sequence), it can introduce small errors which can compound over the prediction sequence, leading to a data distribution shift, which can lead to bad performance towards the end of a sequence. In order to mitigate this issue, we use reinforcement learning (RL) to learn the proposed voting function. RL is designed to account for distribution shifts in auto-regressive processes by evaluating and optimizing over sequences. Specifically, we use a Markov Decision Process (MDP) to model the sequential decision-making process of obtaining an optimal active object bounding. In each step of the sequential decision-making process, the voting function (the MDP policy) takes an initial estimate of the active object bounding box as input (the MDP state) and predicts a new improved active object bounding box (the MDP action and next state). Interestingly, we find that the first estimate of the bounding box of the active object can be seeded using a hand bounding box since active objects are usually near the hands. We demonstrate that using reinforcement learning provides a meaningful improvement over standard supervised learning models for active object detection.

Our approach is evaluated on two large-scale hand-object interaction datasets: 100DOH \cite{100doh} and MECCANO \cite{meccano} datasets. Experiments show that our method achieves new state-of-the-art performance on both hand-object detection and active object detection tasks. We also demonstrate the better generalization ability of our model by evaluating its performance across the datasets. Last, we provide a comprehensive ablation study for our design choices.

\begin{figure}[t]
	\centering
	\includegraphics[width=\linewidth]{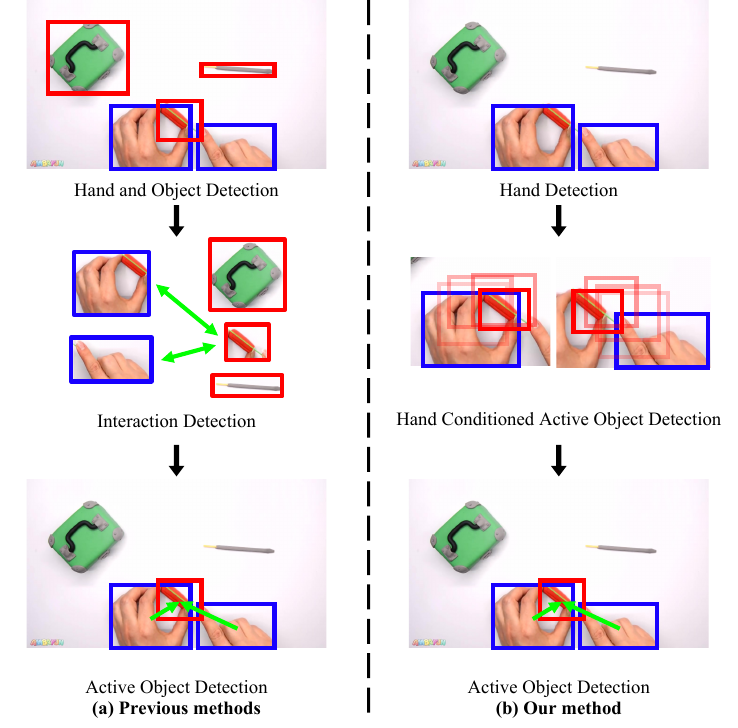}
	\caption{\label{fig:prev_proposed_compare} Previous methods detect active object by first detecting hands and objects \emph{independently}, followed by an interaction detection to match active objects and hands. Our method directly detects the active object corresponding to each hand a sequential decision-making process dependent on hand.}
\end{figure}

\begin{figure*}[t]
\centering
\includegraphics[width=0.99\linewidth]{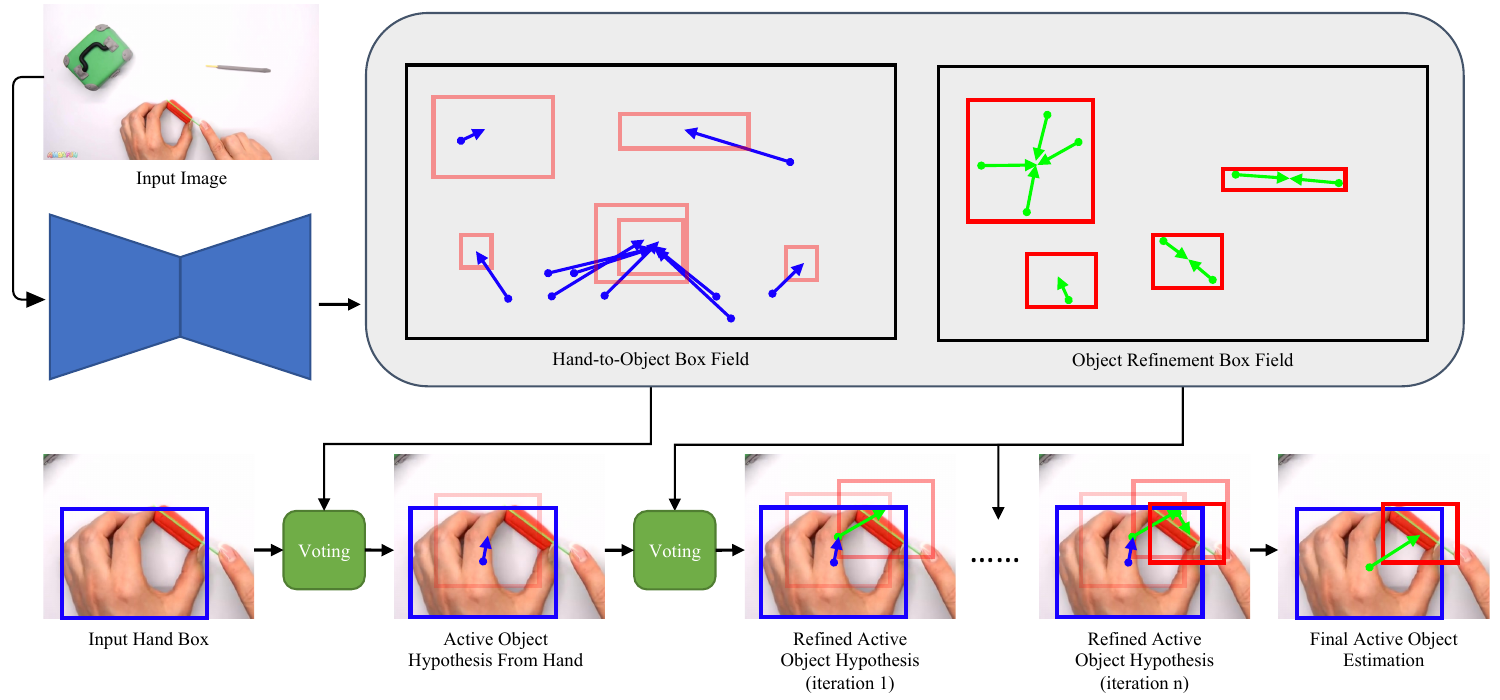}
\caption{\label{fig:overview} Overview of the sequential decision-making process for active object detection. Our approach, seeded using a hand bounding box, progressively refines the current active object bounding box towards an optimal active object estimation. In each step, we use a voting function on the Relational Box Fields to predict a better active object bounding box based on the current input.}
\end{figure*} 

\section{Related Work}
\label{sec:related_work}

\paragraph{Human-object Interaction Detection} 

As shown in \cref{fig:prev_proposed_compare}, existing approaches \cite{ppdm, hoi:dirv, hoi:kaiming, 100doh} in human(or hand)-object interaction usually detect active objects in two steps: (1) detecting hand and object \emph{independently} with a classical detector \cite{fasterrcnn, centernet_zhou, centernet_duan, detr}, followed by (2) an hand-object interaction detection to match hands and objects detected in the first step. One limitation of using classical object detectors is that they are not designed to detect objects under occlusions. Meanwhile, in these methods, the important relationship between hand and object is also not considered when localizing the active object. 
Instead, our method achieves an accurate localization of the active object in a sequential decision-process dependent on hand, which is robust to occlusions and fully exploits the feature of hand, object, and their inter-dependency like human perception \cite{eye_hand, eye_visual_search}.

\paragraph{Object Detection with Reinforcement Learning} 
Different from classic object detection methods \cite{fasterrcnn, centernet_zhou, centernet_duan}, RL-based detection approaches \cite{dlr_activeobjectdetection, rl_tree, rl_detection} progressively narrow down the scope from the initial guess to the object bounding box in a top-down sequential process based on the current observation and historical paths. 
Some RL methods \cite{dlr_activeobjectdetection, rl_tree, rl_detection} localize the visual objects in scenes using top-down sequential policies. Some methods \cite{rl_cascade, rl_efficient_detection} combine RL with classical detectors to improve object detection performance or efficiency. For instance, \cite{rl_efficient_detection} uses RL to efficiently generate bounding box proposals to replace the first stage of two-stage detectors like \cite{fasterrcnn}. 
Similar to \cite{dlr_activeobjectdetection} training a localization agent that starts with the whole scene and narrows down to the object, we design an agent which starts from the hand box and progressively moves to the object touched by the hand. The key difference is our agent does not directly predict a single action that deforms the bounding box using translation and scaling. Instead, we propose a voting function that allows each pixel to have a different prediction and then finds the consensus from pixel-wise predictions, which essentially enables the agent to ignore noisy observation caused by occlusions.

\paragraph{Dense Methods} Most dense methods make a prediction for each pixel or patch in the image, then use voting to aggregate the results into a robust estimation. \cite{pvnet, posecnn} use pixel-wise predictions with Hough voting to localize keypoint for pose estimation. \cite{dehv} uses patch-based voting to remove false-positive hypotheses from the background for joint object detection and depth recovery. \cite{hoi:dirv} uses bounding-box-wise voting for human-object interaction detection. Different from the above methods, our proposed method progressively performs pixel-wise voting on the \emph{Relational Box Field} to predict and improve the active object estimation from hand.

\section{Method}
\label{sec:method}

The overview structure of our framework is illustrated in \cref{fig:overview}. To achieve robust object localization, especially under occlusion, we propose a voting function with the Relational Box Field that allows each pixel in the image to vote for a bounding box of the active object. To further refine the active bounding box estimation, we repeatedly apply the voting function until the bounding box converges.

In the following subsections, we first explain the Relational Box Field with Pixel-wise Voting for improving the active object bounding box estimation. Then we describe how to use an MDP to model the sequential decision-making process that progressively obtains a more accurate active object bounding box. Finally, we describe the implementation details of the model and the hybrid training strategy consisting of imitation learning and reinforcement learning to learn an optimal policy efficiently.

\subsection{Relational Box Field with Pixel-wise Voting}
\label{ssec:rbl_voting}

We describe a voting function for estimating an improved bounding box of an object from the input box. Specially, we first predict a \emph{Relational Box Field} which encodes bounding box predictions for every pixel inside the image. Then the predictions from the pixels inside the input box are aggregated into a single improved object box through pixel-wise voting.

\paragraph{Relational Box Field} The \emph{Relational Box Field} $F$ is a field of active object bounding boxes in relationship to each pixel inside the image. Specifically, the estimated Relational Box Field $\hat{F}\in \mathbb{R}^{H\times W\times 5}$ has the same spatial resolution as the input image $I\in \mathbb{R}^{H\times W\times 3}$. For a location $(u, v) \in \mathbb{R}^2$, $\hat{F}_{u, v} \in \mathbb{R}^5$ represents its related object box in the form of
\begin{align}
    \hat{F}_{u, v} = [\hat{r}_{u,v},\hat{\theta}_{u,v}, \hat{h}_{u,v}, \hat{w}_{u,v}, \hat{c}_{u,v}]
\end{align}
where $\hat{r}_{u,v} \in \mathbb{R}^+$ and $\hat{\theta}_{u,v} \in [0, 2\pi)$ represent the predicted relative displacement from $(u, v)$ to its related bounding box center in polar coordinate system. $\hat{h}_{u,v} \in (0, H]$ and $\hat{w}_{u,v} \in (0, W]$ are the height and width of the bounding box prediction at $(u, v)$. $\hat{c}_{u,v} \in [0, 1]$ is the confidence score of the prediction at $(u, v)$.

\paragraph{Loss for Relational Box Field}
During training, we supervise the Relational Box Field with respect to the set of ground truth bounding boxes $\mathcal{A}$.
Note that bounding boxes may overlap with each other.
To alleviate the confusion due to overlapping, we remove the overlapped regions from each bounding box $a\in \mathcal{A}$ during training as clipped bounding boxes $a^\star$:

\begin{equation}
    a^\star = a \cap (\coprod \mathcal{A})  \\
\end{equation}
where $\coprod$ denotes the disjoint union, \emph{i.e.,} the union of non-overlapped areas in $\mathcal{A}$.

For each pixel $(u, v)$ inside the clipped bounding box $a^\star$, we apply localization loss $L^{loc}_{u, v}$ -- smooth-$L_1$ regression loss on the predicted box parameters $(\hat{r}_{u ,v}, \hat{\theta}_{u, v}, \hat{h}_{u, v}, \hat{w}_{u, v})$, and Focal Loss $L^{C}_{u, v}$ on the predicted confidence score $\hat{c}_{u, v}$. 
Both $L^{loc}_{u, v}$ and $L^{C}_{u, v}$ are only applied in $a^\star$.
Given predicted $\hat{F}$, the overall loss applied to a bounding box $a$ is the average of pixel-wise losses:

\begin{equation}
    L_{\hat{F}}(a) =
    \frac{\sum_{u, v \in a^\star} L^{loc}_{u, v} + L^{C}_{u, v}}{|a^\star|}
\end{equation}

To reduce the bias towards large bounding boxes,
we compute the total loss of Relational Box Field by averaging the loss of each bounding box
\begin{equation}
\label{eq:box_field_loss}
    L_{\hat{F}}(\mathcal{A}) = \frac{\sum_{a \in \mathcal{A}} L_{\hat{F}}(a)}{|\mathcal{A}|}
\end{equation}

\paragraph{Weighted Pixel-wise Voting}
To improve the robustness against noise in the Relational Box Field prediction, we use pixel-wise voting to aggregate the predicted bounding box results.
The voting function summarizes the bounding box predictions from all pixels belonging to the input box $a$ 
into an improved bounding box estimation $(\hat{x}^a, \hat{y}^a, \hat{h}^a, \hat{w}^a)$ as
\begin{equation}
\label{eq:vote}
    \hat{x}^a, \hat{y}^a, \hat{h}^a, \hat{w}^a = \text{Vote}_{\hat{F}}(a)
\end{equation}

To identify the best location of the improved bounding box, we create a histogram of votes over bounding box parameters, where the predictions with larger confidence scores $\hat{c}_{u, v}$ contributes with larger weights
Specifically, we compute the voting scores of the target object box center $S^a_{\text{center}} \in \mathbb{R}^{H\times W}$, width $S^a_{\text{width}} \in \mathbb{R}^{W}$, and height $S^a_{\text{height}} \in \mathbb{R}^{H}$ for the improved bounding box estimation from the input box $a$ as
\begin{multline}
\label{eq:distribution}
S_{\text{center}}^a(x, y) = \sum_{u, v\in a} \hat{c}_{u, v} \cdot \mathbbm{1}(\nint{u+\hat{r}_{u, v}\sin(\hat{\theta}_{u, v})} = y) \cdot\\ \mathbbm{1}(\nint{v+\hat{r}_{u, v}\cos(\hat{\theta}_{u, v})} = x)\\
S_{\text{width}}^a(w)= \sum_{u, v\in a} \hat{c}_{u, v} \cdot \mathbbm{1}(\nint{\hat{w}_{u, v}} = w)\\
S_{\text{height}}^a(h) = \sum_{u, v\in a} \hat{c}_{u, v} \cdot \mathbbm{1}(\nint{\hat{h}_{u, v}} = h)\\
\end{multline}
where $\mathbbm{1} (\cdot)$ is the indicator function and $\nint{\cdot}$ is the function of rounding to the nearest integer.
The optimal bounding box parameters $(\hat{x}_a, \hat{y}_a, \hat{h}_a, \hat{w}_a)$ are then obtained by retrieving the candidate with maximum voting score in $S^a_{\text{center}}, S^a_{\text{width}}, S^a_{\text{height}}$ respectively.

\begin{figure*}[t]
	\centering
	\includegraphics[width=0.19\linewidth]{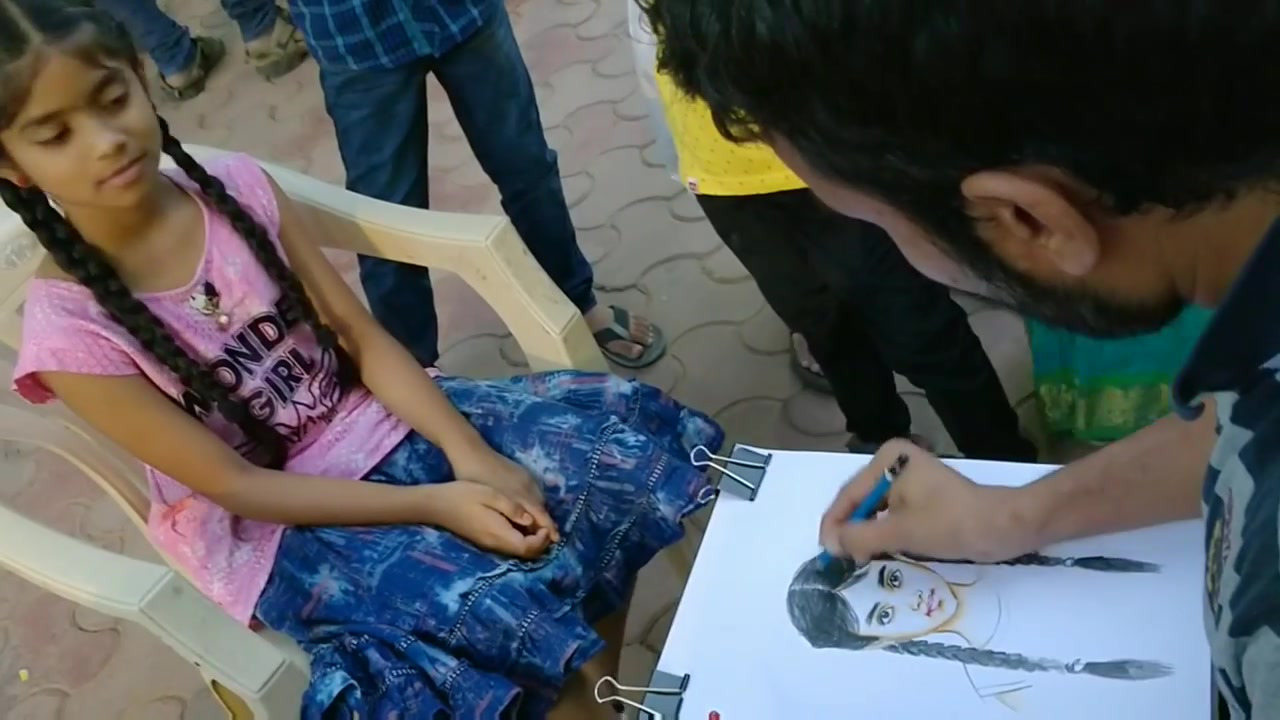}
	\includegraphics[width=0.19\linewidth]{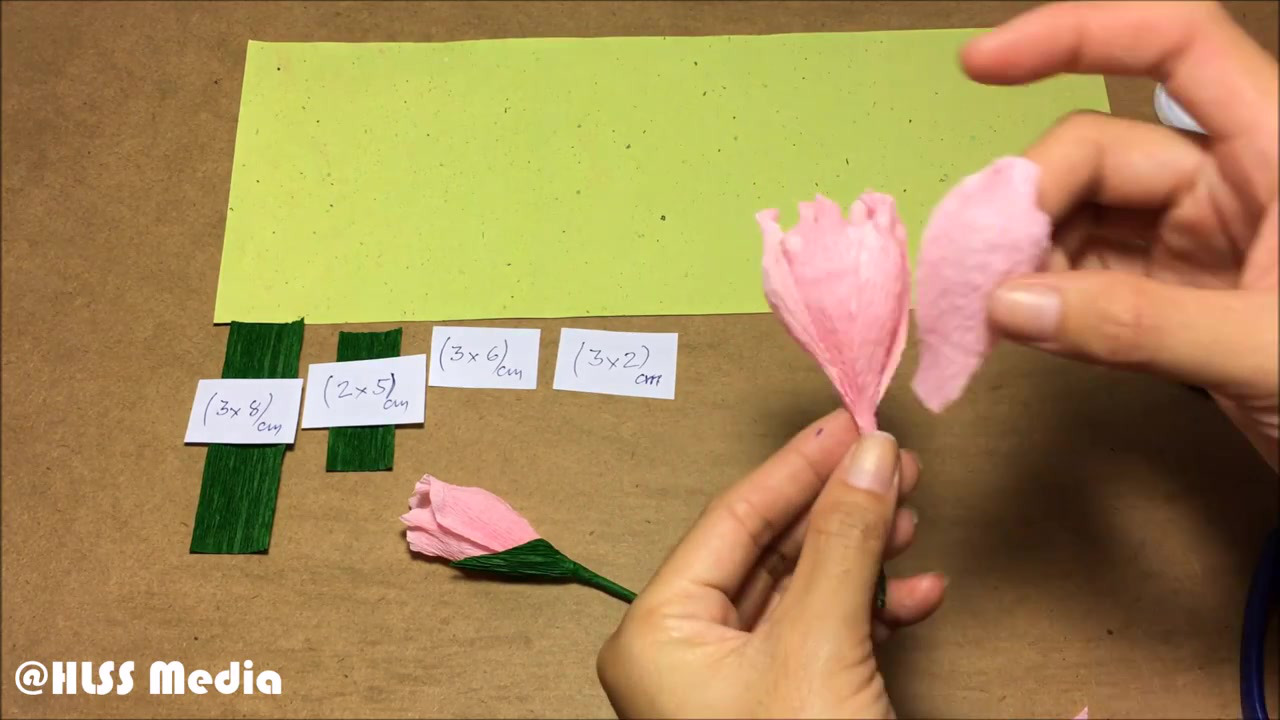}
	\includegraphics[width=0.19\linewidth]{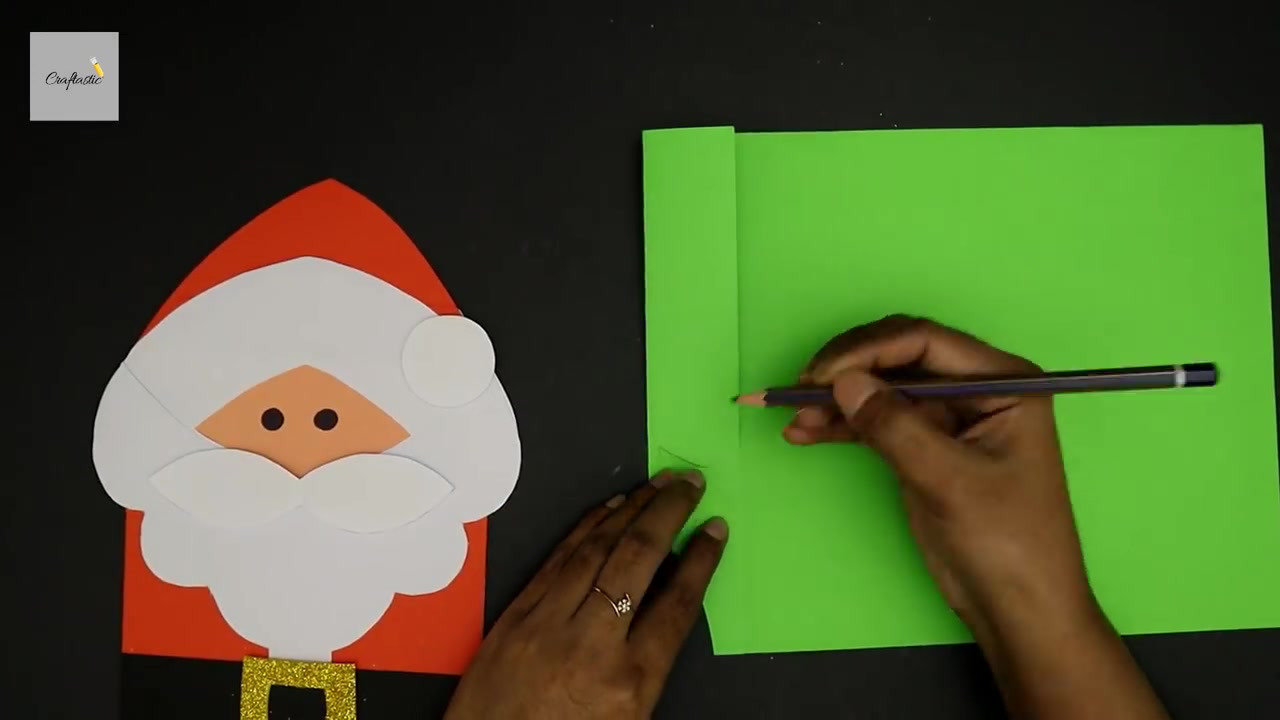}
	\includegraphics[width=0.19\linewidth]{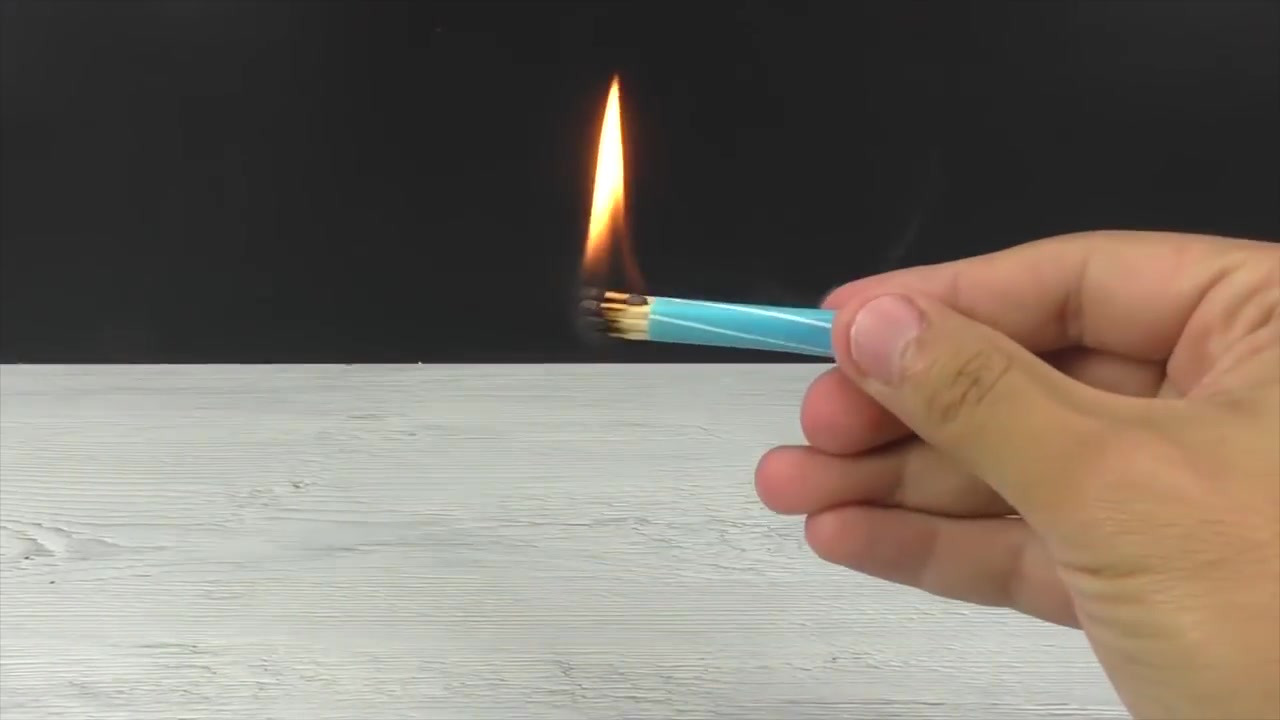}
	\includegraphics[width=0.19\linewidth]{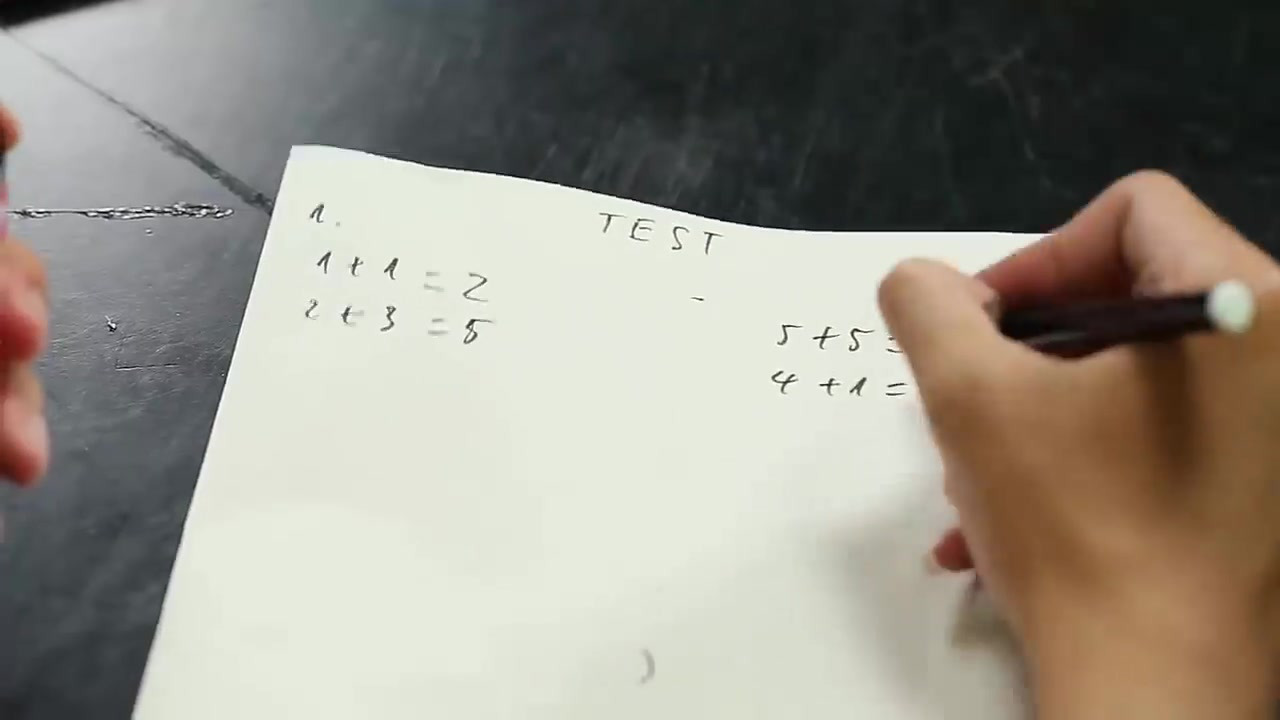}
	\\
	\includegraphics[width=0.19\linewidth]{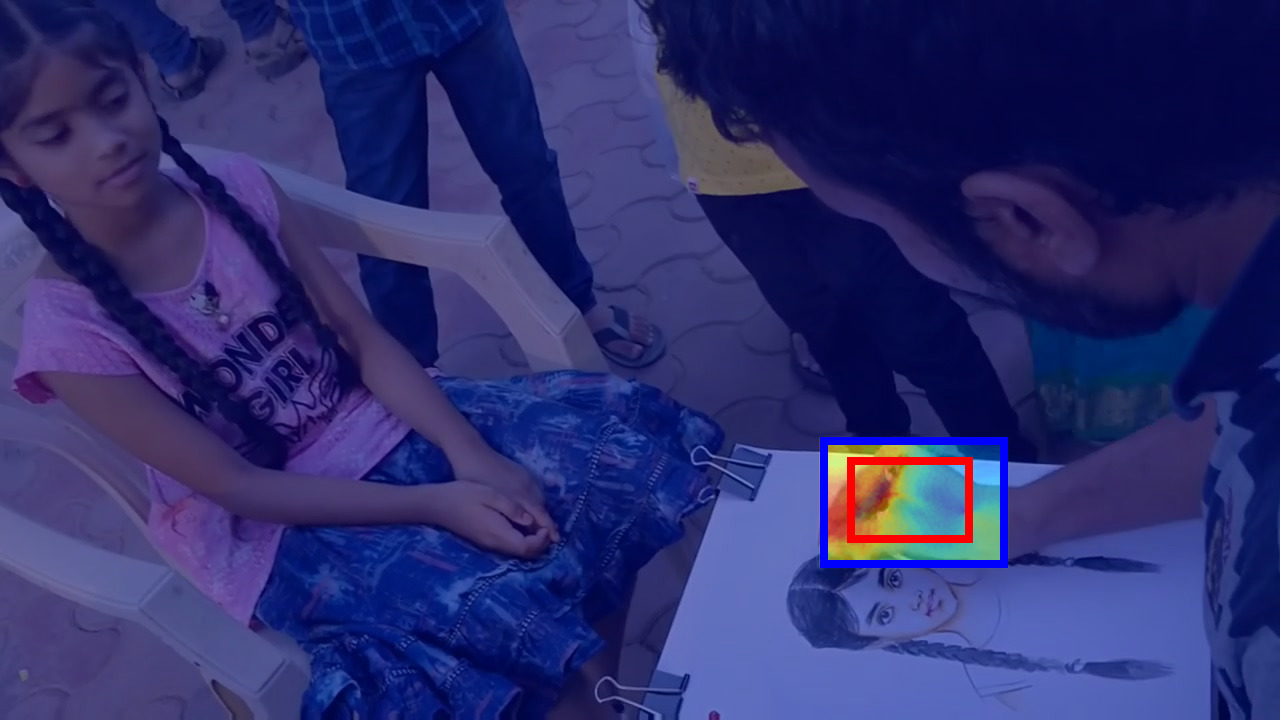}
	\includegraphics[width=0.19\linewidth]{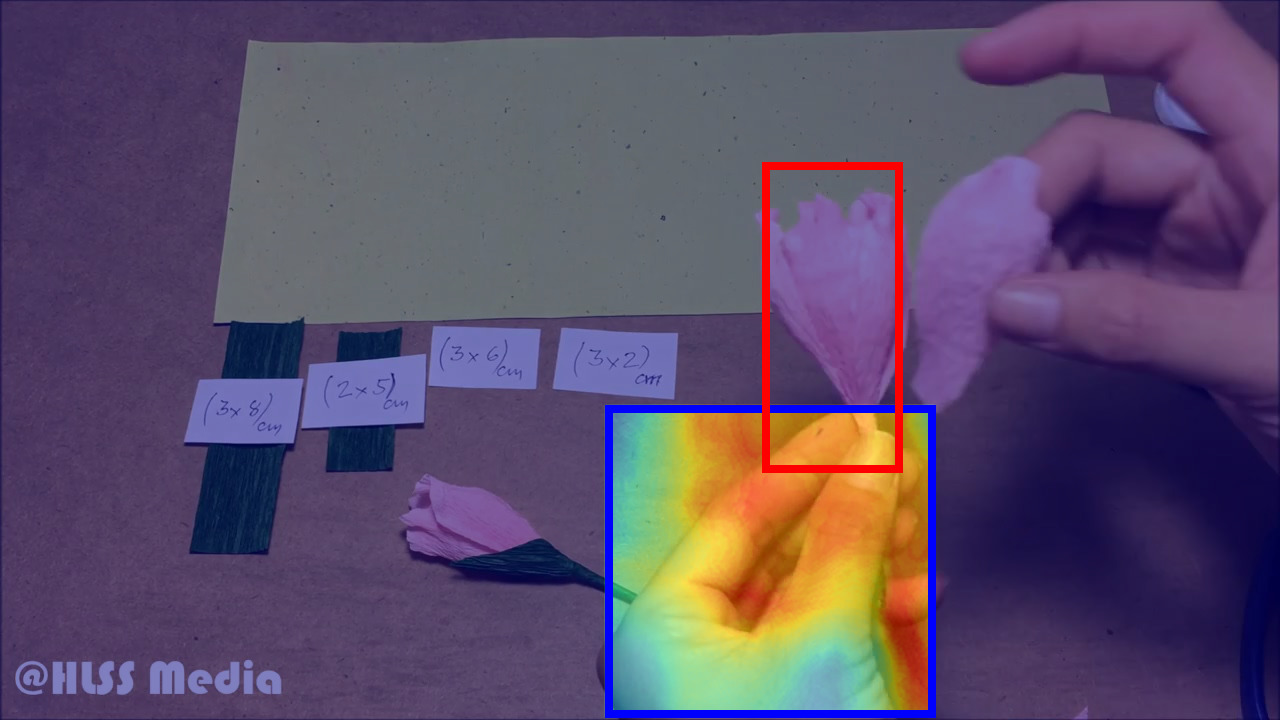}
	\includegraphics[width=0.19\linewidth]{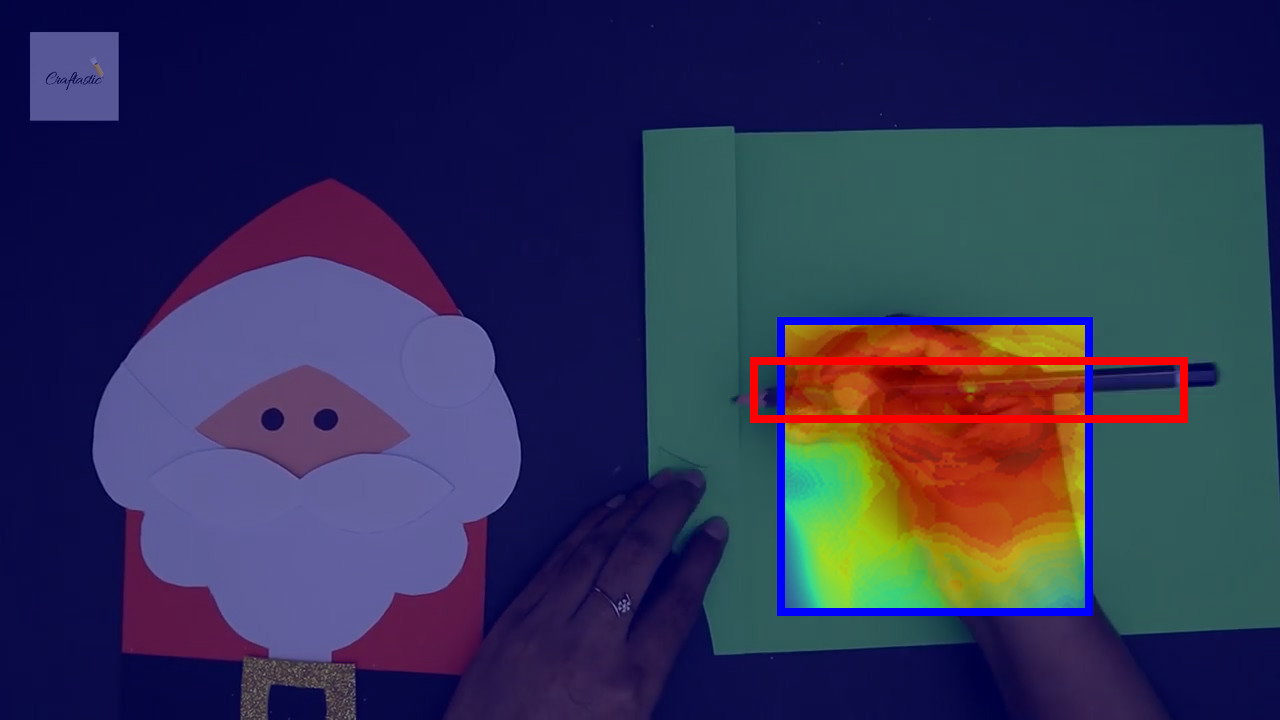}
	\includegraphics[width=0.19\linewidth]{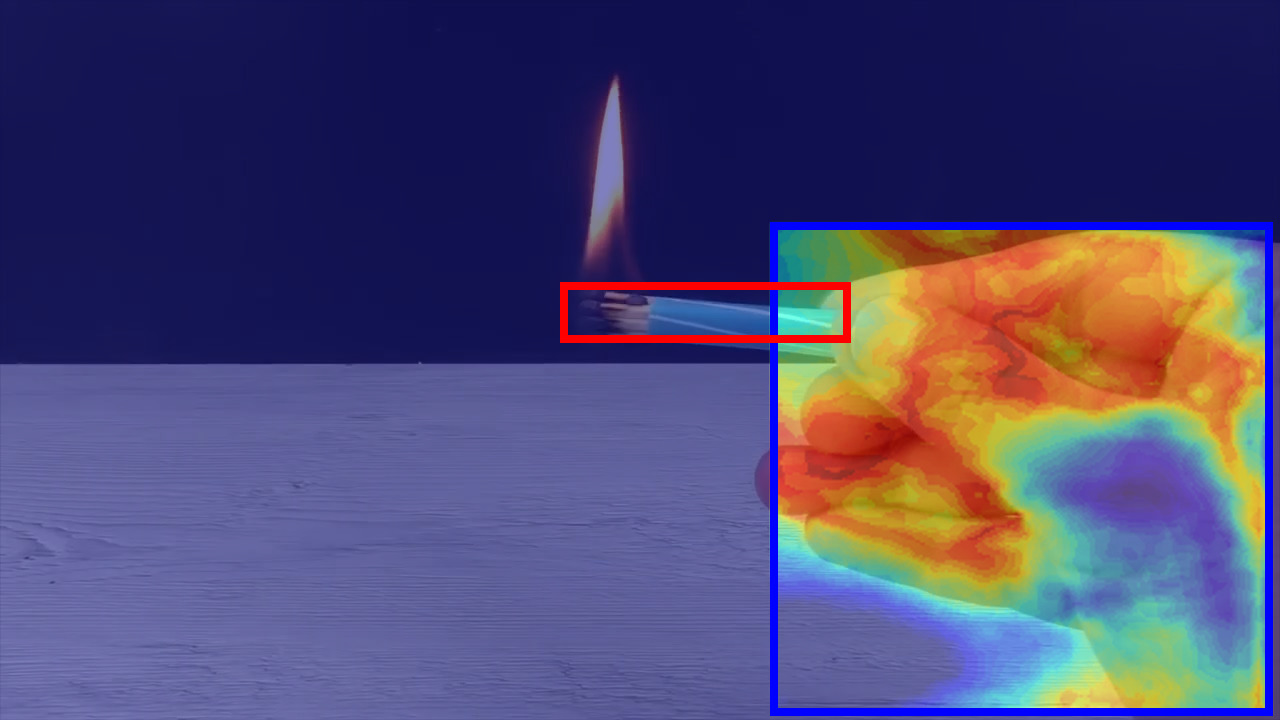}
	\includegraphics[width=0.19\linewidth]{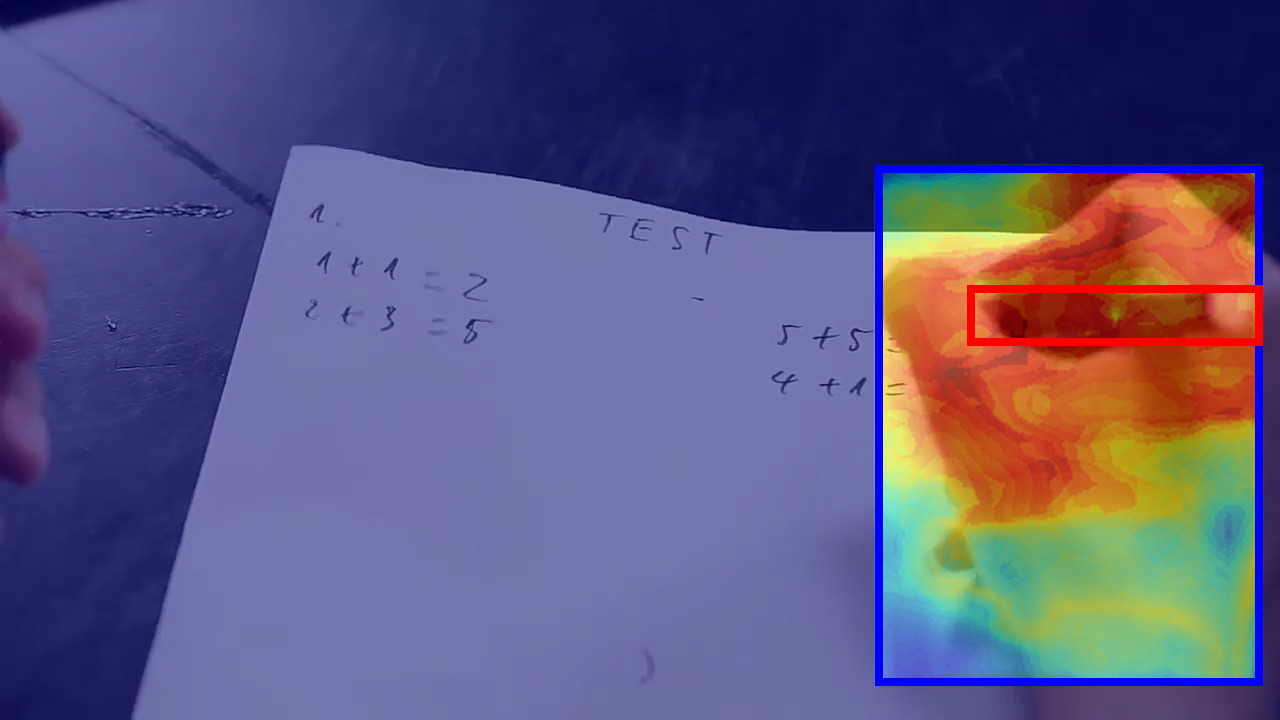}
	\\
	\centering
	\caption{\label{fig:vote_correctness} We visualize the IoU (red indicates higher IoU) between the final active object box estimation (red) and the pixel-wise predictions inside the hand bounding box (blue). This figure shows voting is able to adapt predictions from informative hand parts like fingers as opposed to irrelevant parts like wrist and background.}
\end{figure*}

As shown in \cref{fig:vote_correctness}, the informative patterns (such as fingers and objects) produce more consistent predictions compared to irrelevant information such as background and occlusions. Since the voting function picks the box with the most votes as output, it allows the model to explicitly focus on informative patterns instead of irrelevant information. 

\subsection{MDP}
\label{ssec:mdp}

Recall the goal of our method is to localize the active object bounding box $b^o = [x^o, y^o, w^o, h^o]\in \mathbb{R}^4$ corresponding to a hand bounding box $b^h = [x^h, y^h, w^h, h^h]\in \mathbb{R}^4$ in an image $I$, where the $(x^o, y^o)$ and $(x^h, y^h)$ are the centers of the active object and hand box, and $w^o, h^o, w^h, h^h$ stand for their widths and heights. Considering the remarkable success in hand detection \cite{hand_det_cplxego, hand_det_mp, 100doh}, we assume the hand boxes is given. As the grasp appearance of human hands is highly indicative of the location and shape of the object being manipulated, we leverage this important clue by first generating an active object bounding box hypothesis using visual clues inside the hand bounding box. However, the object hypothesis based on only hand appearance could be ambiguous because a similar hand grasp could hold various objects. Thus, it is important to further refine the object estimation by incorporating object patterns from the estimated object bounding box as direct evidence. 

We find only one step of pixel-wise voting from the Relational Box Field could be insufficient to localize the active object accurately. Therefore, we propose to further refine the active object localization based on the updated estimation of the bounding box in a sequential decision-making process, which is modeled by an MDP.
The state, action, dynamics, policy, and reward of the MDP are defined as follows.

\paragraph{State} The state $s_t$ at each timestamp $t$ consists of two parts: a local state and a global state feature which is consistent throughout the sequence. We use the current active object bounding box estimation $\hat{b}^o_t = [\hat{x}_t^o, \hat{y}_t^o, \hat{w}_t^o, \hat{h}_t^o]\in \mathbb{R}^4$ as the local state, where $(\hat{x}_t^o, \hat{y}_t^o)$ represents the center of estimated bounding box and $\hat{w}_t^o, \hat{h}_t^o$ stand for its width and height. The global state representation is a combination of the detected hand bounding box $b^h$ and the image feature $\mathcal{F}\in \mathbb{R}^{H\times W\times 256}$ extracted from image $I$.
\begin{equation}
\label{eq:state_reper}
    s_t = (\hat{b}^o_t, b^h, \mathcal{F})
\end{equation}

Our method does not require an initial active object bounding box guess. Instead, we exploit the visual clue inside the hand bounding box to generate the initial active object bounding box hypothesis. 
In other words, the local state $\hat{b}^o_0$ at timestamp $t=0$ can be arbitrary initialized because it is ignored by the policy.

\paragraph{Action} Each action at timestamp $t$ is in the format of $a_t = [\Delta x_t, \Delta y_t, \Delta w_t, \Delta h_t]\in \mathbb{R}^4$, which are the offsets to update the center coordinate, width and height of the active bounding box estimation.

Actions that do not significantly change the bounding box lead to the \emph{terminal} state. We use the relative changes of the center, height, and width of the bounding box as the criterion for termination. Specifically, we check:
\begin{equation}
 \frac{\abs{\Delta x_t}}{\hat{w}_t^o}, \frac{\abs{\Delta y_t}}{\hat{h}_t^o}, \frac{\abs{\Delta w_t}}{\hat{w}_t^o}, \frac{\abs{\Delta h_t}}{\hat{h}_t^o}
\end{equation}
The \emph{terminal} state will be triggered if all the relative changes are below a threshold. We set the threshold value to $0.05$ in all of our experiments.

\paragraph{Dynamics} Our state transition dynamics $h: (s_t, a_t) \mapsto s_{t+1}$ is a deterministic function, which leads to exactly one next state from one state-action pair. It updates the current state $s_t$ by adding the action $a_t$ (the offsets for bounding box parameters) to the current active object bounding box estimation $\hat{b}^o_t$. Except when timestamp $t=0$, the action is applied to the hand bounding box $b^h$ since our method is conditioned on the hand location and appearance for the initial active object hypothesis.

\paragraph{Policy} 

The policy generates an action $a_t$ by applying the voting function on two relational box fields: hand-to-object box field $\hat{F}^{ho}$ and object refinement box field $\hat{F}^{oo}$, which are predicted based on the image feature $\mathcal{F}$ in the current state $s^t$.

The hand-to-object box field $F^{ho}$ addresses the \textit{contact} relation, which is used to generate an initial hypothesis of the active object box $\hat{b}^o_1$. Specifically, $F^{ho}$ encodes the mapping from the hand pixels to the corresponding active object bounding box

\begin{equation}
    F^{ho}: (u, v) \mapsto (x^o, y^o, w^o, h^o) \qquad \text{for } u, v \in b^h
\end{equation}

In $F^{ho}$, the ground truth confidence score of contact relation $c^{ho}_{u, v}=1$ if $(u,v)$ lies in a hand which is touching an object, while $c^{ho}_{u, v}=0$ otherwise.

The object refinement relational box field $F^{oo}$ exploits object local patterns inside the current object box estimation to further improve it. 
Specifically, $F^{oo}$ encodes the mapping from the pixels of active objects to their box center, width, and height.
\begin{equation}
    F^{oo}: (u, v) \mapsto (x^o, y^o, w^o, h^o) \qquad \text{for } u, v\in b^o
\end{equation}

To summarize, at $t=0$, the policy generates an initial active object hypothesis by applying the voting in \cref{eq:vote} to bounding box predictions within $\hat{F}^{ho}$ from pixels belonging to the hand bounding box $h^t$. When $t > 0$, the policy applies the voting in \cref{eq:vote} to predictions within $\hat{F}^{oo}$ from pixels belonging to the current active object estimation $\hat{b}^o_t$ to vote for a refined active object bounding box. Specifically, the policy finds a local optimal active object bounding box estimation $\hat{b}_{t+1}^o$ as:
\begin{equation}
    \hat{b}_{t+1}^o = \begin{cases}
     \text{Vote}_{\hat{F}^{ho}}(b^h) & \text{if } t = 0\\
     \text{Vote}_{\hat{F}^{oo}}(\hat{b}_t^o)& \text{otherwise}\\
    \end{cases}
\end{equation}
where $\hat{b}^o_t$ is the local state and $b^h$ belongs to the global state representation in $s_t$ defined in \cref{eq:state_reper}. 

The output action of the policy $a_t = \pi(s_t)$ is defined as the displacement towards the local optimal active object bounding box estimation $\hat{b}^o_{t+1}$ from either current estimation $\hat{b}^o_t$ when $t > 0$, or the hand bounding box $b^h$ when $t = 0$.

\paragraph{Reward} We design the reward $r$ as the accuracy of the active object bounding box estimation. As the action $a_t$ updates the active object bounding box from $\hat{b}^o_{t}$ to $\hat{b}^o_{t+1}$ , we use the GIoU \cite{GIoU} between $b^o_{t+1}$ and the ground truth $b_o$ as the reward at timestamp $t$
\begin{equation}
    r(s_t, a_t) = \text{GIoU}(\hat{b}^o_{t+1}, b^o)
\end{equation}

\subsection{Implementation Details}
\label{ssec:method:implemetation}

In this subsection, we describe the implementation details of our image feature extractor and the policy model.

\paragraph{Image Feature Extractor} Our image feature extractor $f$ is in an encoder-decoder style with self-attention. The network takes an input image $I\in\mathbb{R}^{H\times W\times 3}$ and outputs a feature map $\mathcal{F}\in \mathbb{R}^{H\times W\times 256}$. For the encoder, we use a pre-trained ResNet101 on ImageNet and modify it by changing the last two layers into dilated convolution to increase the spatial size of the feature map. To further exploit the synergy between hand and object, we apply one layer of image-wide self-attention (with details in the supplementary) on the deep feature before forwarding it to the decoder. In the decoder, the Atrous Spatial Pyramid Pooling from DeepLabV3+ \cite{dlv3+} and bi-linear up-sampling are repeatedly performed to efficiently expand the feature map until its size matches the input size.

\paragraph{Policy Model} Our light-weighed policy model is a $1\times 1$ convolution layer with an input channel of size $256$ and an output channel of size $14$. The policy model takes the image feature $\mathcal{F}$ and outputs the prediction of hand-to-object relational box field $F^{ho}$ and object refinement relational box field $F^{oo}$.

\begin{figure*}[t]
	\centering
	\rotatebox{90}{Ground Truth}\hfill
	\includegraphics[width=0.19\linewidth]{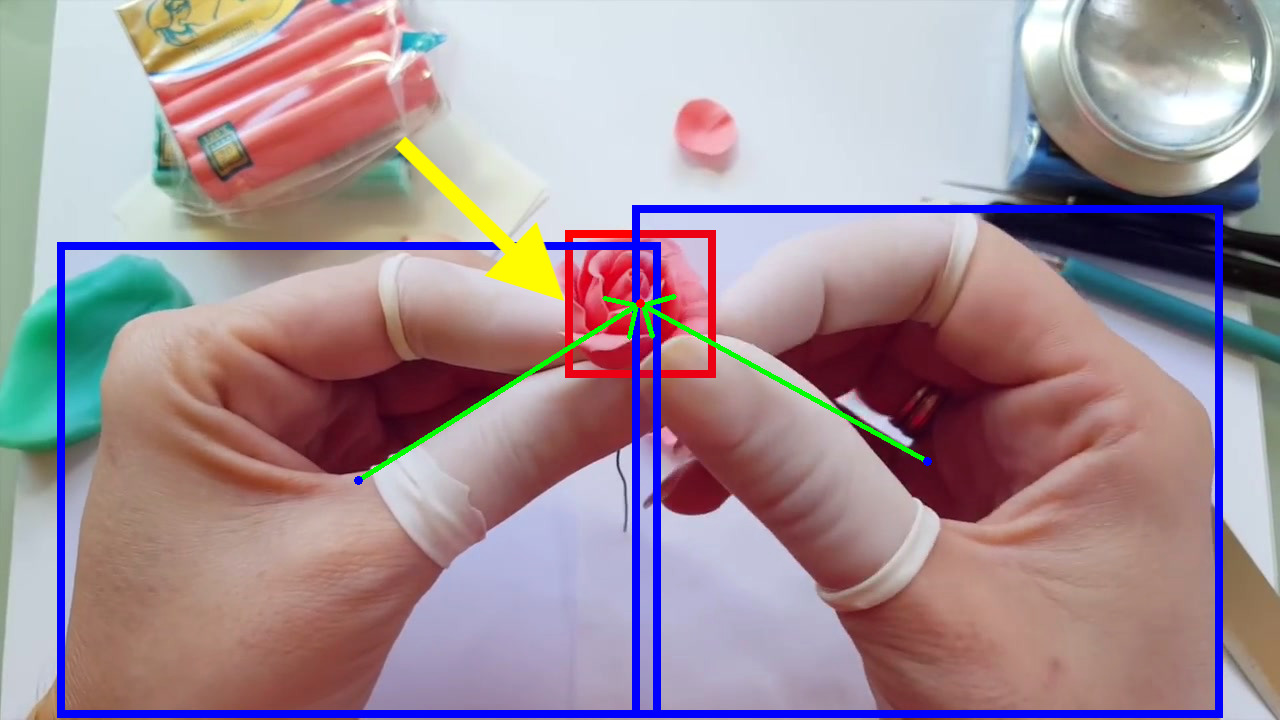}
	\includegraphics[width=0.19\linewidth]{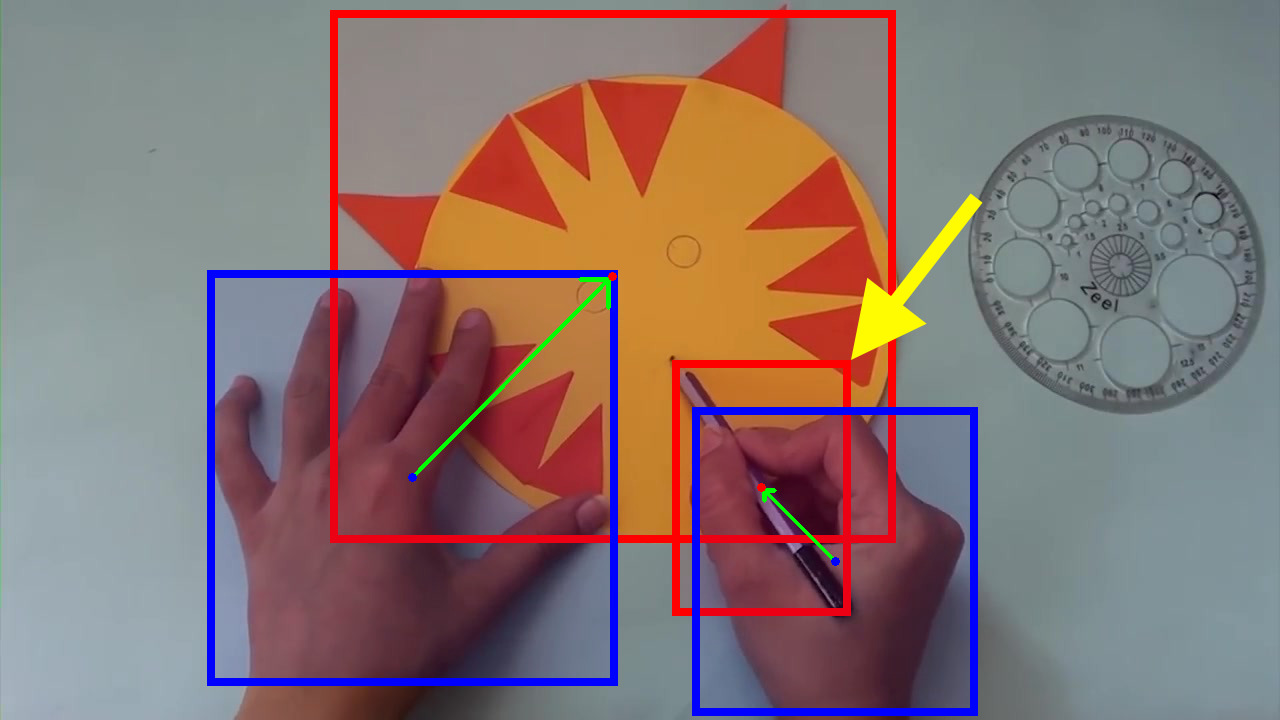}
	\includegraphics[width=0.19\linewidth]{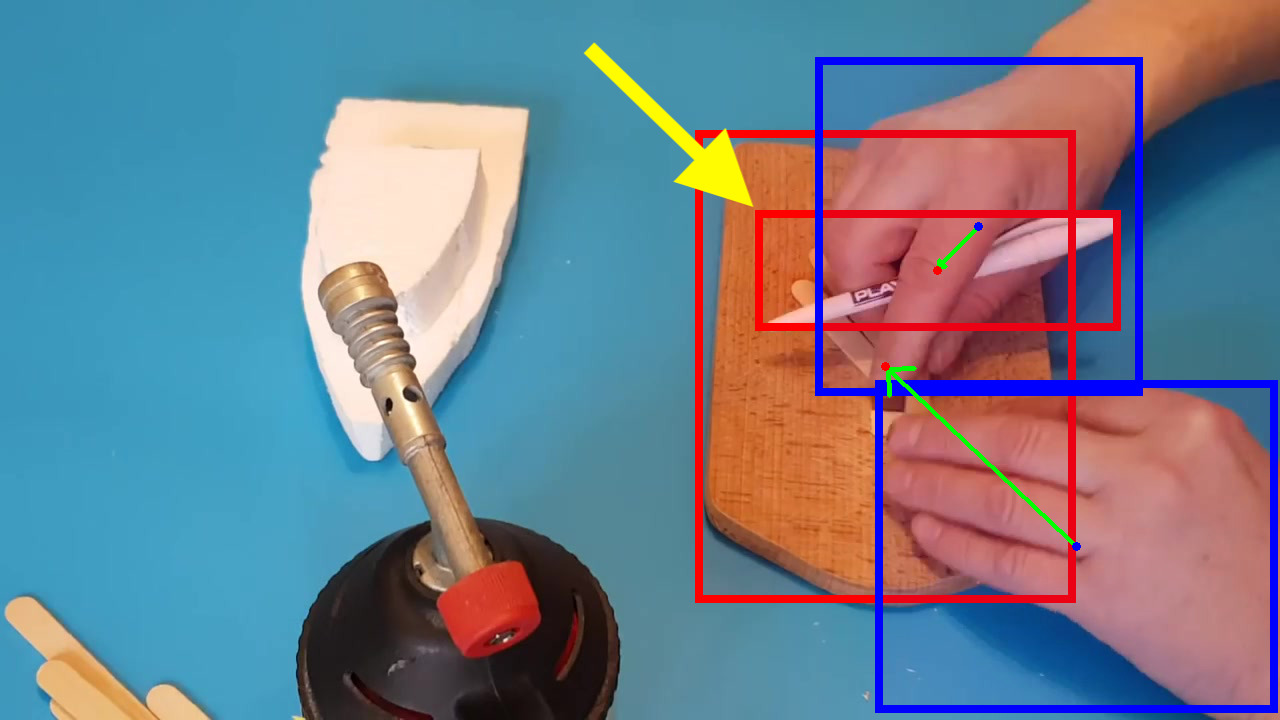}
	\includegraphics[width=0.19\linewidth]{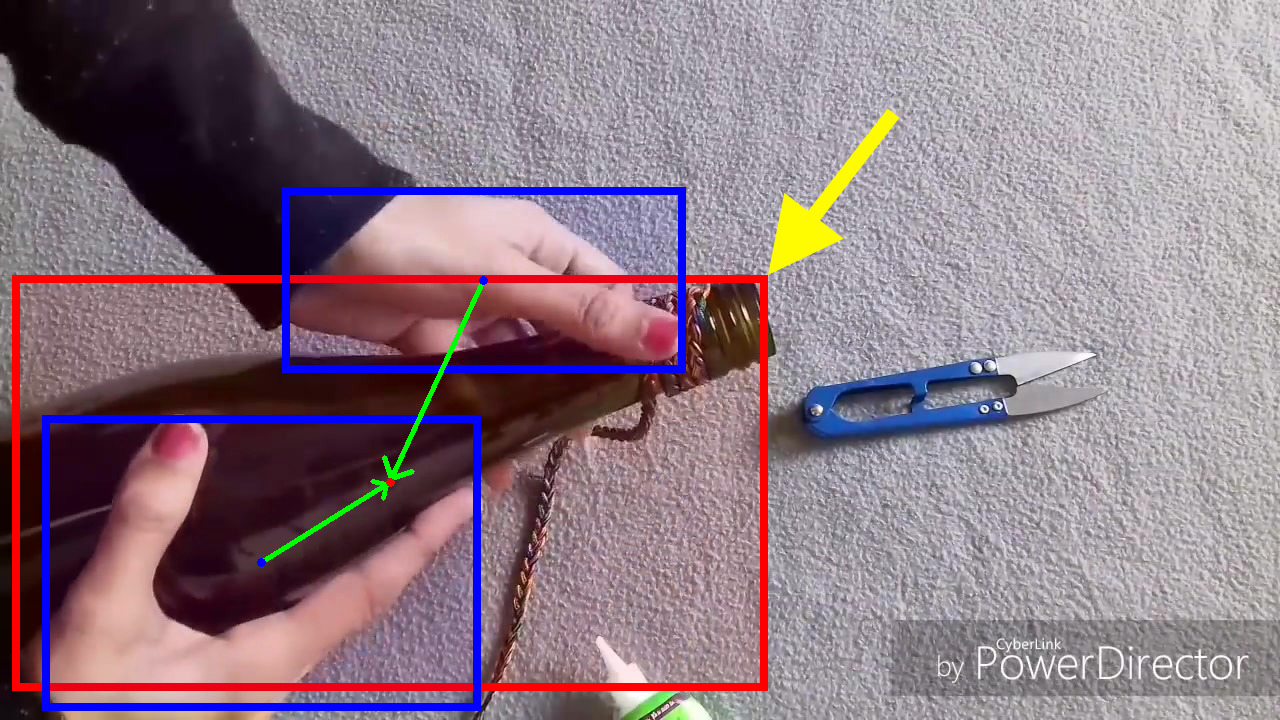}
	\includegraphics[width=0.19\linewidth]{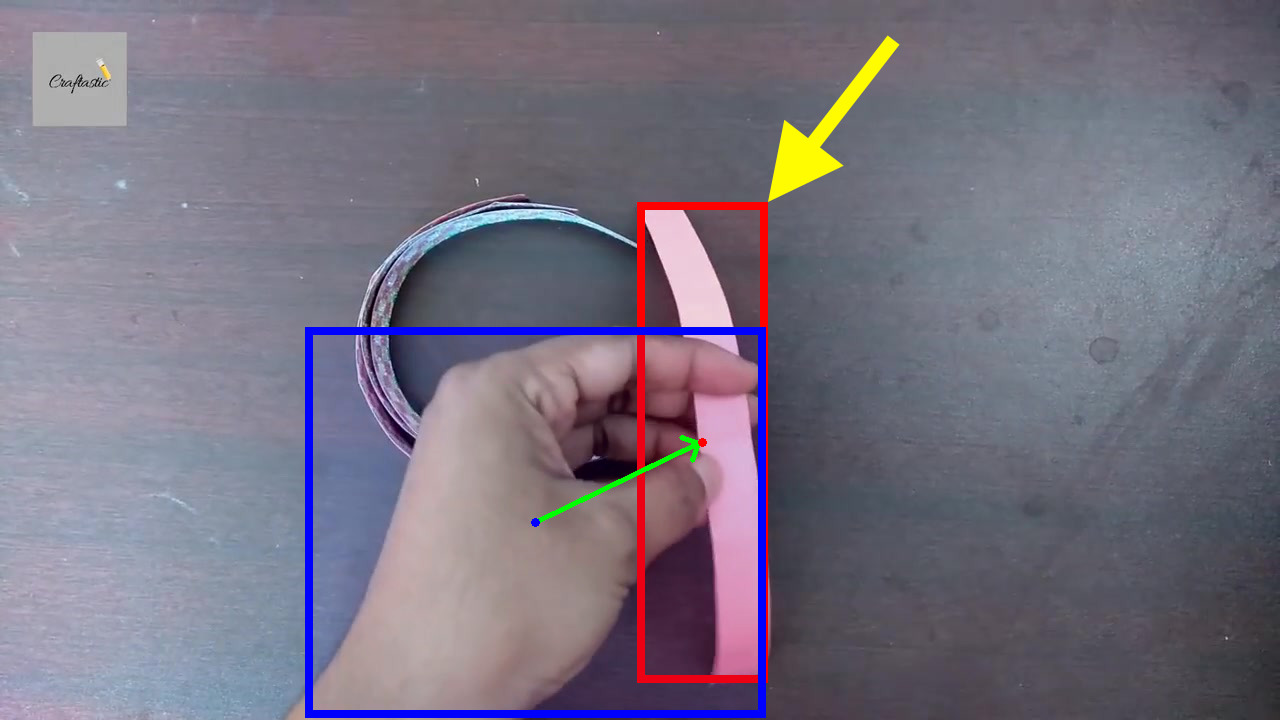}
	\\
	\rotatebox{90}{HO Detector}\hfill
	\includegraphics[width=0.19\linewidth]{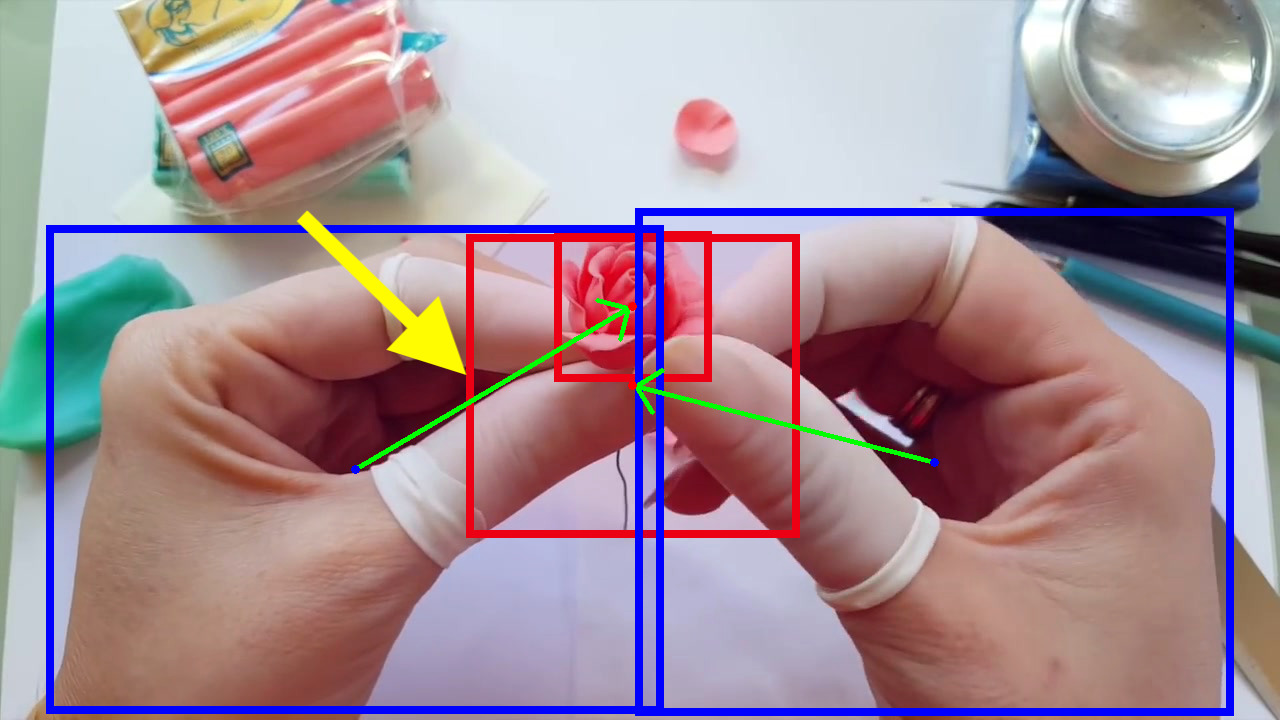}
	\includegraphics[width=0.19\linewidth]{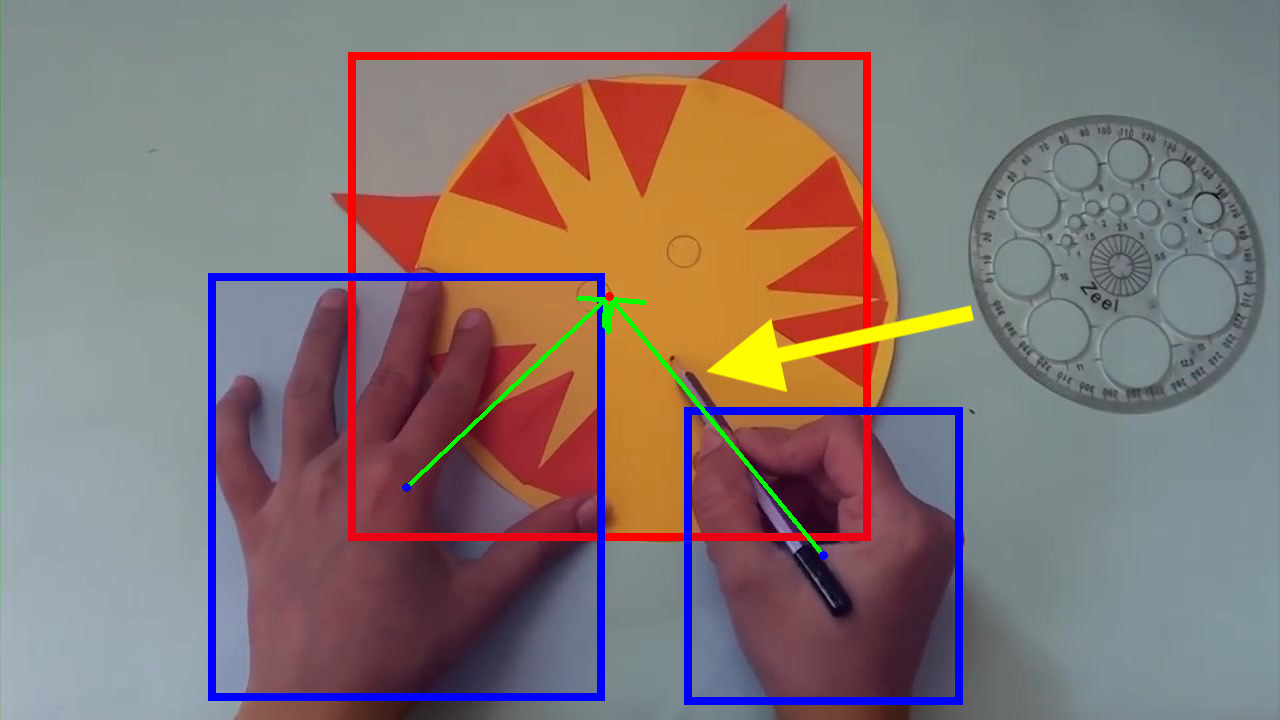}
	\includegraphics[width=0.19\linewidth]{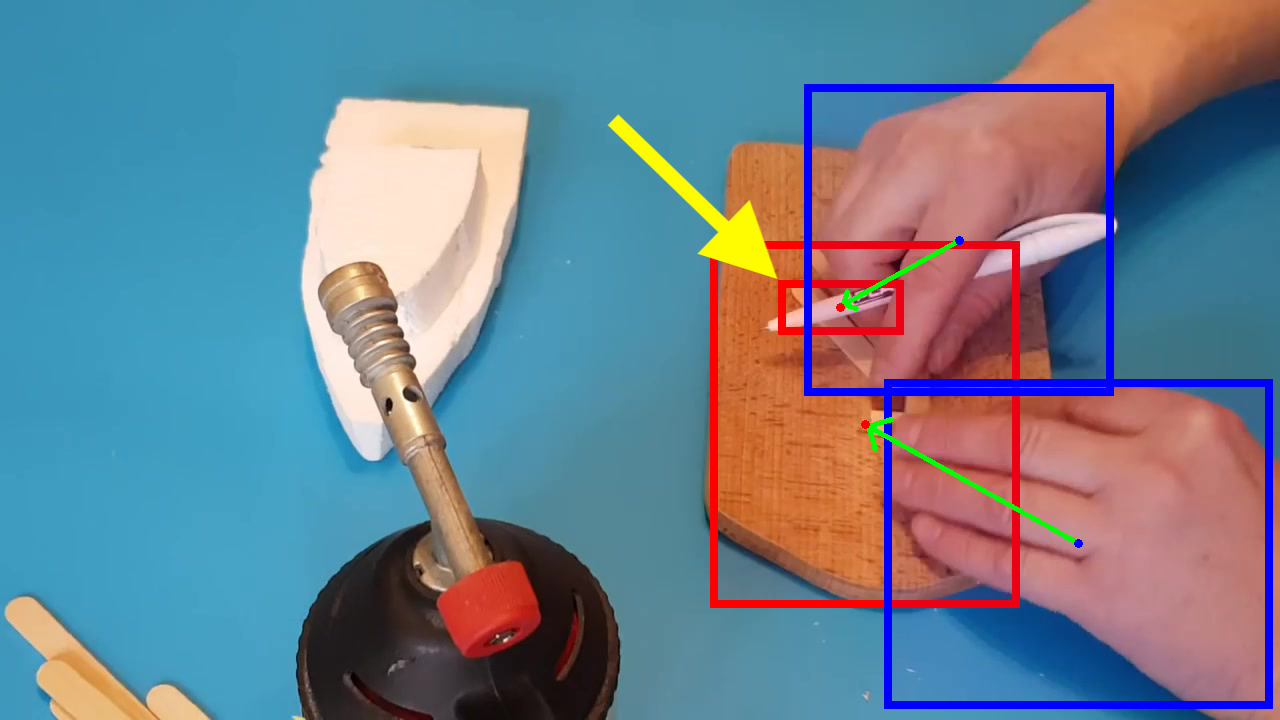}
	\includegraphics[width=0.19\linewidth]{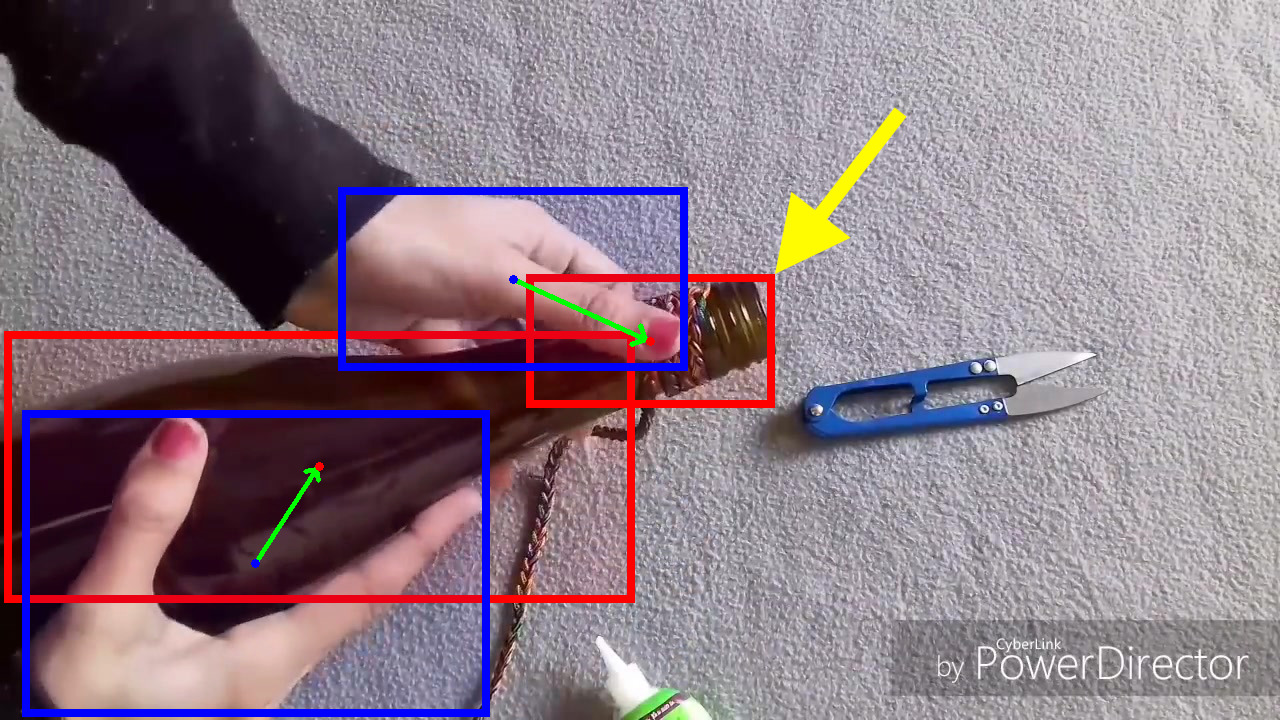}
	\includegraphics[width=0.19\linewidth]{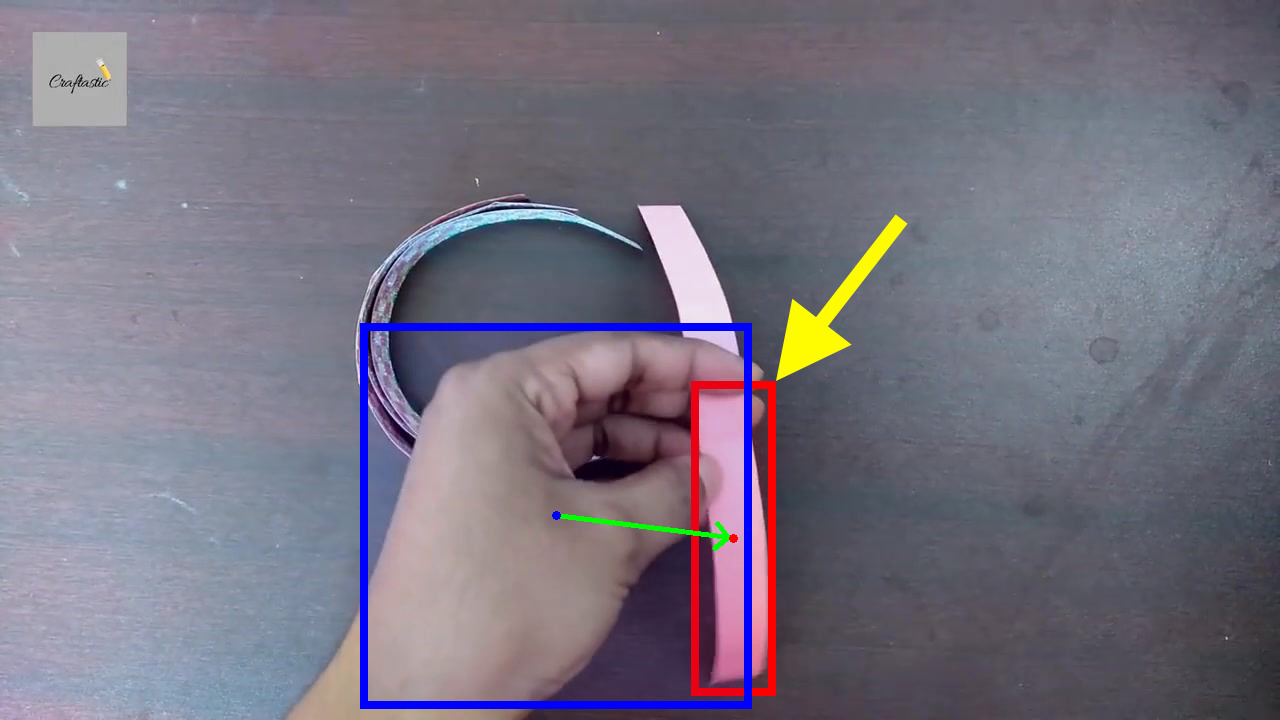}
	\\
	\rotatebox{90}{Proposed}\hfill
	\includegraphics[width=0.19\linewidth]{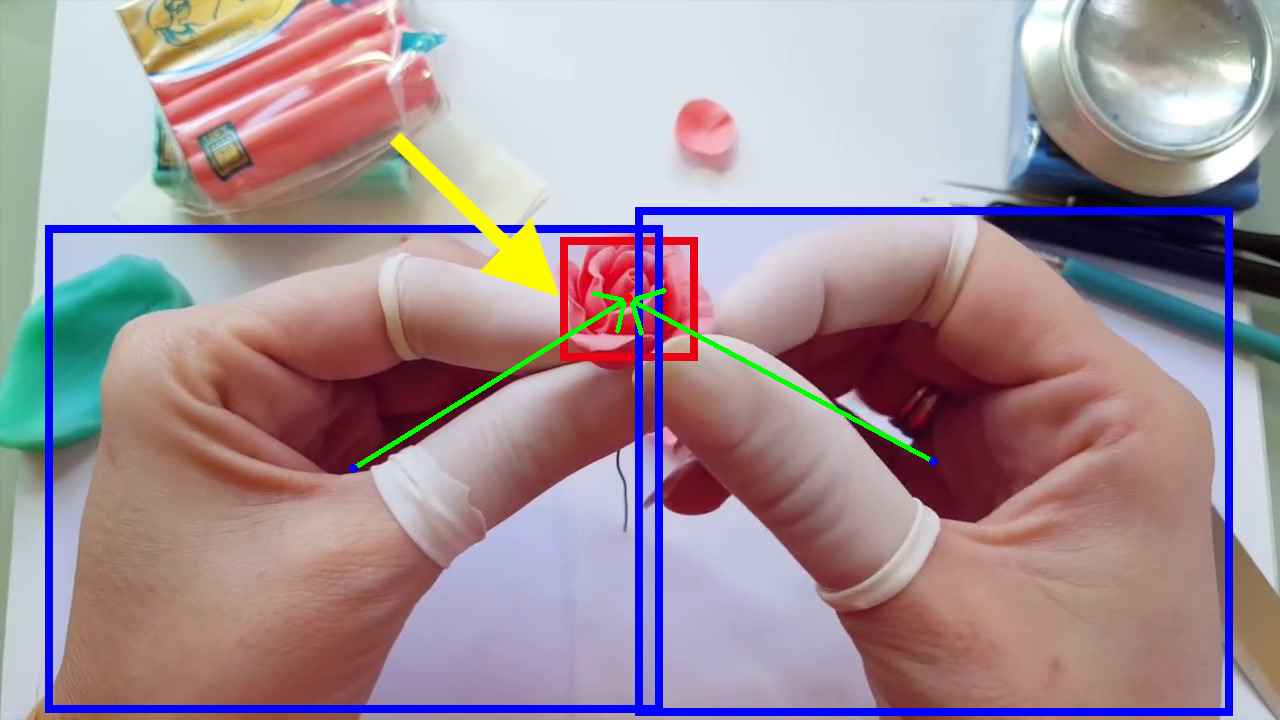}
	\includegraphics[width=0.19\linewidth]{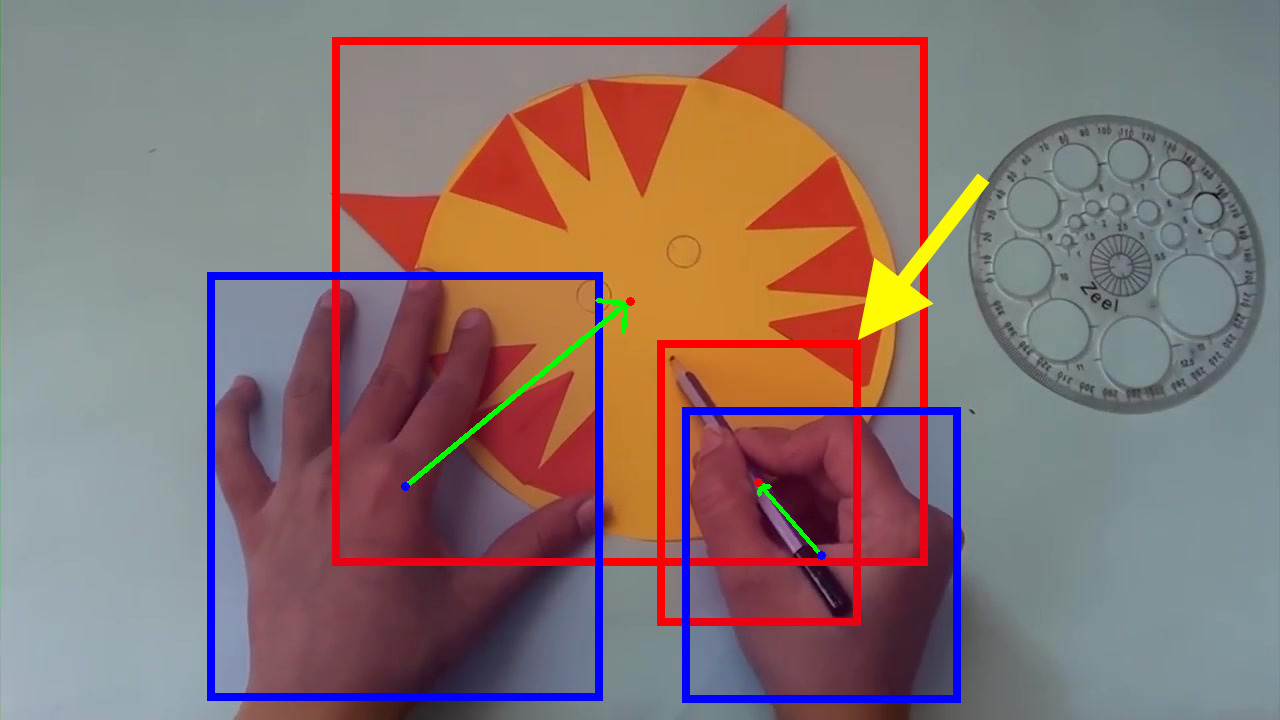}
	\includegraphics[width=0.19\linewidth]{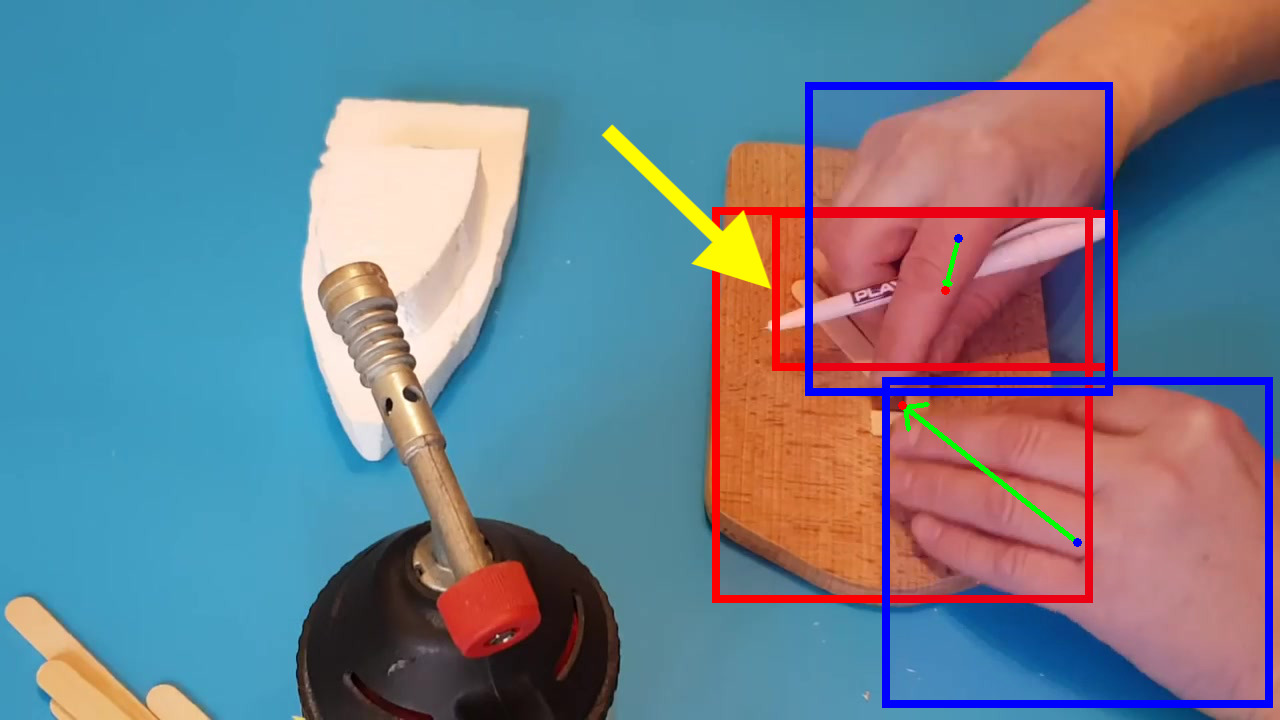}
	\includegraphics[width=0.19\linewidth]{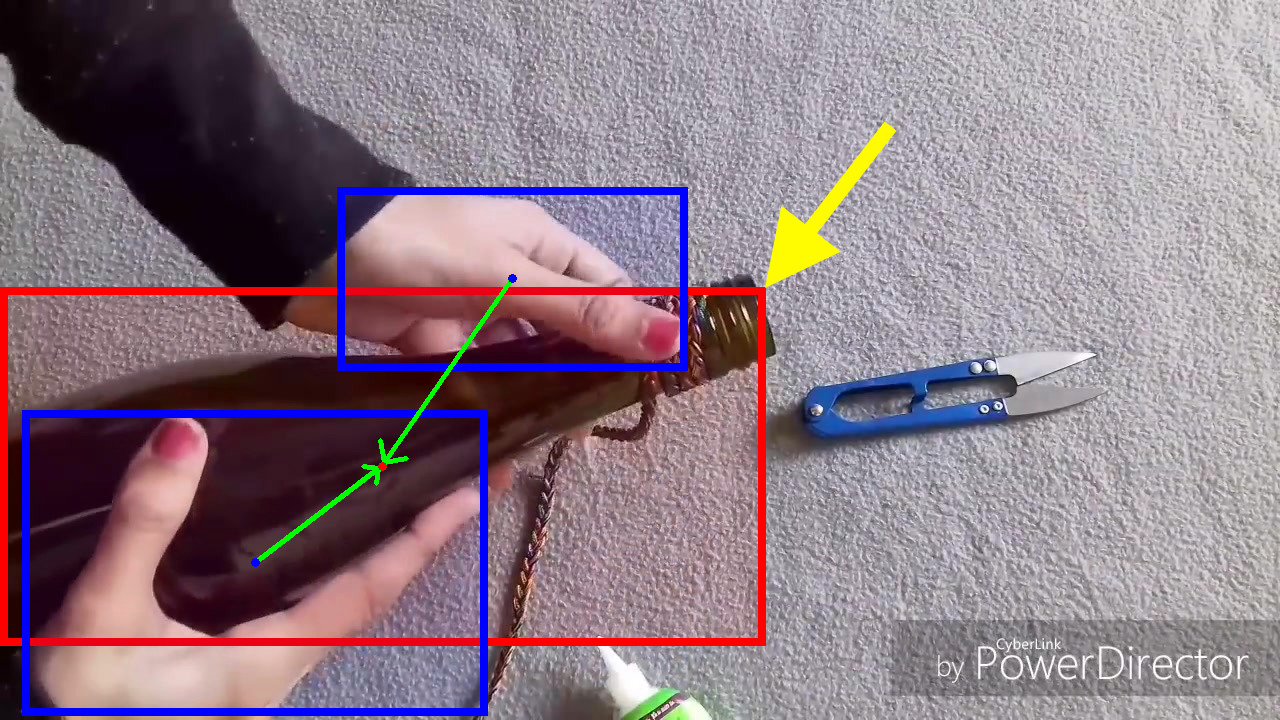}
	\includegraphics[width=0.19\linewidth]{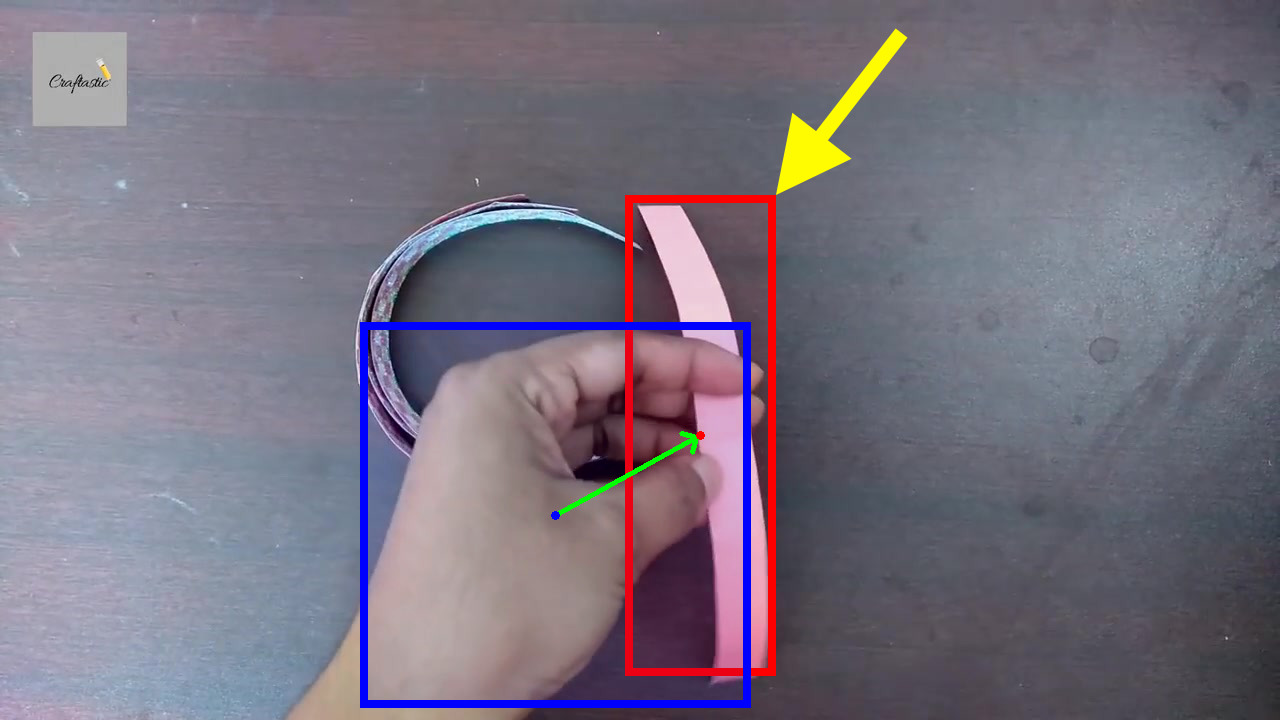}
	\\
	\caption{\label{fig:comparision_100doh} Comparison of qualitative hand-object interaction detection results on the 100DOH dataset, where HO Detector denotes 100DOH Detector \cite{100doh}. The differences are highlighted using bold yellow arrows. It demonstrates our method not only provides more precise object localization (col. 1) but also is more robust to the scenes where objects are overlapped (col. 2, 3) or occluded by hands (col. 2, 3, 4, 5).}
\end{figure*}
\subsection{Finding a Policy with Hybrid Training}
\label{ssec:method:training}

\begin{table*}[ht]
\centering
\small{
\begin{tabular}{ccccc|cccc}
\Xhline{1.0pt}  
Methods & Backbone & \# of Params & Hand Source & $AP_{hand}^{50}$ & $AP^{75}$ & $AP^{50}$ & $AP^{25}$\\
\Xhline{1.0pt}
 Simple Baseline & R101 & 47M & FasterRCNN \cite{fasterrcnn} & $89.59$ & $28.15$ & $ 44.73$ & $47.57$ \\
 100DOH Detector \cite{100doh} & R101 & 47M & FasterRCNN \cite{fasterrcnn} & $89.59$ & $28.50$ &$46.95$ & $51.80$ \\
 PPDM \cite{ppdm}& DLA34 & 21M & CenterNet \cite{centernet_zhou} & $89.64$ & $26.89$ & $45.80$ & $53.04$ \\
 HOTR \cite{hotr}& R50 & 51M & DETR \cite{detr} & $90.26$ & $29.30$ & $49.27$ & $\mathbf{57.80}$ \\
 Ours& R101 & 48M & FasterRCNN \cite{fasterrcnn} & $89.59$ & $\mathbf{29.90}$ & $\mathbf{53.02}$ & $57.15$ \\
\hline 
 Simple Baseline & R101 & 47M & Ground Truth & $100$ & $34.51$ &$44.68$ & $52.35$ \\
 Ours& R101 & 48M & Ground Truth & $100$ & $\mathbf{40.05}$ & $\mathbf{54.82}$ & $\mathbf{64.86}$ \\
\Xhline{1.0pt}
\end{tabular}
\caption{Results of hand-object interaction detection on 100DOH. The Simple Baseline method is described in \cref{ssec:exp:100doh}.}
\label{tab:res_100doh}}
\end{table*}

It is challenging to learn the image feature extractor and policy model from scratch: a massive amount of random actions, in the beginning, provide weak training signals, and optimizing the image feature extractor for each state is slow. 

While our policy uses a deterministic voting scheme to predict the action, a set of optimal hand-to-object relational box field $F^{ho}$ and object refinement relational box field $F^{oo}$ would generate the same desired behavior as learning by the reward function. This design enables us to first use imitation learning to learn these two relational box fields in a fully supervised mode. Specifically, we perform multi-task training with an overall objective which is a linear combination of losses from hand-to-object and object refinement box fields following \cref{eq:box_field_loss} to jointly pretrain the image feature extractor and policy model
\begin{align}
    L_{IL} = L_{F^{ho}}(\mathcal{H}) + L_{F^{oo}}(\mathcal{O})
\end{align}
where $\mathcal{H}$ and $\mathcal{O}$ are the set of hand bounding boxes and active object bounding boxes in the image. In imitation learning, we use a batch size of 28 and an initial learning rate of $10^{-4}$ for 100 epochs with $10\times$ learning rate drop every 30 epochs. 

After pretraining, we freeze the weights in the image feature extractor and further train the policy model using reinforcement learning (RL) on the sequential predictions for the active object $b_o$. We follow the standard policy gradient on the cumulative reward. For computational efficiency, we set the horizon $T = 5$. In consistent with the ultimate objective of getting a final accurate active object, we do not discount the future reward by setting a discount factor $\gamma = 1$. To improve the training stability, we add the losses of relational box fields as an auxiliary loss
\begin{equation}
    L_{RL} = \sum_{t=0}^{T} (1-r(s^t, a^t)) + L_{F^{ho}}(b^h) + L_{F^{oo}}(b^o)
\end{equation}

For reinforcement learning, we use a batch size of 48 and a learning rate of $10^{-5}$ for 5 epochs. We show that RL boosts the precision of localization in the supplementary.

\section{Experiments}
\label{sec:exp}

In this section, we demonstrate the performance of our approach on hand-object detection and active object detection tasks. In order to show our method is capable of tackling both third-person and egocentric real-life applications, we evaluate on two challenging datasets: 100DOH \cite{100doh} and MECCANO \cite{meccano}. 

\subsection{Experiments on 100DOH}
\label{ssec:exp:100doh}
100DOH \cite{100doh} is a large-scale benchmark for hand-object interaction. It has 99,899 frames (79,921 for training, 9,995 for validation and 9,983 for testing). Among the dataset, there are 189.6K annotated hands with 110.1K objects. For each hand, it is either paired with an active object bounding box or has no object.

We use the same hand-object detection evaluation metrics used in \cite{100doh}, which calculates the average precision (AP) of the tuple $(hand, object)$. Specifically, a tuple is considered as true positive if and only if: 
\begin{inparaenum}[1)] 
\item the IoU between the predicted hand bounding box and ground truth is greater than or equal to the IoU threshold; and 
\item the IoU between the predicted object bounding box and ground truth is greater than or equal to the IoU threshold.
\end{inparaenum}

Since active objects are usually close to hands, we construct a \textit{Simple Baseline} by first detecting hands and objects independently using Faster-CNN \cite{fasterrcnn}, and then assigning the closest object (in terms of box center Euclidean distance) to each hand as the active object. Besides \textit{Simple Baseline} and the baseline in the dataset paper, we also adapt two recent methods on human-object interaction detection, PPDM \cite{ppdm} and HOTR \cite{hotr}, to our task by changing the subject from human to hand. We assign the object with the highest score in detected interaction tuple $(hand, object)$ to the corresponding hand for evaluation.

We report the AP at IoU thresholds of 0.75, 0.5, and 0.25. As a fair comparison against the baselines, we use detected hand bounding boxes from Faster-RCNN \cite{fasterrcnn} as the input of the proposed method. As illustrated in \cref{tab:res_100doh}, our method outperforms all baseline with similar hand detection as input. We also exhibit the maximum performance when ground truth hand bounding boxes are available. Some qualitative results with comparisons are displayed in \cref{fig:comparision_100doh}, which shows the proposed method not only provides more precise object localization but also is more robust to occlusions. More qualitative results and visualization of object refinement are shown in the supplementary.

\subsection{Experiments on MECCANO}
\label{ssec:exp:meccano}

MECCANO \cite{meccano} is an egocentric dataset for human-object interaction understanding. It contains 64,349 frames which are annotated with active object boxes. Since there is no hand bounding box and hand-object correspondence annotation in the MECCANO dataset, we perform Pseudo labeling in two steps to train our model. First, we adapt the pre-trained hand detector \cite{100doh} to detect hands in all frames. Second, we assign the closest annotated active object to each hand box in terms of box centers. There is no human annotation involved in the Pseudo labeling, which leads to a fair comparison.

Since MECCANO does not have the interaction annotation to train PPDM \cite{ppdm} and HOTR \cite{hotr}, we compare our methods against 100DOH Detector \cite{100doh} in terms of standard object detection metrics: average precision (AP) at IoU thresholds of 0.75, 0.5, and 0.25. We first compare the generalization ability by directly evaluating models trained on 100DOH. Then we compare the performance by retraining the proposed method using automatically generated pseudo labels of MECCANO. \cref{tab:res_meccano} shows that our method has a better generalization ability when adapted to a new dataset without retraining, and better performance after retraining on the MECCANO dataset. More qualitative results and visualization of object refinement are shown in the supplementary.

\subsection{Ablation Studies}
\label{ssec:exp:ablation}

\begin{table}[t]
\centering
\small{
\setlength\tabcolsep{1pt}
\begin{tabular}{ccc|ccc}
\Xhline{1.0pt}  
Method & Backbone & Finetune & $AP^{75}$ & $AP^{50}$ & $AP^{25}$ \\
\Xhline{1.0pt}
 100DOH Detector \cite{100doh} & R101 & \xmark & - &$11.17$ & - \\
 Ours& R101 & \xmark  & $\mathbf{9.09}$ & $\mathbf{16.61}$ & $\mathbf{23.97}$ \\
\hline 
 100DOH Detector \cite{100doh} & R101 & \cmark & - &$20.18$ & - \\
 Ours& R101 & \cmark & $\mathbf{12.99}$ & $\mathbf{26.25}$ & $\mathbf{34.88}$\\
\Xhline{1.0pt}
\end{tabular}
\caption{Results of active object detection on MECCANO. We compare our method against 100DOH Detector \cite{100doh}.}
\label{tab:res_meccano}
}
\end{table}

\begin{table}[t]
\centering
\small{
\begin{tabular}{cc|ccc}
\Xhline{1.0pt}  
Dataset & Aggregation & $AP^{75}$ & $AP^{50}$ & $AP^{25}$ \\
\Xhline{1.0pt}
 100DOH & Center & $25.00$ & $52.31$ & $56.88$\\
 100DOH & Average & $25.59$ & $51.77$ & $56.41$\\
 100DOH & Vote & $\mathbf{29.90}$ & $\mathbf{53.02}$ & $\mathbf{57.15}$ \\
\hline 
 MECCANO & Center & $9.59$ & $22.85$ & $32.78$\\
 MECCANO & Average & $12.83$ & $25.76$ & $34.66$\\
 MECCANO & Vote & $\mathbf{12.99}$ & $\mathbf{26.25}$ & $\mathbf{34.88}$\\
\Xhline{1.0pt}
\end{tabular}
\caption{Ablation studies on different aggregation methods to retrieve estimation from dense predictions.}
\label{tab:ablation_voting}
}
\end{table}

We perform ablation studies on both 100DOH and MECCANO datasets. Specifically, we examine the effects of pixel-wise voting and the sequential decision-making process. More ablation study, runtime analysis, and discussion about limitations are included in the supplementary. 

\paragraph{Effect of Pixel-wise Voting} The pixel-wise weighted voting is the key component of the proposed method to retrieve a single bounding box estimation from dense predictions while being robust to occlusions.
To understand the effect of pixel-wise weighted voting, we visualize the heatmap of the IoU between pixel-wise bounding box predictions and the final predicted bounding box after voting aggregation in \cref{fig:vote_correctness}. As expected, voting picks the final estimated bounding box mostly based on the predictions in the regions of informative patterns such as fingers and objects as opposed to irrelevant information such as the background. We further quantitatively examine the effectiveness of pixel-wise voting by comparing it with two aggregation methods:
\begin{inparaenum}[1)] 
\item using the prediction from the central pixel of the input box, and
\item averaging the bounding box parameters of all predictions inside the input box.
\end{inparaenum} Table \ref{tab:ablation_voting} reveals the superiority of using voting to aggregate dense predictions.

\begin{table}[t]
\centering
\small{
\begin{tabular}{cc|ccc}
\Xhline{1.0pt}  
Dataset & \# of Voting & $AP^{75}$ & $AP^{50}$ & $AP^{25}$ \\
\Xhline{1.0pt}
 100DOH & 1 & $29.18$ & $49.58$ & $53.11$\\
 100DOH & 2 & $29.80$ & $\mathbf{53.11}$ & $\mathbf{57.47}$\\
 100DOH & 3 & $\mathbf{29.92}$ & $53.10$ & $57.23$\\
 100DOH & $\infty$ & $29.90$ & $53.02$ & $57.15$ \\
\hline 
 MECCANO & 1 & $7.33$ & $22.61$ & $32.94$\\
 MECCANO & 2 & $8.97$ & $22.56$ & $32.46$\\
 MECCANO & 3 & $10.38$ & $23.91$ & $33.03$\\
 MECCANO & $\infty$ &  $\mathbf{12.99}$ & $\mathbf{26.25}$ & $\mathbf{34.88}$\\
\Xhline{1.0pt}
\end{tabular}
\caption{Ablation studies on the number of voting of the sequential decision-making process.}
\label{tab:ablation_horizon}
}
\end{table} 

\paragraph{Effect of Sequential Decision-Making Process} We analyze whether applying the voting function multiple times could sequentially improve the active object bounding box estimation. Specifically, we report the performance after applying different numbers of the voting function. The results are shown in \cref{tab:ablation_horizon}. For the 100DOH dataset, the performance converges after applying the voting function two times. On the MECCANO dataset, the performance is progressively improved with more iterations of voting.

\section{Conclusion}
\label{sec:conclusion}

In this paper, we propose a voting function with Relational Box Field to leverage each pixel as evidence to robustly predict the bounding box of the active object, despite under occlusions. The voting function is applied repeatedly to improve the active object estimation. We use an MDP to model the sequential decision-making process and apply reinforcement learning to learn an optimal policy.
Our method achieves state-of-the-art performance on both hand-object detection and active object detection tasks, as well as better generalization ability across datasets. 
In the future, we will try to develop a stochastic MDP policy to further explore the state space.

\noindent\textbf{Acknowledgement: } This work is funded in part by JST AIP Acceleration, Grant Number JPMJCR20U1, Japan.

{\small
\bibliographystyle{ieee_fullname}
\bibliography{egbib}
}

\newpage
\appendix
\section*{Supplementary}

\section{Overview}

In this document, we provide additional implementation and experimental details, as well as qualitative results and analysis. 
We present the details of the self-attention layer in \cref{apdx:ssec:self-attention} and the confidence calculation in \cref{apdx:ssec:confidence}.
We validate that the proposed method is more robust to detect active objects under occlusion in \cref{apdx:sec:occulusion}.
We show the effect of reinforcement learning with additional ablation study in \cref{apdx:ssec:rl_effect}.
We illustrate additional qualitative results and visualizations in \cref{apdx:ssec:qualitative}. 
The inference running time is presented in \cref{apdx:ssec:runtime}.

\section{Implementation Details}

\subsection{Self-attention Layer}
\label{apdx:ssec:self-attention}

\begin{figure}[t]
	\centering
	\includegraphics[width=0.6\linewidth]{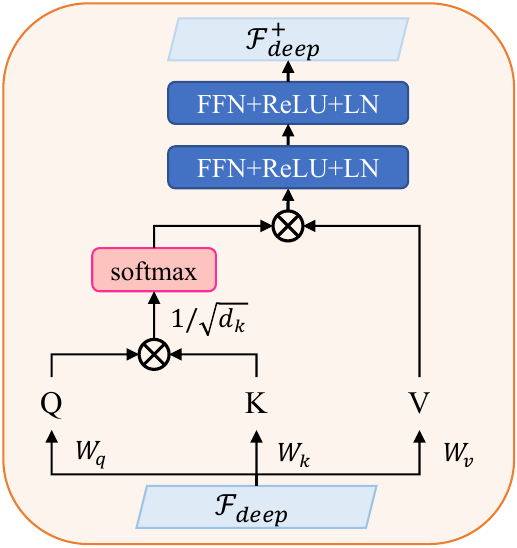}
	\centering
	\caption{\label{fig:self-attention} The architecture of the self-attention layer}
\end{figure}

As described in Sec. 3.3 of the main paper, inside the image feature extractor, we use a self-attention layer between the encoder and the decoder to further exploit the synergy between hands and objects.
The architecture of the self-attention layer is illustrated in \cref{fig:self-attention}. The self-attention layer takes the image feature $\mathcal{F}_{\text{deep}}$ from the encoder as input and computes the query, key, and value embeddings ($Q$, $K$, and $V$) from  $\mathcal{F}_{\text{deep}}$ using learnable embedding matrices $W_q$, $W_k$ and $W_v$. Then the relationships between every spatial location in the feature map are computed using query $Q$ and key $K$, which is used as the weight to average $V$. Finally, a two-layer MLP with layer normalization \cite{layernorm} is applied as
\begin{equation}
    \begin{split}
    &Q = W_q \mathcal{F}_{\text{deep}}, K = W_k \mathcal{F}_{\text{deep}}, V = W_v \mathcal{F}_{\text{deep}}\\
    &\mathcal{F}_{\text{deep}}^+ = \text{MLP}(\text{softmax}(\frac{QK^T}{\sqrt{d_k}})V)
    \end{split}
\end{equation}
where $d_k$ is the feature dimension of the key and $\mathcal{F}_{\text{deep}}^+$ is the feature map after applying self-attention, which is forwarded to the decoder. Empirically, we find marginal improvement from positional encoding, so we omit it for simplicity.

In order to avoid exhaustive computation, we set $d_k = 256$, and reduce the feature dimension of $\mathcal{F}_{\text{deep}}$ from $2048$ to $512$ by a convolutional layer with a kernel size of $1\times 1$. Following \cite{transformer} , we use $8$ attention heads to address multiple relations between hands and objects. 

\subsection{Confidence Calculation}
\label{apdx:ssec:confidence}

By definition, an active object must be manipulated by a human hand. We first predict a contact score $s^{\text{contact}}_{b^h}$ representing the probability that a given hand $b^h$ is manipulating an object. Besides, we predict a object probability score $s^{\text{obj}}_{\hat{b}^o}$ for the final object estimation $\hat{b}^o$ of the given hand. To compute $s^{\text{contact}}_{b^h}$ and $s^{\text{obj}}_{\hat{b}^o}$, we use the average of confidence scores inside predicted hand-to-object $\hat{F}^{ho}$ and object refinement $\hat{F}^{oo}$ box fields as

\begin{equation}
     s^{\text{contact}}_{b^h} = \frac{\sum_{u,v \in b^h} \hat{c}^{ho}_{u, v}}{|b^h|}, 
     s^{\text{obj}}_{\hat{b}^o} = \frac{\sum_{u,v \in \hat{b}^o} \hat{c}^{oo}_{u, v}}{|\hat{b}^o|}
\end{equation}

We suppress the object detection by a object probability threshold $t_{\text{obj}}$. The final confidence $\hat{c}_{b^h}$ of the hand $b^h$ is the fusion of the hand contact score and the object probability score defined as
\begin{equation}
    \hat{c}_{b^h} =\begin{cases}
     1 - s^{\text{contact}}_{b^h}& \text{if }s^{\text{contact}}_{b^h} < t_{\text{contact}}\\
     s^{\text{contact}}_{b^h} \cdot s^{obj}_{\hat{b}^o}& \text{otherwise}\\
    \end{cases}
\end{equation}
We use $t_{\text{obj}} = 0.2$, $t_{\text{contact}} = 0.1$ in all our experiments.

\section{Analysis: Robustness to Occlusions}
\label{apdx:sec:occulusion}
To analyze the robustness of our method to occlusions, we compute the recall on the hand-object pairs with three different occlusion levels on 100DOH dataset. The occlusion level of a hand-object pair is measured by the IoU between their bounding boxes. In 100DOH dataset, there are 2222 hand-object pairs with an IoU $\in[.25,.5)$, 348 pairs with an IoU $\in[.5,.75)$, and 14 pairs with an IoU $\in[.75,.1]$. The quantitative comparison in \cref{tab:occluded_recall} shows that our method is more robust in detecting active objects under occlusions over all baselines.

\begin{table}[h]
    \centering
    \resizebox{\linewidth}{!}{
    \begin{tabular}{c|ccc}
        \hline
        Method & Recall(IoU$\in[.25,.5)$) & Recall (IoU$\in [.5,.75)$) & Recall(IoU$\in [.75,1]$) \\ \hline
        100DOH Detector & $68.68$ & $63.22$ & $78.57$ \\ \hline
        PPDM & $53.24$ & $53.45$ & $64.29$ \\ \hline
        HOTR & $71.69$ & $68.10$ & $71.43$ \\ \hline
        Ours & $\mathbf{77.22}$ & $\mathbf{78.45}$ & $\mathbf{100}$ \\ \hline
    \end{tabular}}
    \caption{Results of hand-object interaction detection for hand-object pairs with different occlusion levels on 100DOH dataset.}
    \label{tab:occluded_recall}
\end{table}

\section{Ablation: Effect of Reinforcement Learning}
\label{apdx:ssec:rl_effect}
Repeatedly applying the voting function trained for one-step prediction (supervised learning) could result in a data distribution shift issue. Specifically, the small error at each step could compound the sequential predictions, which leads to a bad performance towards the final prediction. The application of RL is to mitigate this issue by optimizing over the sequence with an accumulative loss for the sequential predictions. We examine the effect of RL by comparing the performance with and without using RL. The results are shown in \cref{tab:ablation_rl}, which demonstrate that RL gives significant improvements for $AP^{75}$ and $AP^{50}$ on 100DOH dataset.

\begin{table}[t]
\centering
\small{
\begin{tabular}{cc|ccc}
\Xhline{1.0pt}  
Dataset & RL & $AP^{75}$ & $AP^{50}$ & $AP^{25}$ \\
\Xhline{1.0pt}
 100DOH & \xmark & $23.64$ & $46.84$ & $\mathbf{57.44}$\\
 100DOH  & \cmark & $\mathbf{29.90}$ & $\mathbf{53.02}$ & $57.15$ \\
\hline 
 MECCANO & \xmark & $\mathbf{13.13}$ & $26.21$ & $34.88$\\
 MECCANO  & \cmark & $12.99$ & $\mathbf{26.25}$ & $\mathbf{34.88}$\\
\Xhline{1.0pt}
\end{tabular}
\caption{Ablation studies on reinforcement learning (RL) on 100DOH and MECCANO datasets.}
\label{tab:ablation_rl}
}
\end{table}

\section{Visualizations}
\label{apdx:ssec:qualitative}

\begin{figure*}[t]
	\centering
	\rotatebox{90}{General}\hfill
	\includegraphics[width=0.19\linewidth]{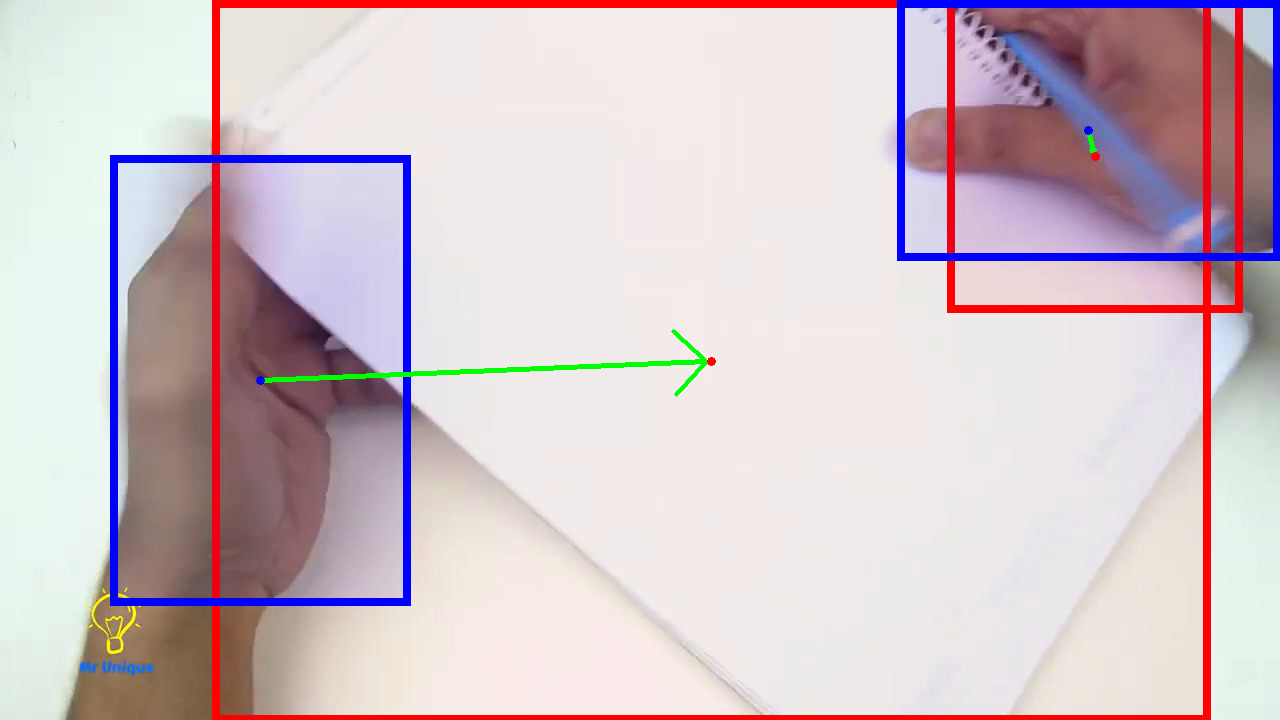}
	\includegraphics[width=0.19\linewidth]{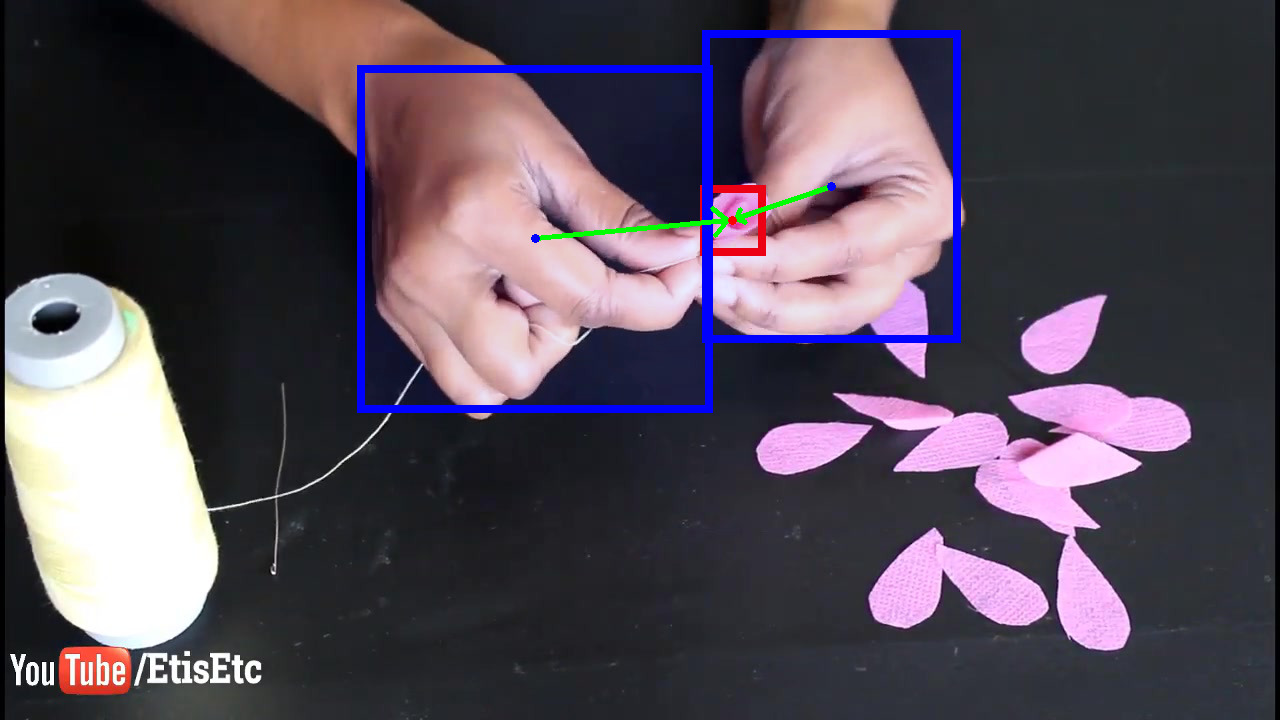}
	\includegraphics[width=0.19\linewidth]{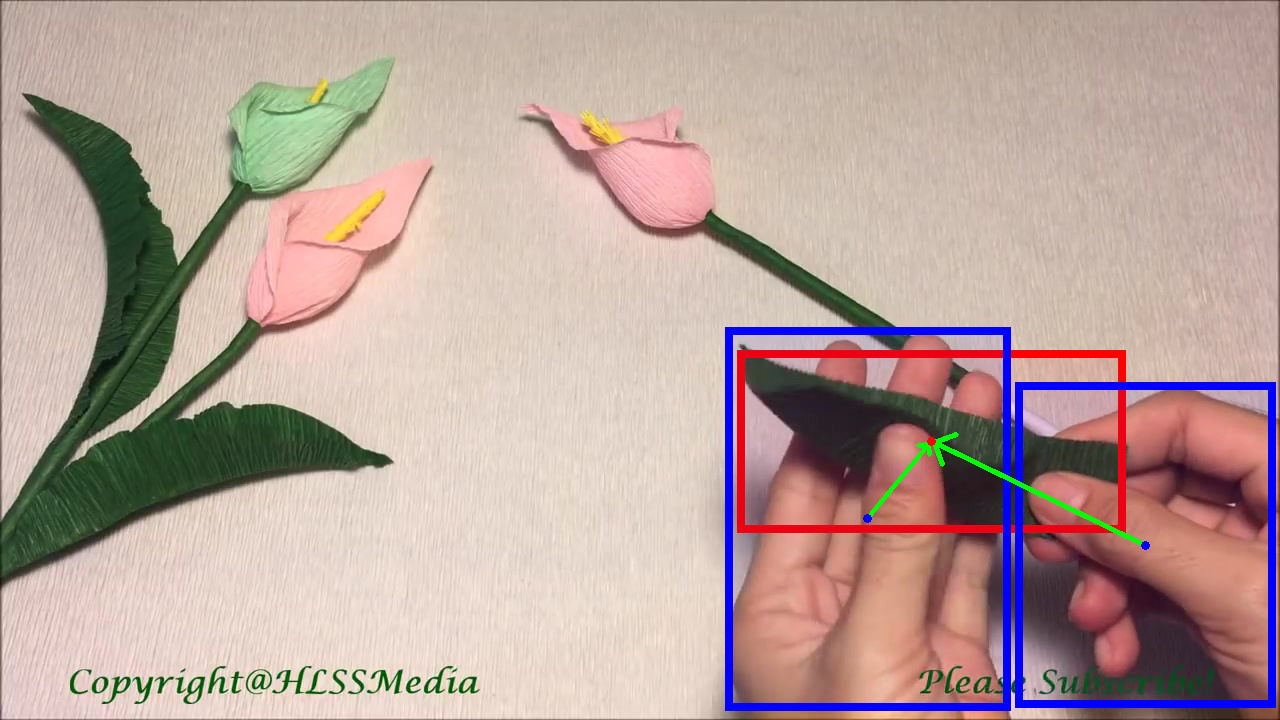}
	\includegraphics[width=0.19\linewidth]{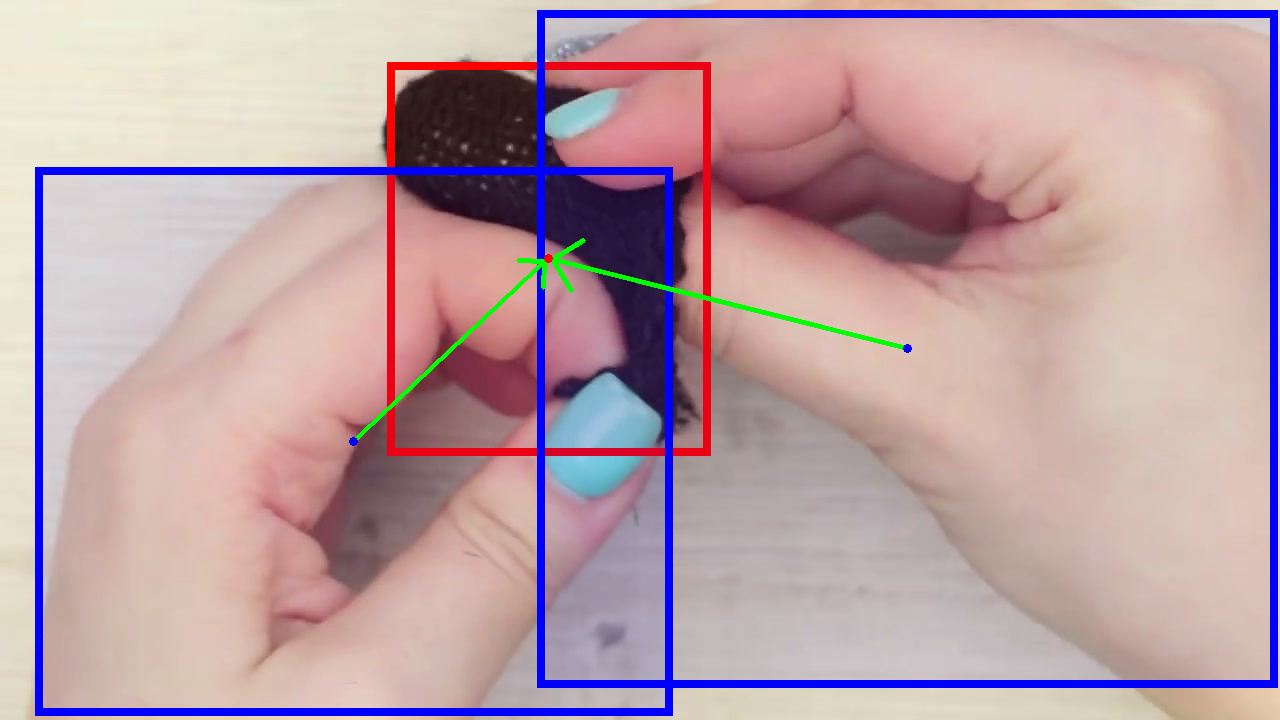}
	\includegraphics[width=0.19\linewidth]{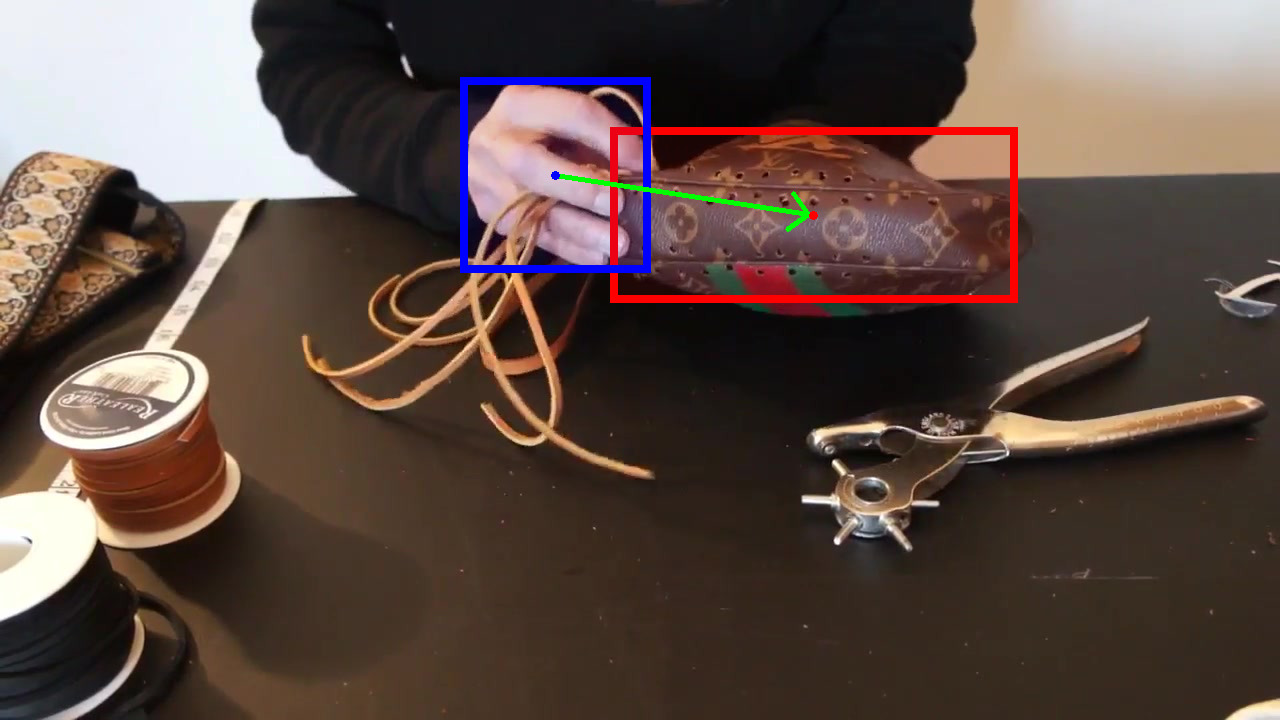}\\
	\rotatebox{90}{Occlusion}\hfill
	\includegraphics[width=0.19\linewidth]{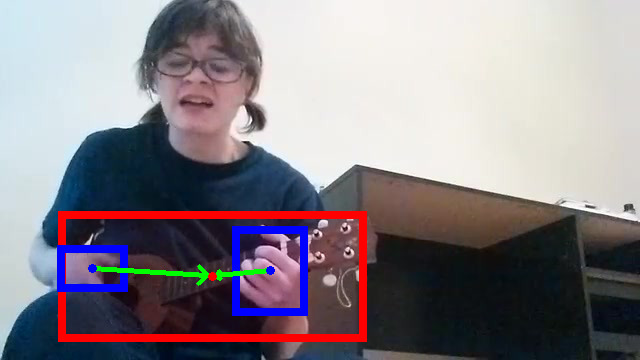}
	\includegraphics[width=0.19\linewidth]{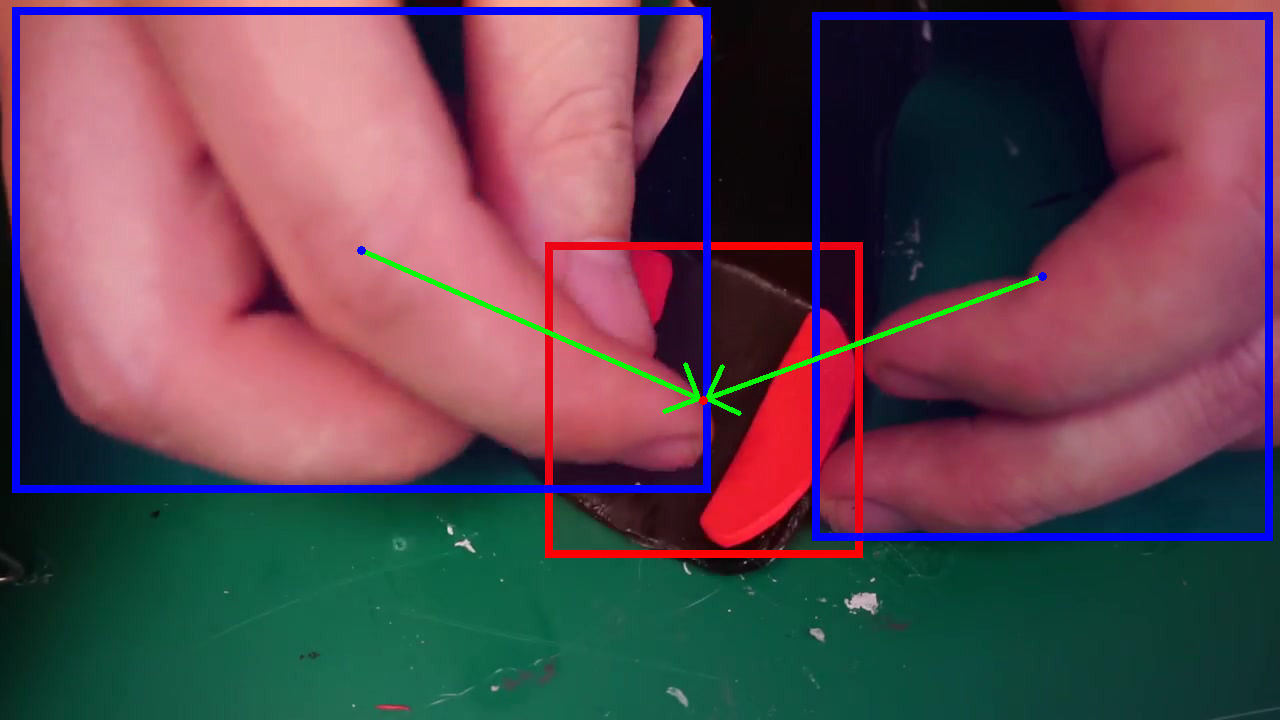}
	\includegraphics[width=0.19\linewidth]{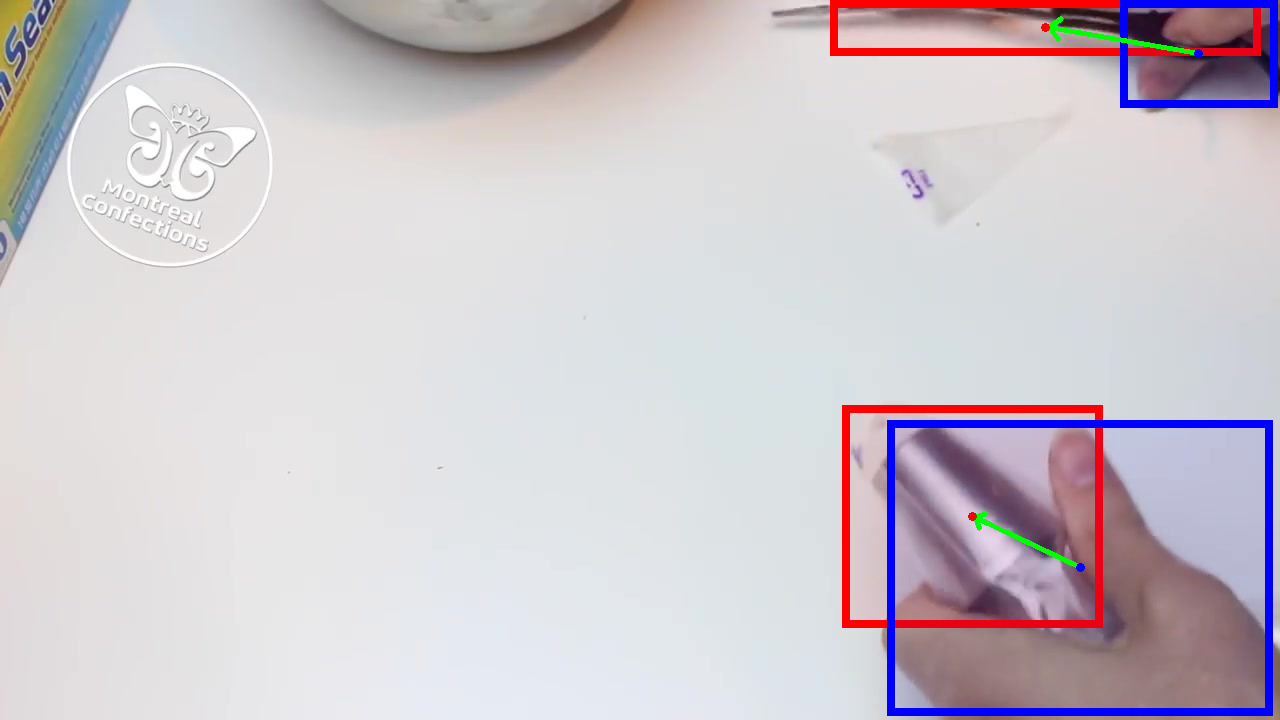}
	\includegraphics[width=0.19\linewidth]{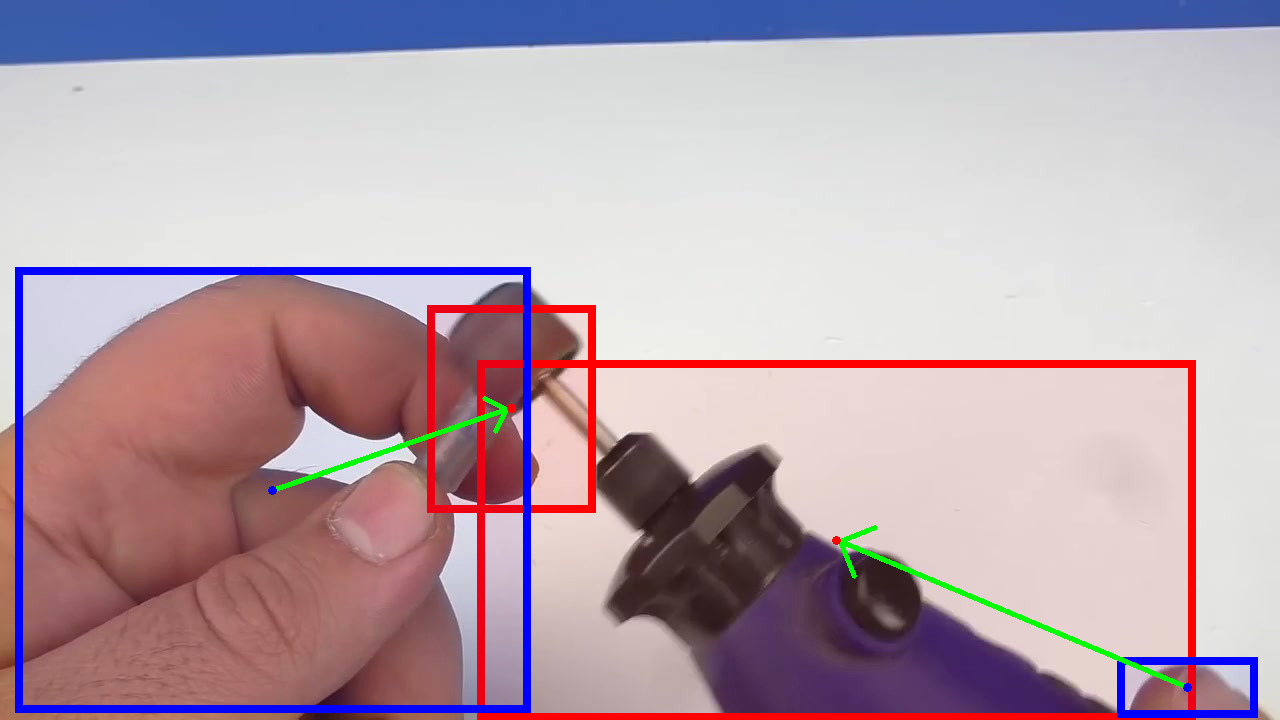}
	\includegraphics[width=0.19\linewidth]{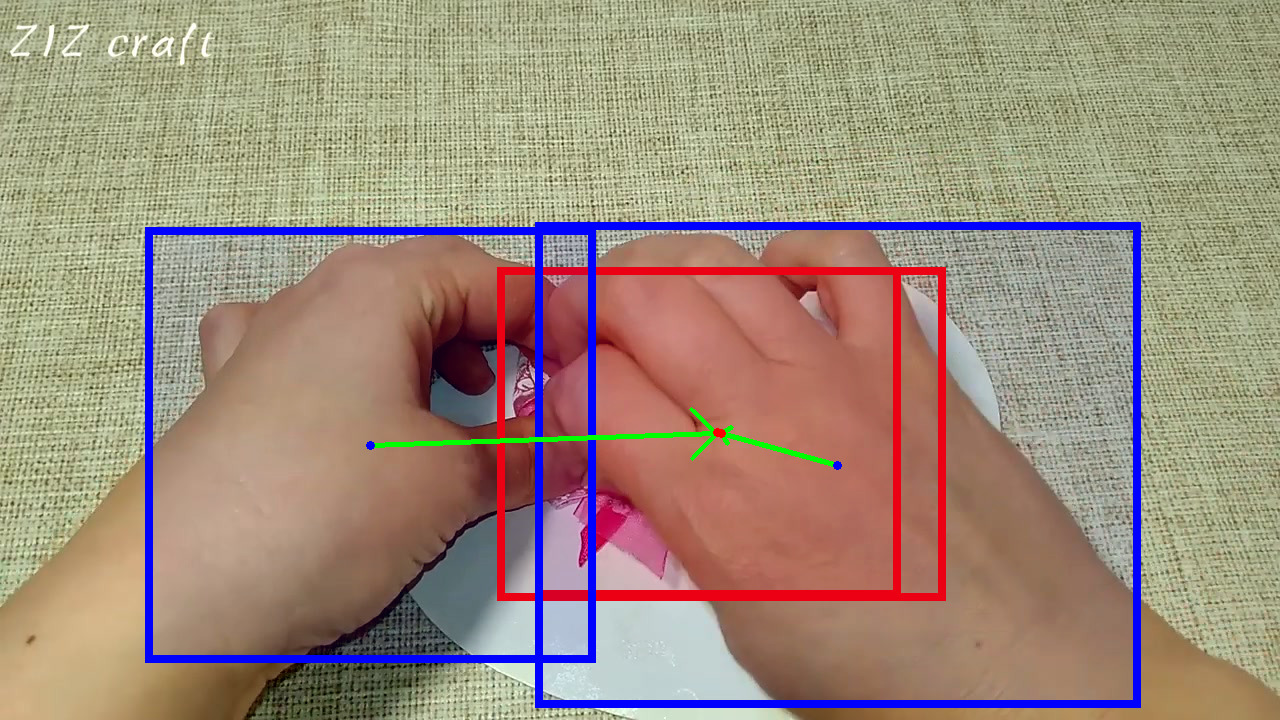}\\
	\rotatebox{90}{Failure}\hfill
	\includegraphics[width=0.19\linewidth]{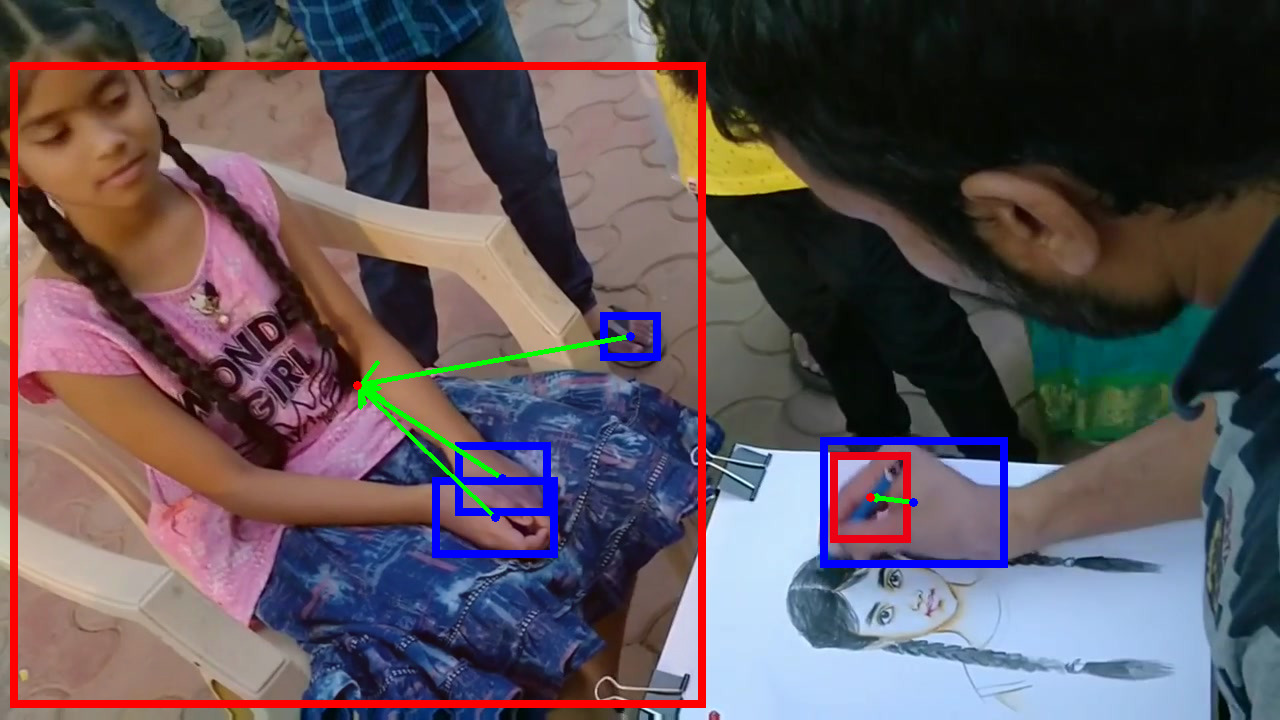}
	\includegraphics[width=0.19\linewidth]{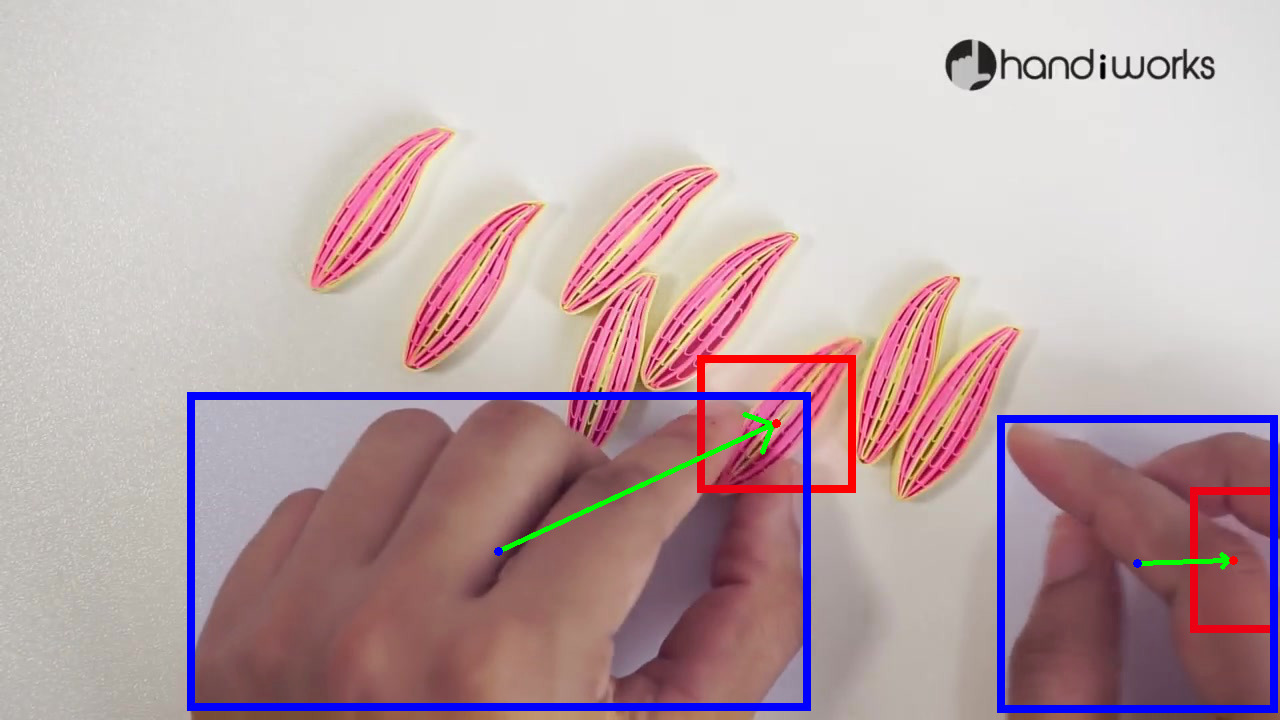}
	\includegraphics[width=0.19\linewidth]{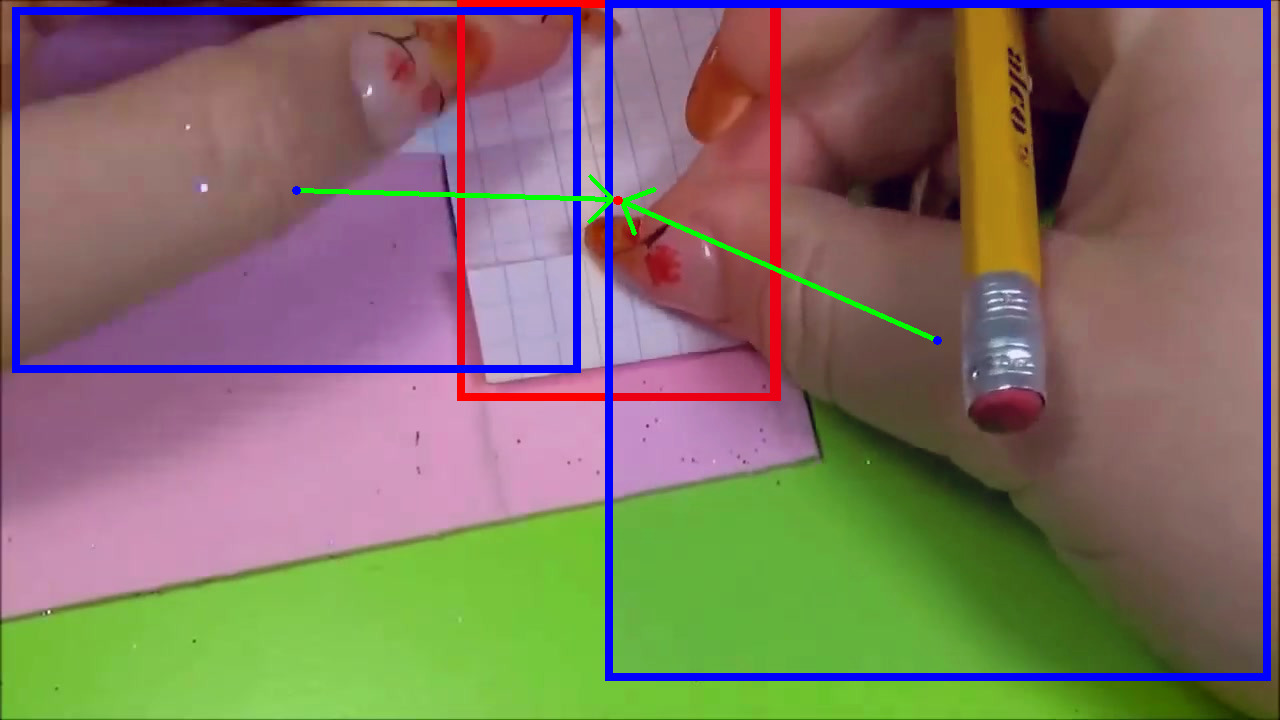}
	\includegraphics[width=0.19\linewidth]{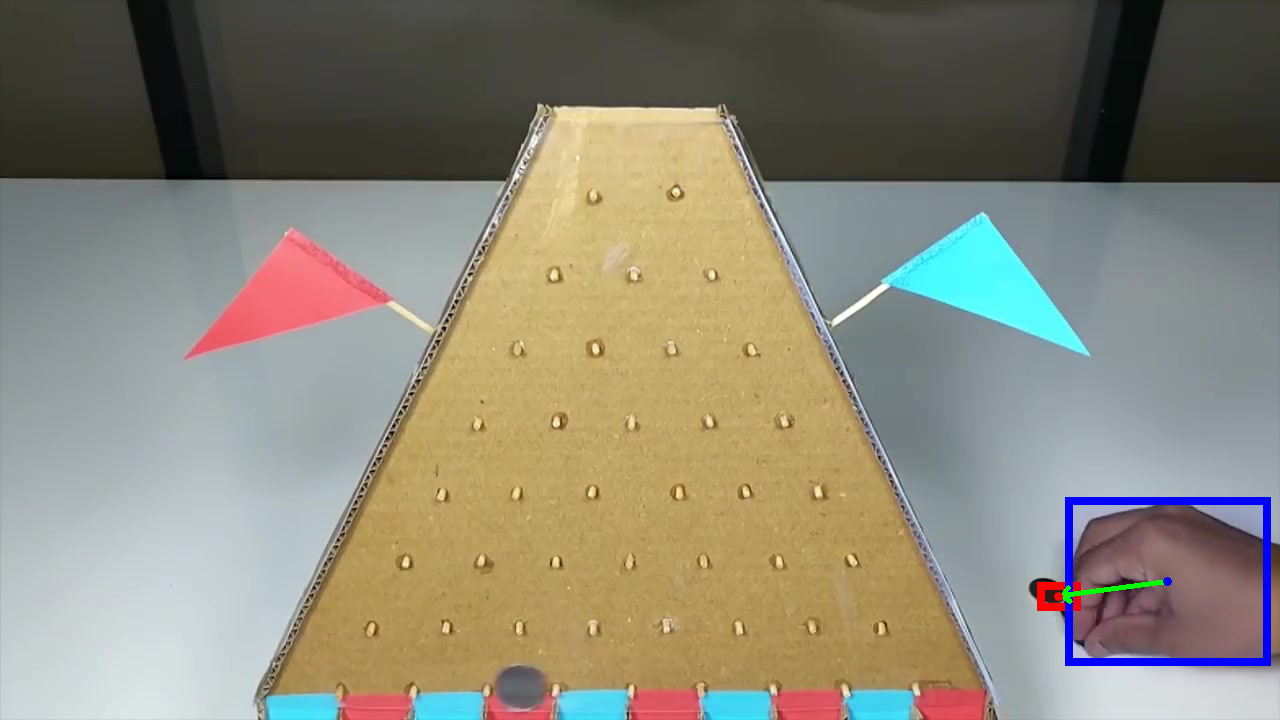}
	\includegraphics[width=0.19\linewidth]{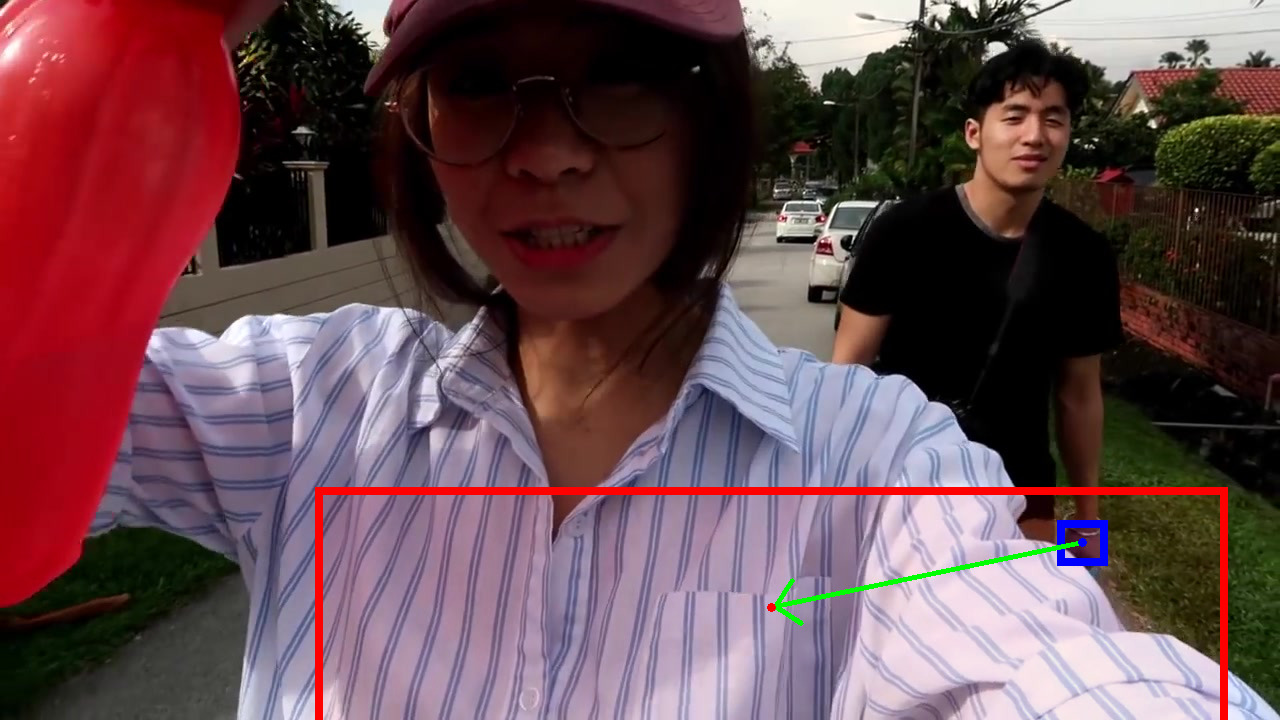}\\
	\centering
	\caption{\label{fig:qualitative_100doh} Qualitative Results on the 100DOH dataset. Each green arrow points from a hand bounding box (blue) to the corresponding active object bounding box (red).}
\end{figure*}

\begin{figure*}[t]
	\centering
	\rotatebox{90}{General}\hfill
	\includegraphics[width=0.19\linewidth]{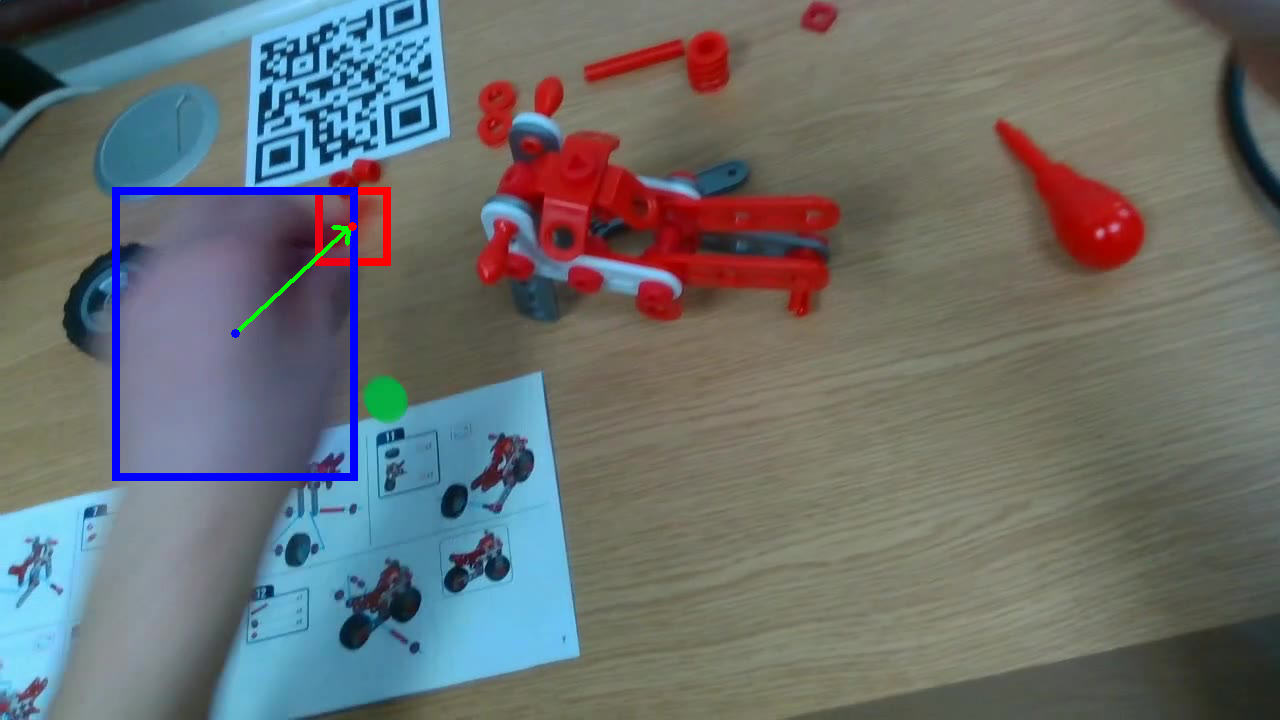}
	\includegraphics[width=0.19\linewidth]{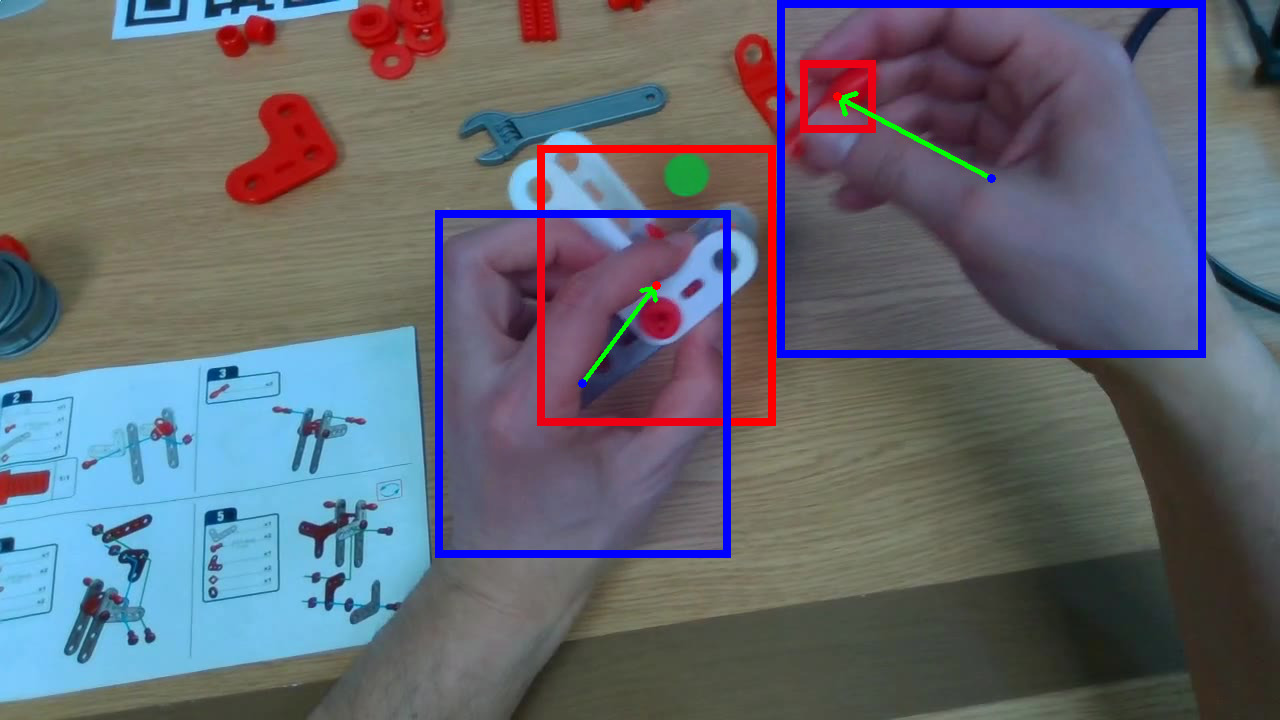}
	\includegraphics[width=0.19\linewidth]{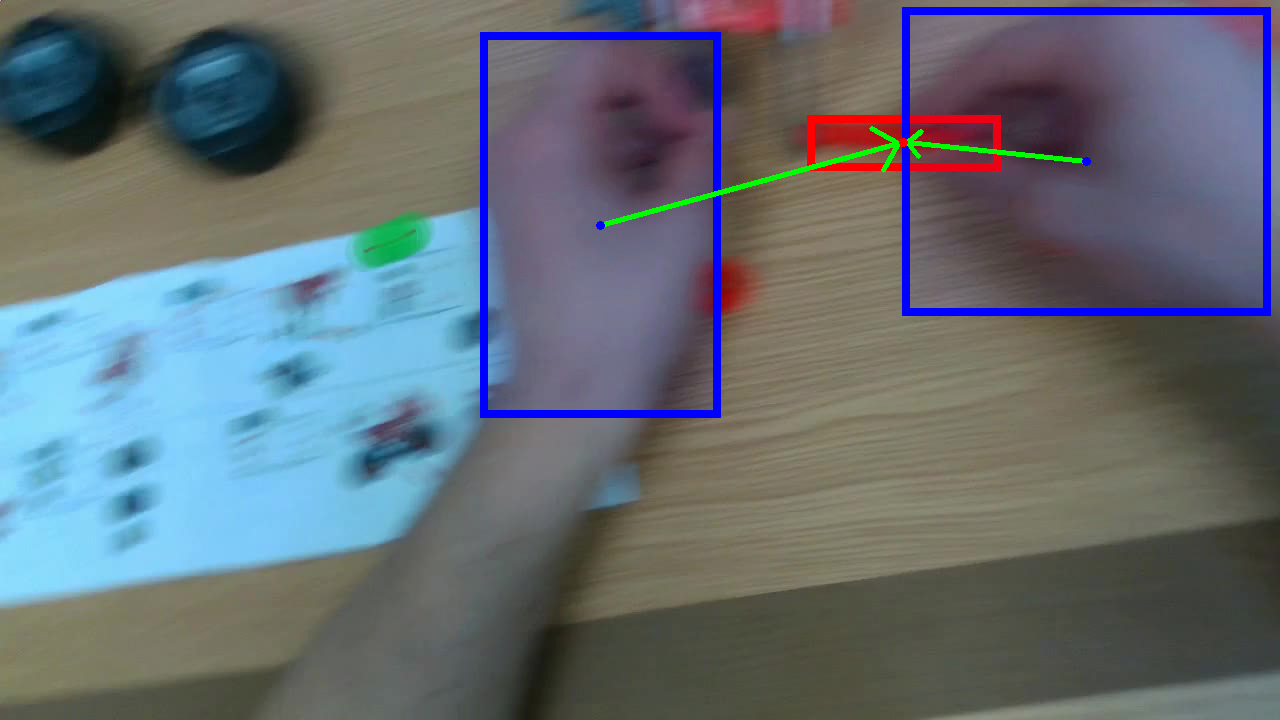}
	\includegraphics[width=0.19\linewidth]{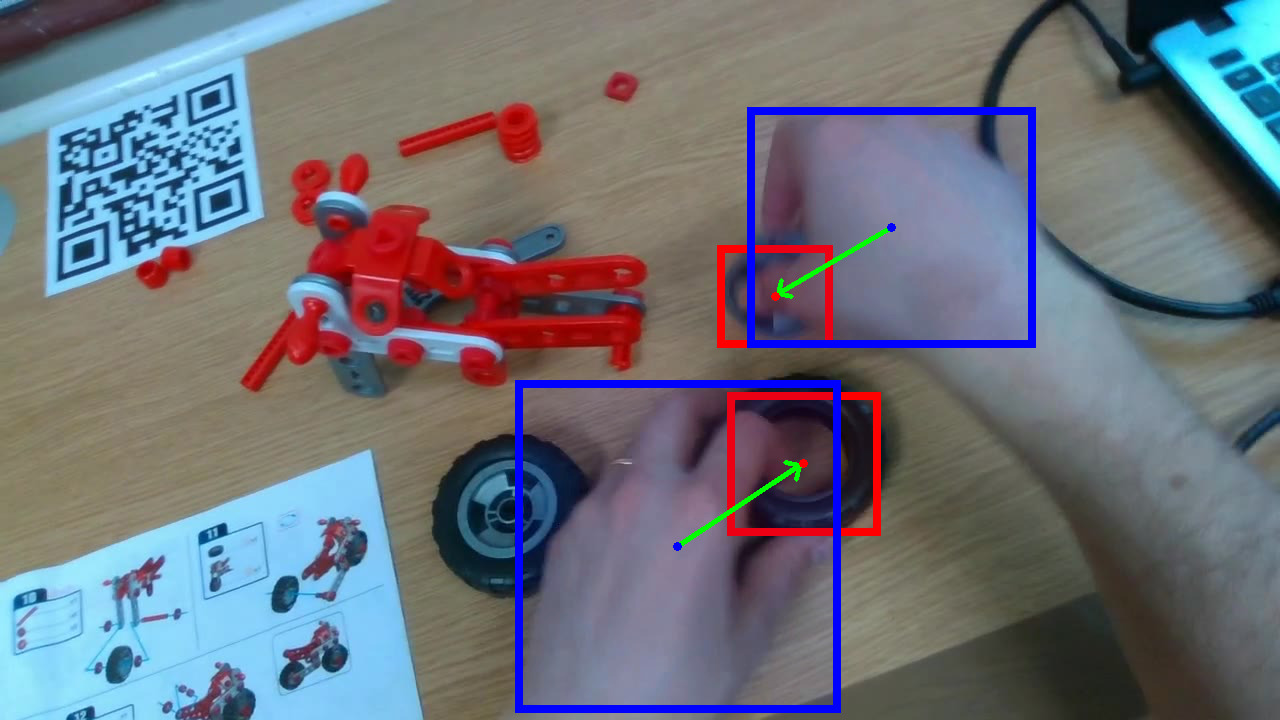}
	\includegraphics[width=0.19\linewidth]{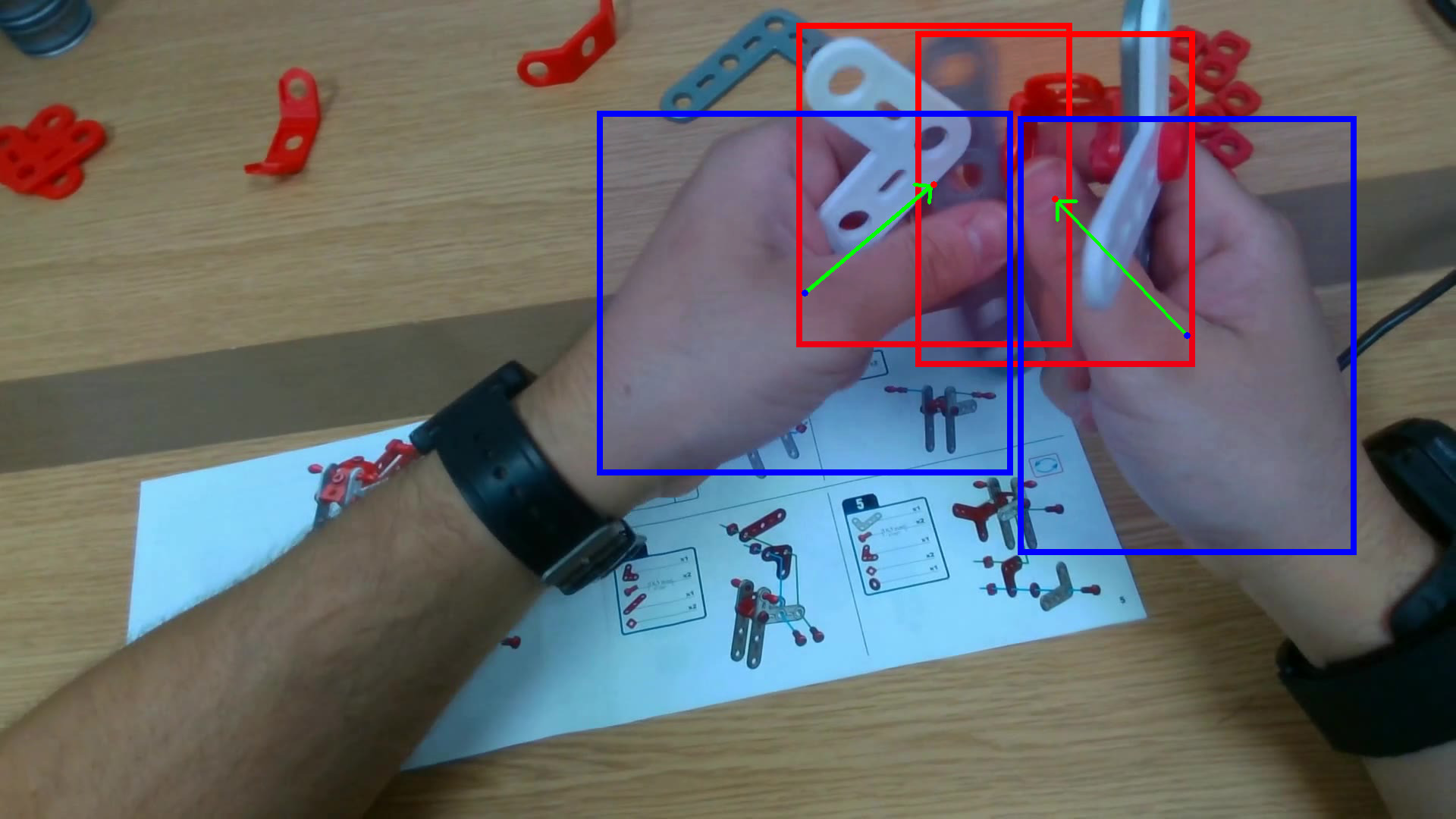}\\
	\rotatebox{90}{Occlusion}\hfill
	\includegraphics[width=0.19\linewidth]{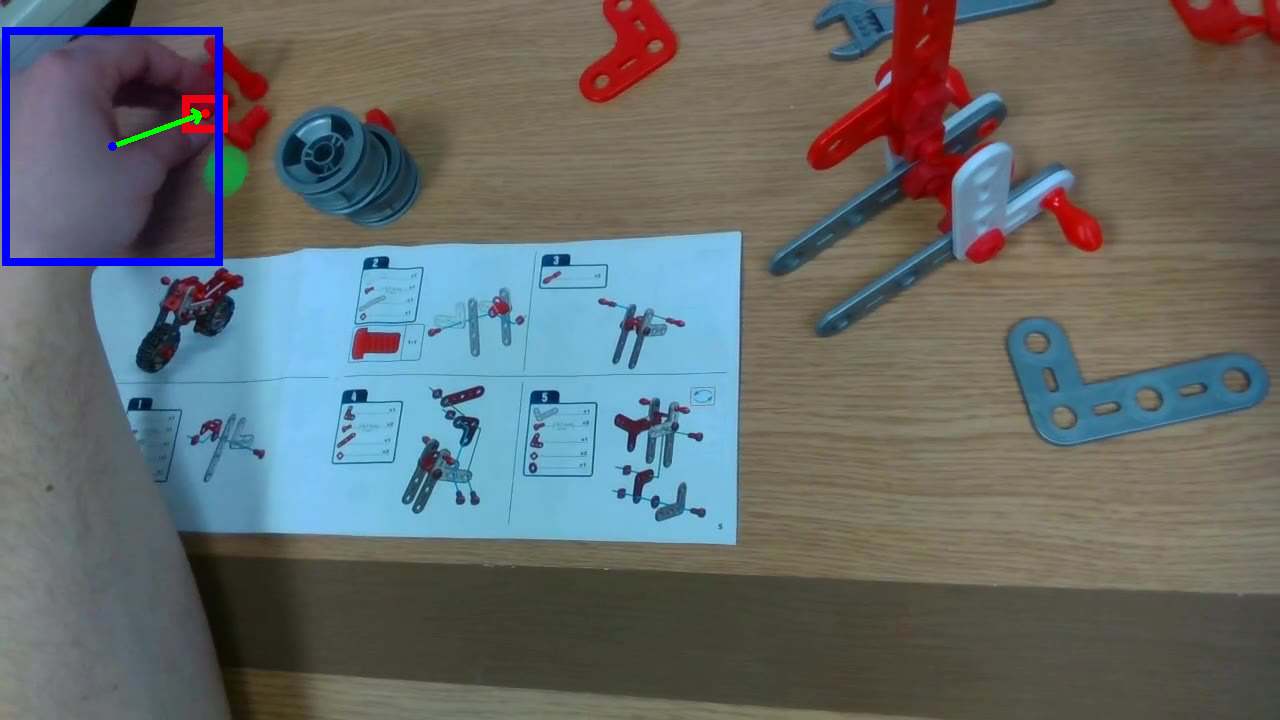}
	\includegraphics[width=0.19\linewidth]{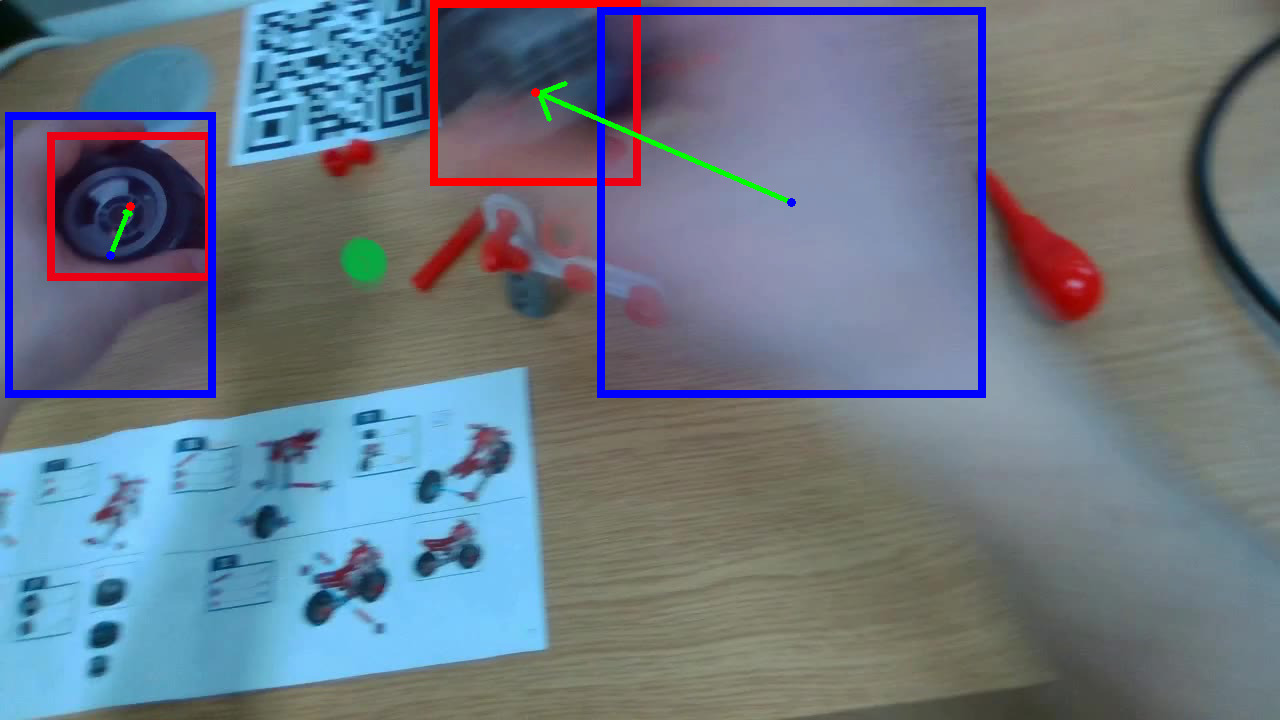}
	\includegraphics[width=0.19\linewidth]{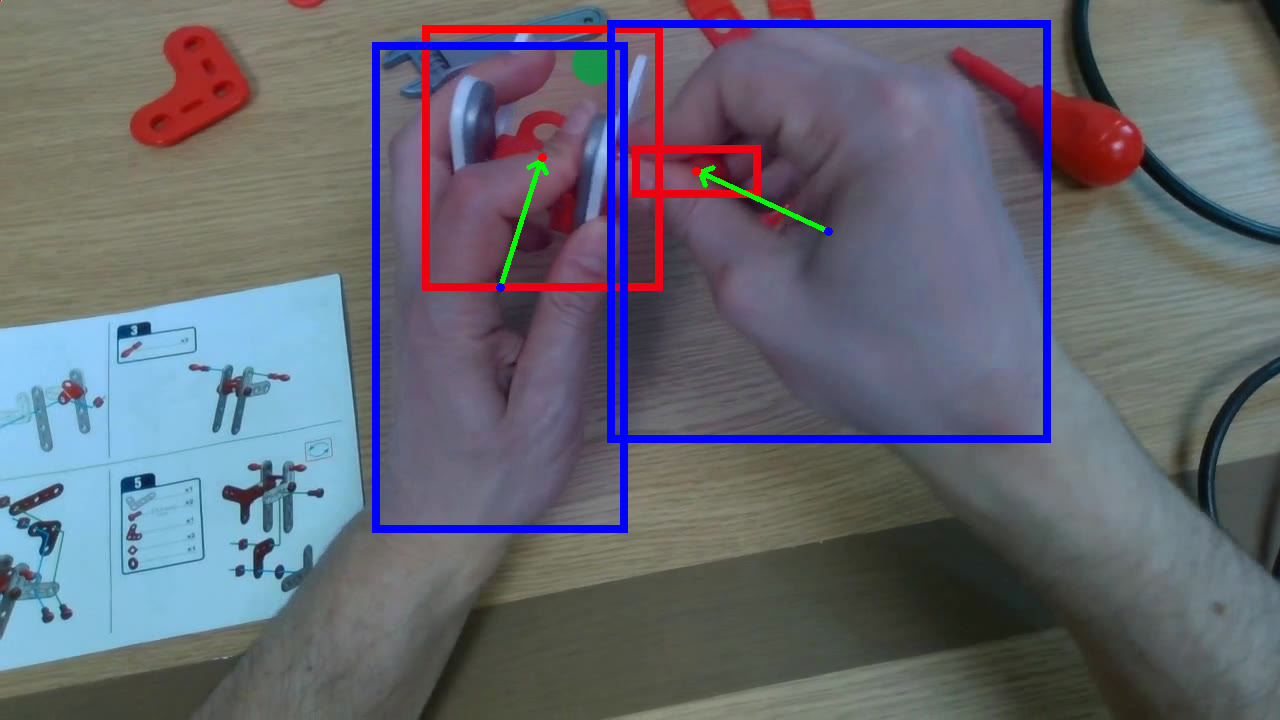}
	\includegraphics[width=0.19\linewidth]{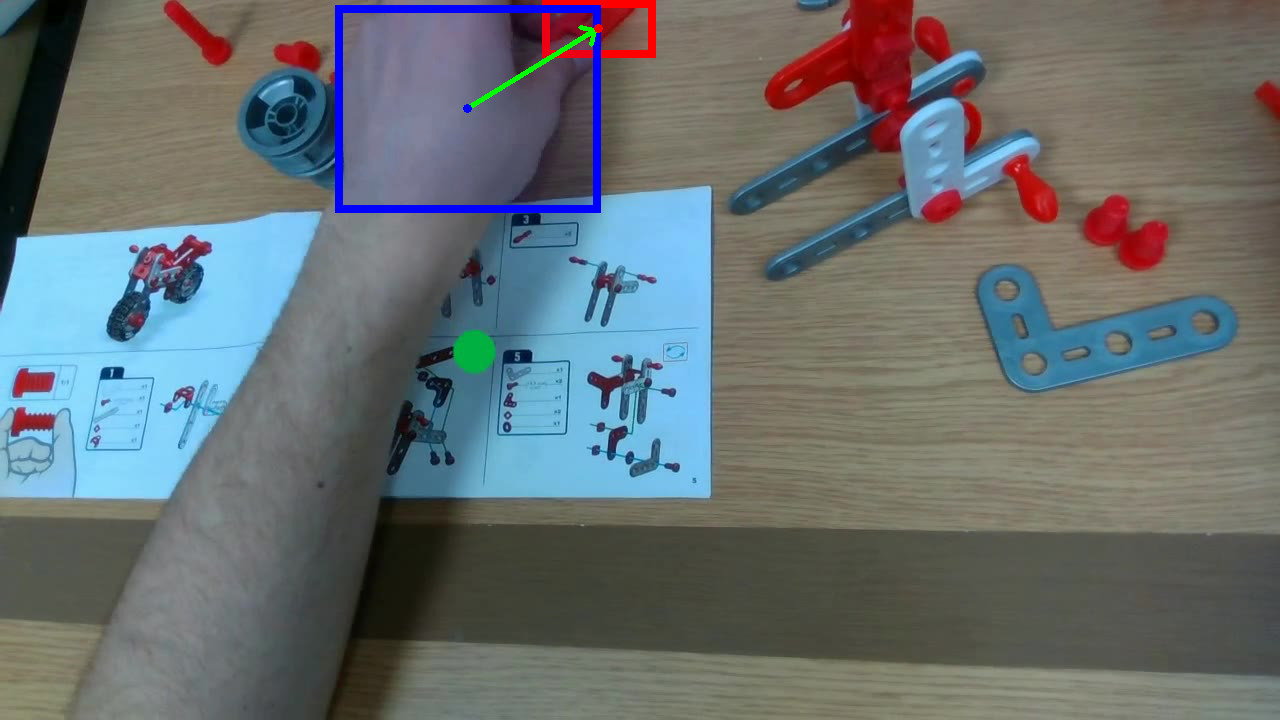}
	\includegraphics[width=0.19\linewidth]{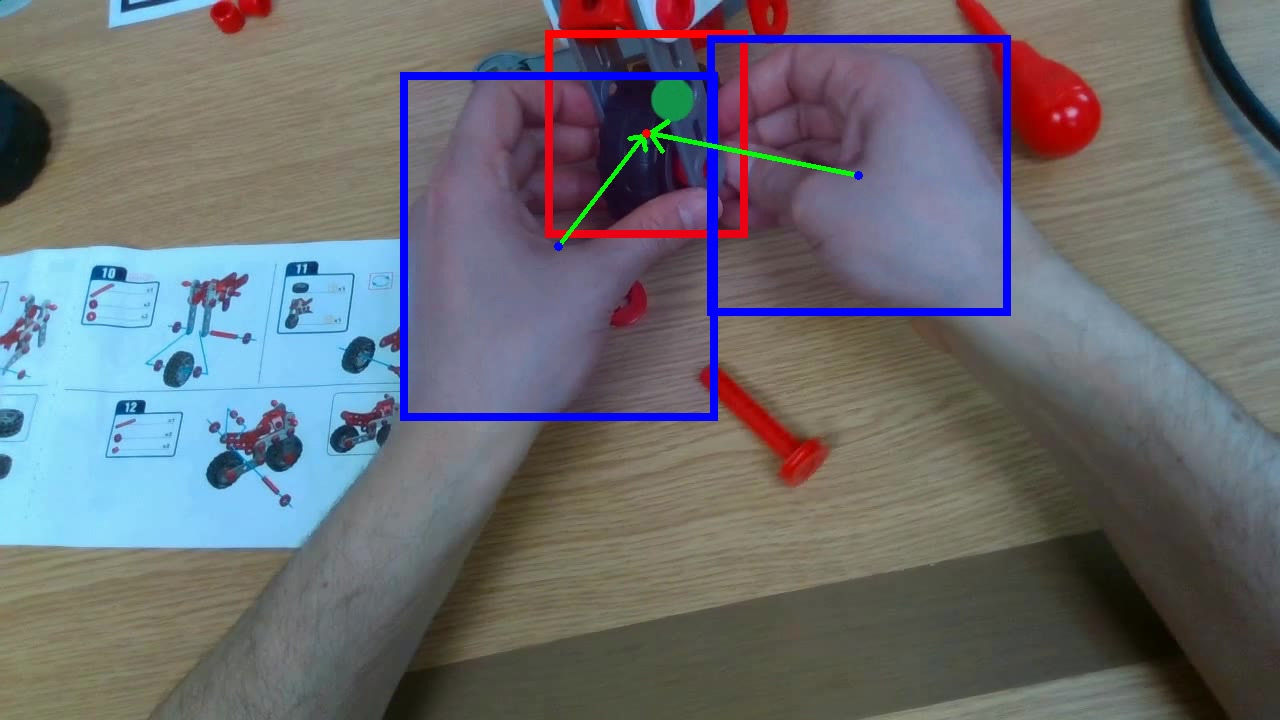}\\
	\rotatebox{90}{Failure}\hfill
	\includegraphics[width=0.19\linewidth]{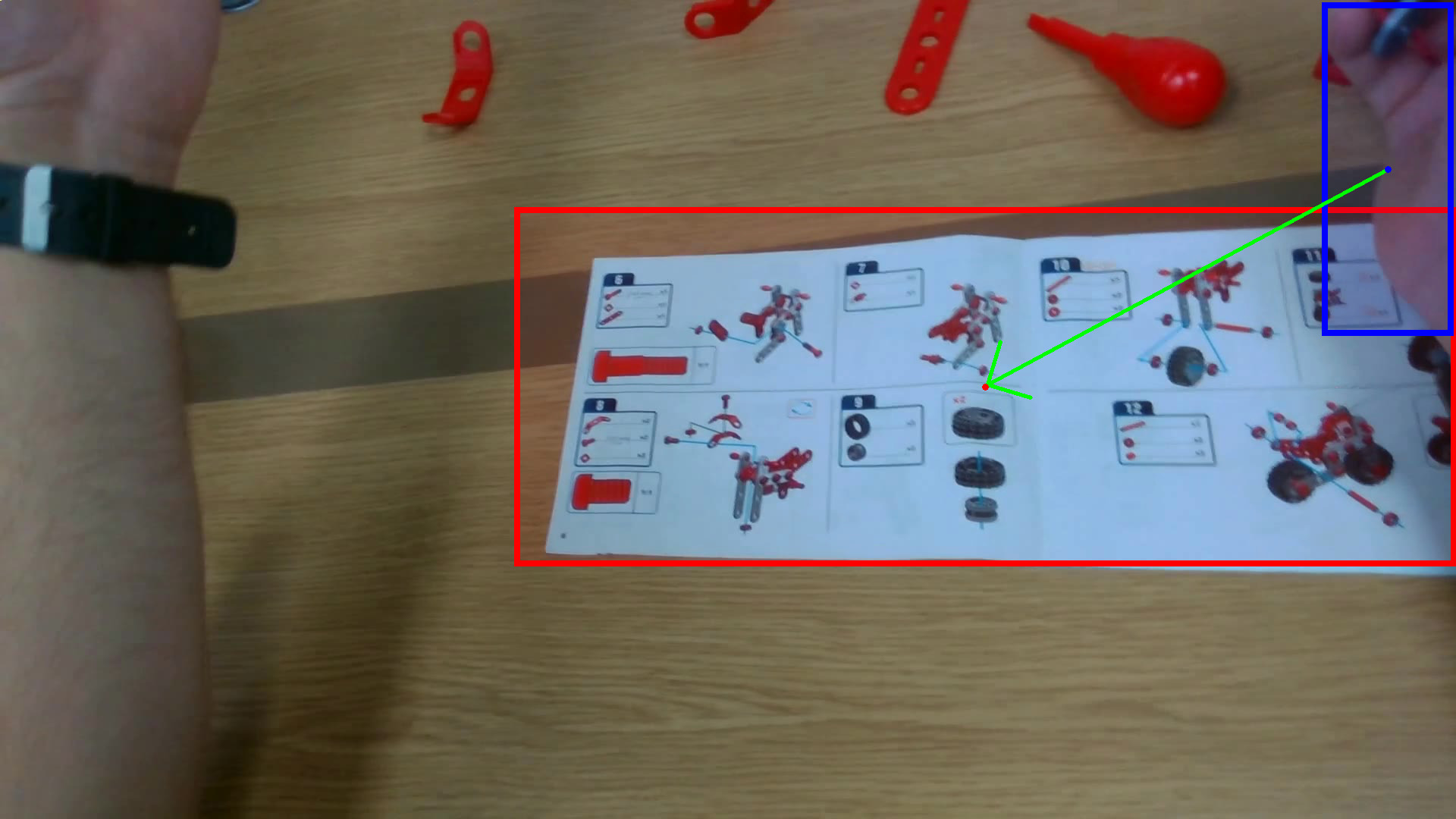}
	\includegraphics[width=0.19\linewidth]{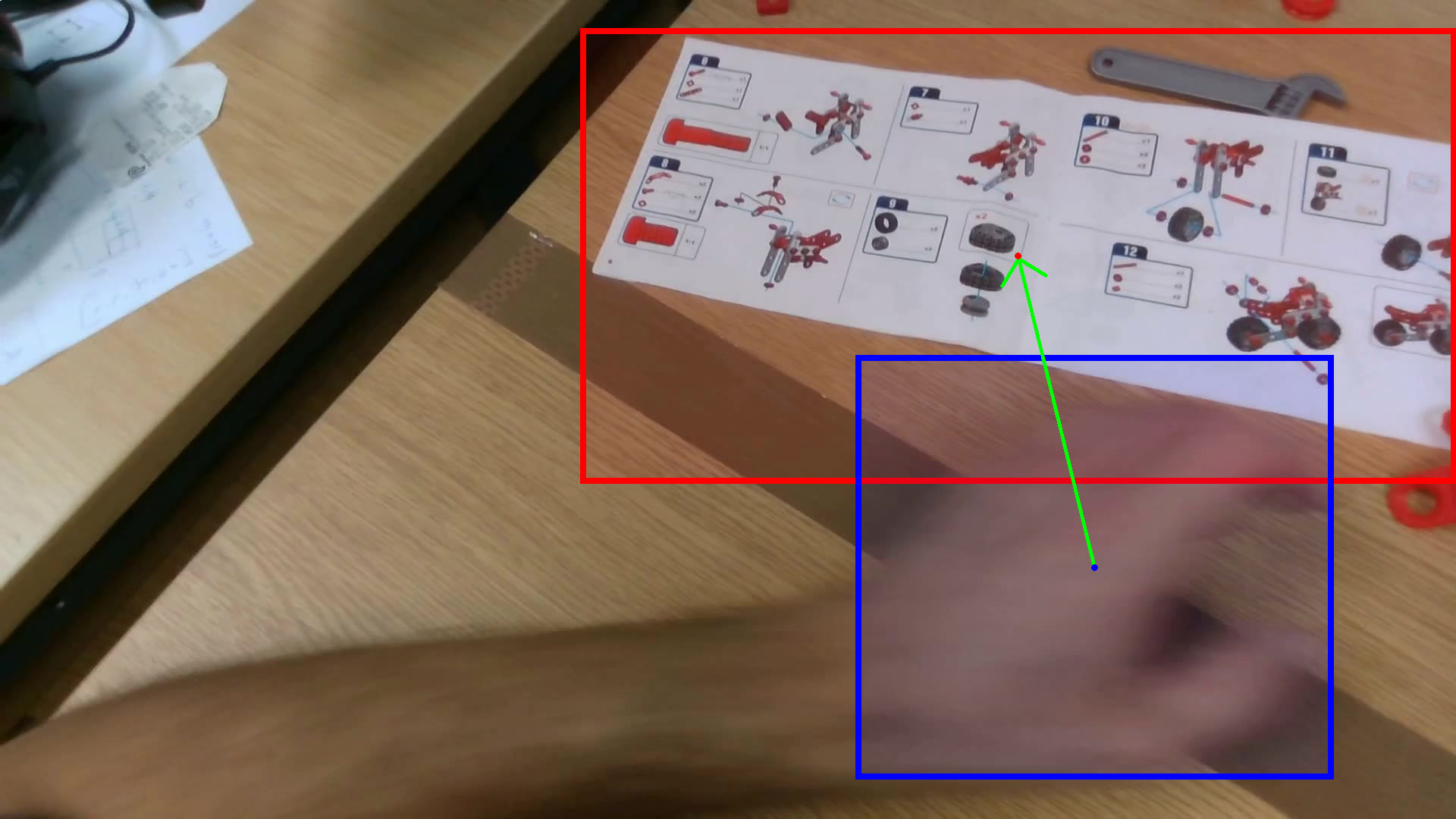}
	\includegraphics[width=0.19\linewidth]{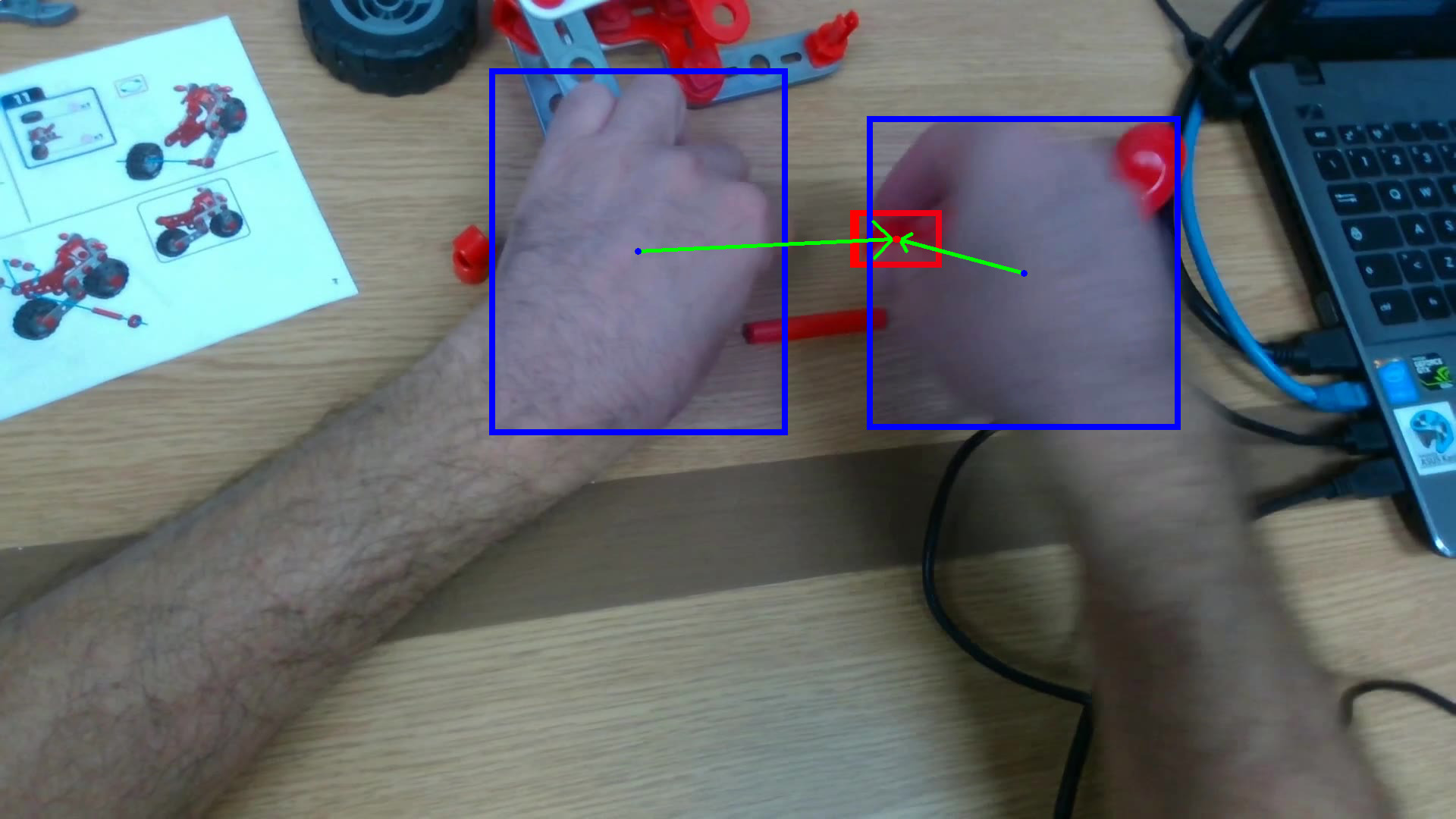}
	\includegraphics[width=0.19\linewidth]{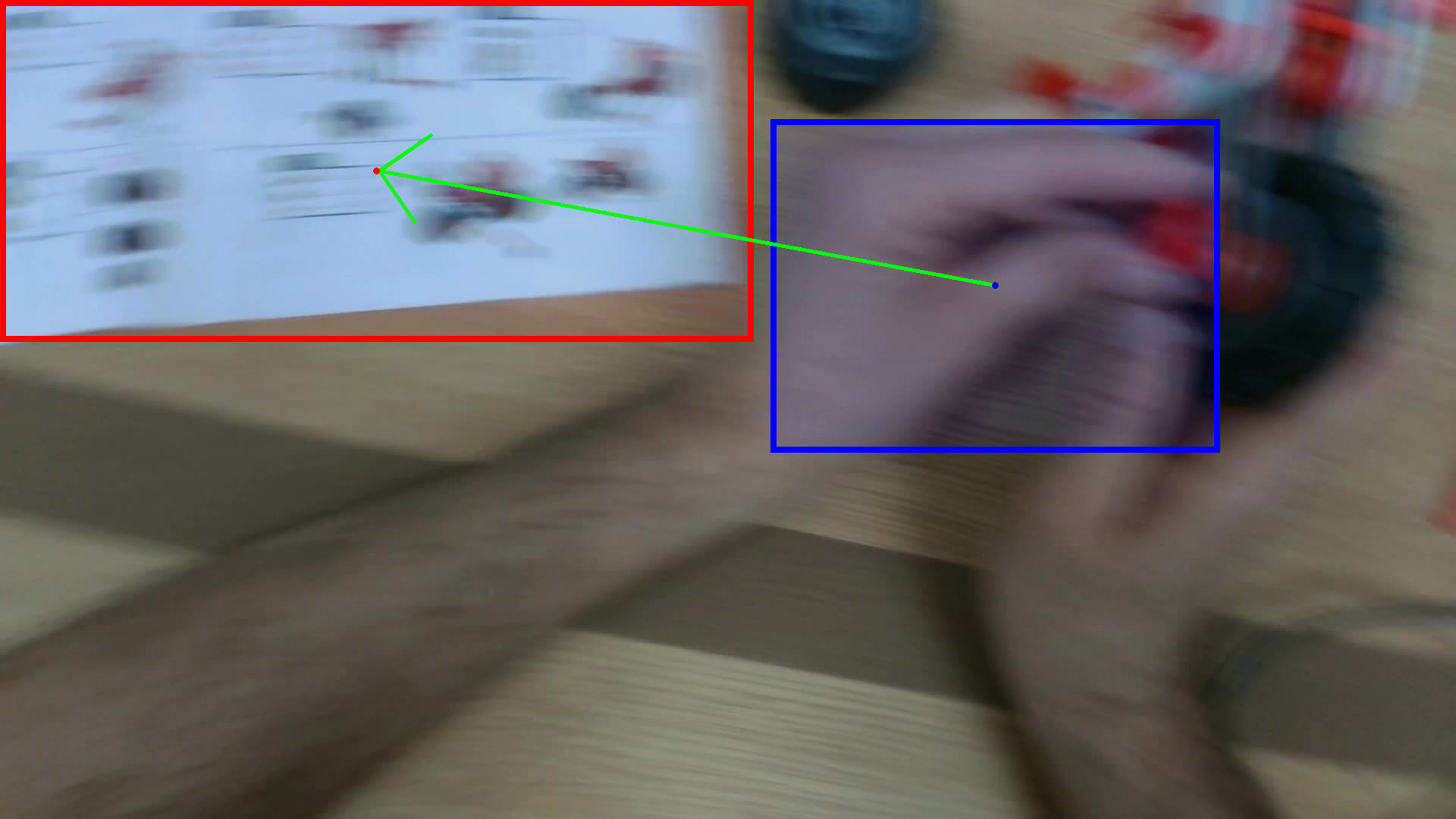}
	\includegraphics[width=0.19\linewidth]{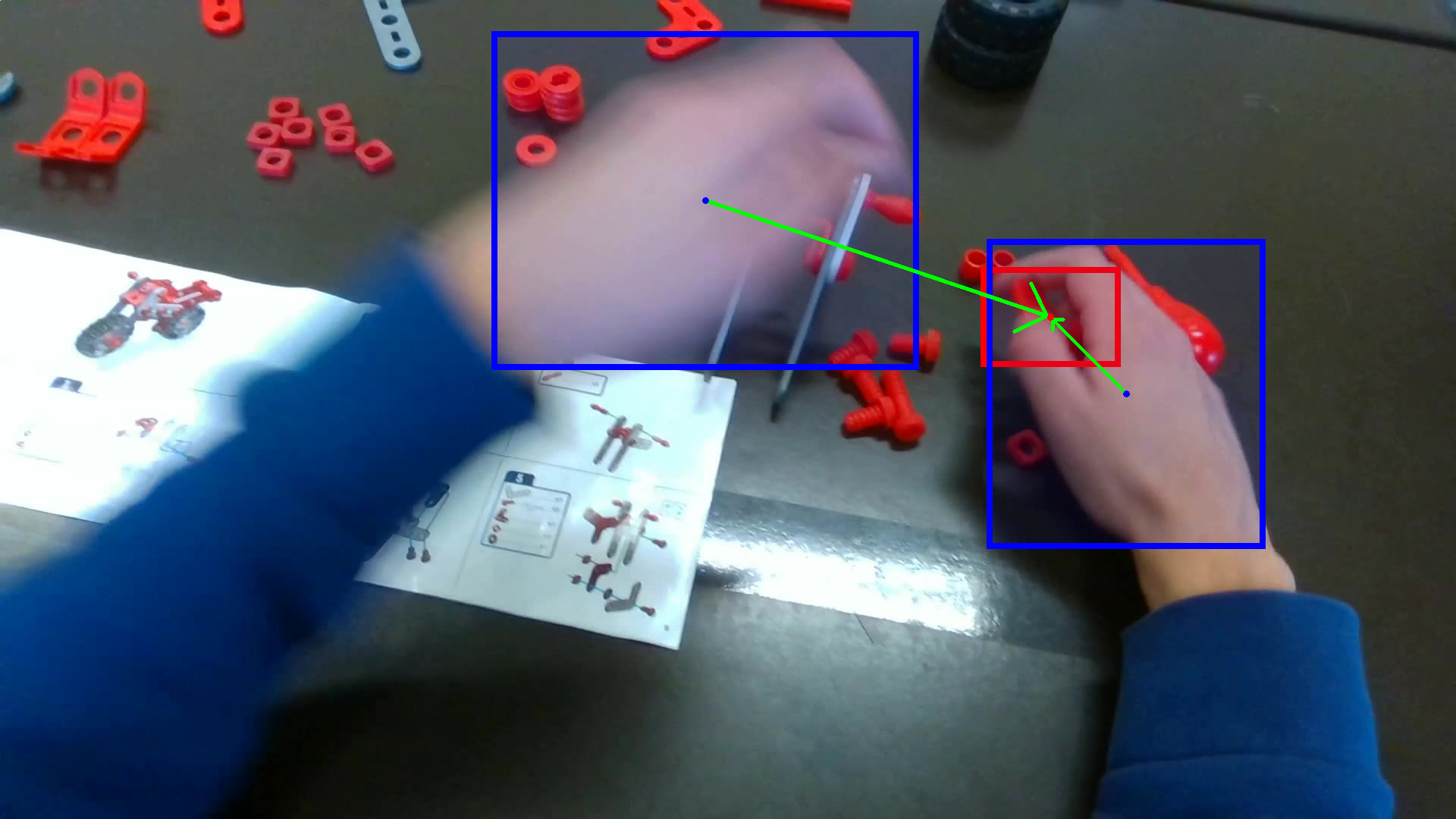}\\
	\centering
	\caption{\label{fig:qualitative_meccano} Qualitative Results on the MECCANO dataset. Each green arrow points from a hand bounding box (blue) to the corresponding active object bounding box (red).}
\end{figure*}

\paragraph{Qualitative Results} The qualitative results on 100DOH dataset \cite{100doh} and MEECANO dataset \cite{meccano} are presented in \cref{fig:qualitative_100doh} and \cref{fig:qualitative_meccano} respectively. Each green arrow points from a hand bounding box (blue) to the corresponding active object bounding box (red). The visualization shows that our method is able to robustly detect the active object under scenes with overlapping objects and severe occlusions. Most failure cases are due to wrong hand detection, motion blur, and insufficient feature from tiny hands and objects.

\begin{figure*}[t]
	\centering
	\includegraphics[width=0.24\linewidth]{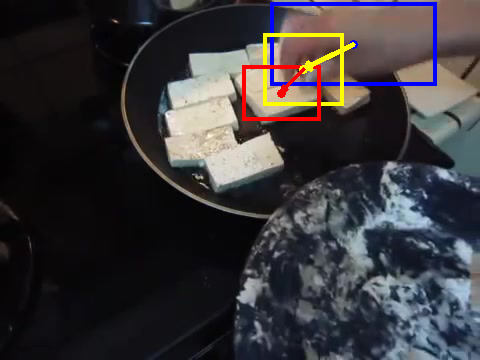}
	\includegraphics[width=0.24\linewidth]{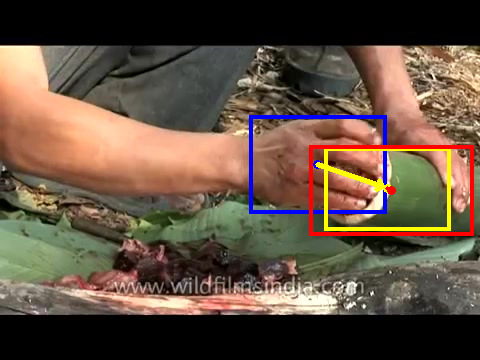}
	\includegraphics[width=0.24\linewidth]{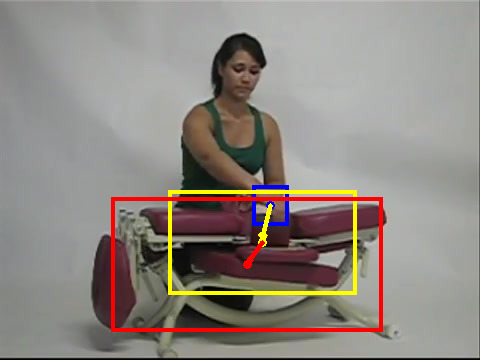}
	\includegraphics[width=0.24\linewidth]{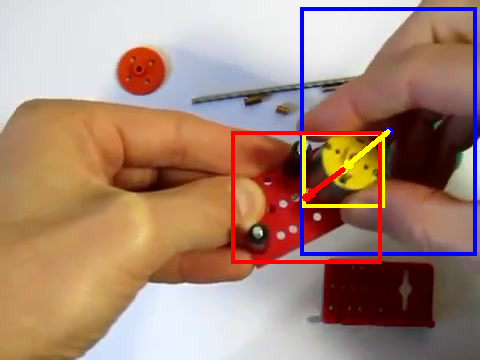}\\
	\includegraphics[width=0.24\linewidth]{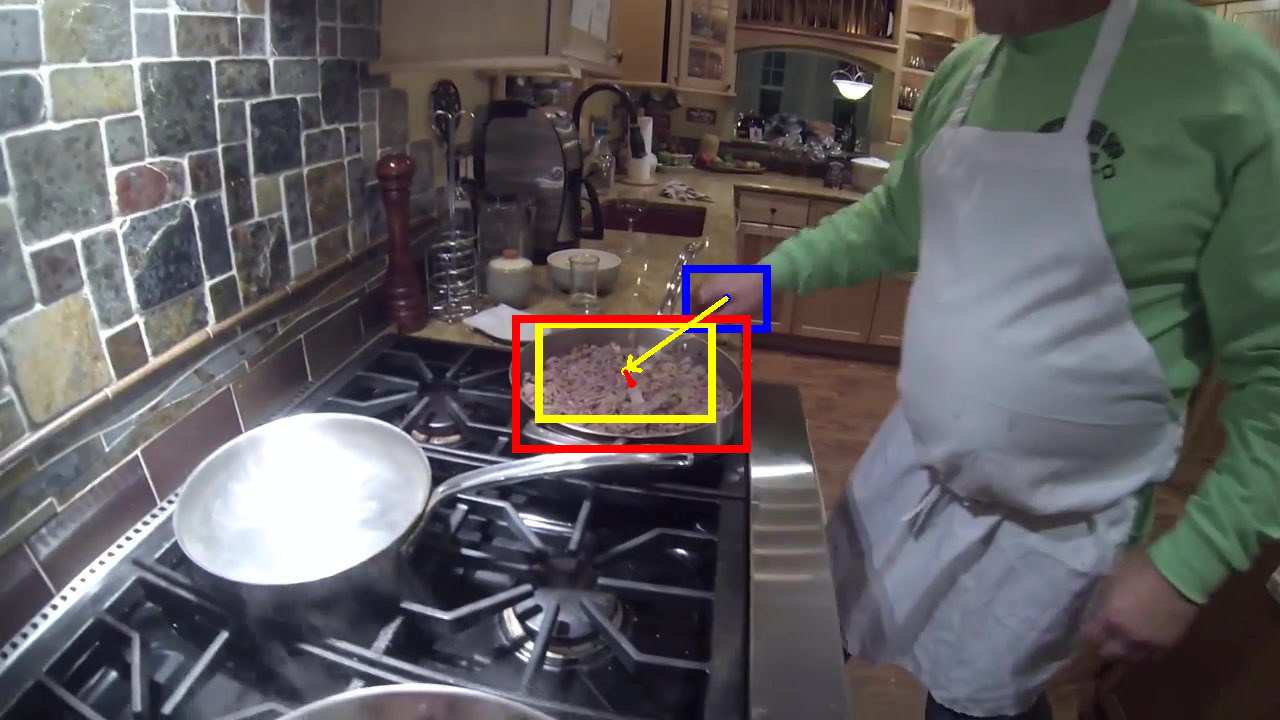}
	\includegraphics[width=0.24\linewidth]{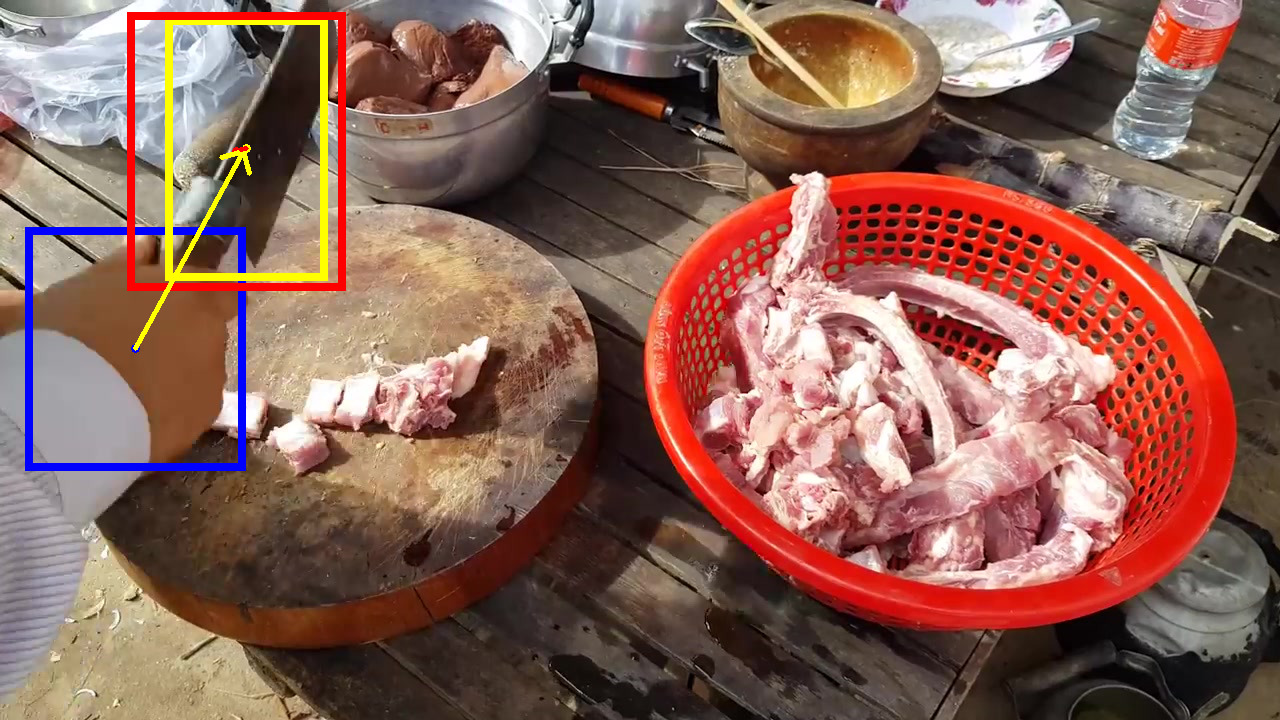}
	\includegraphics[width=0.24\linewidth]{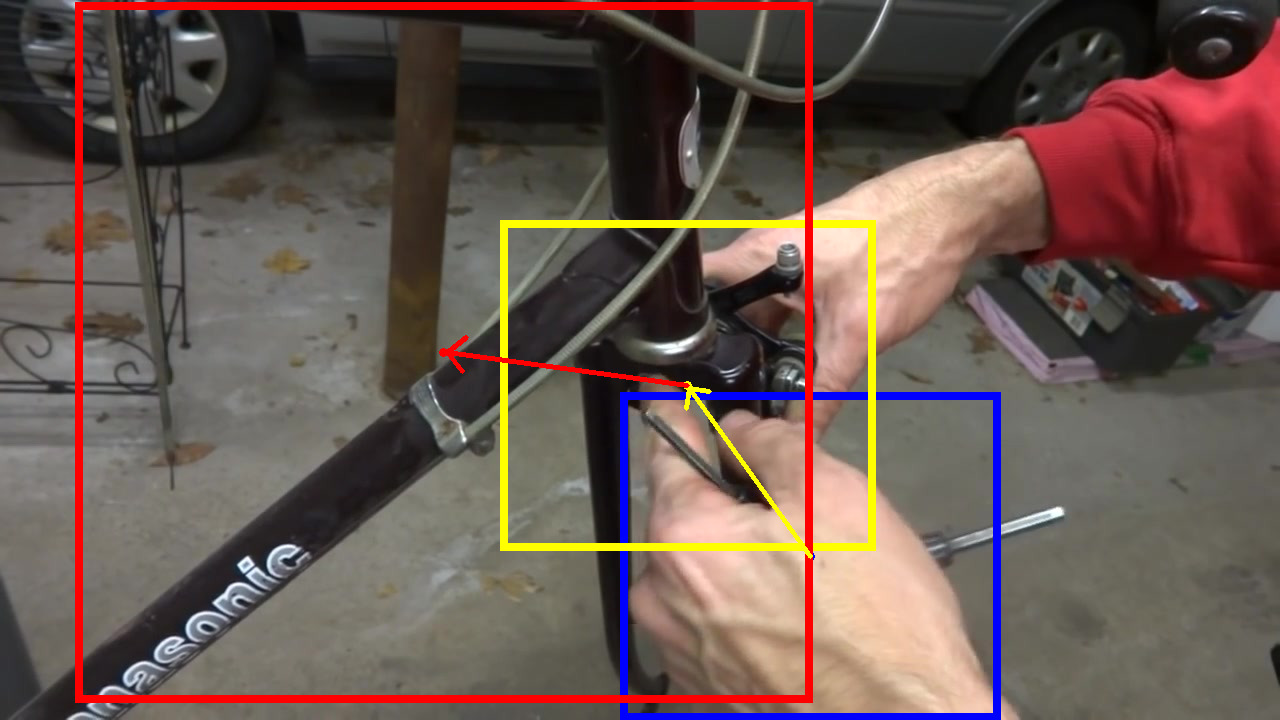}
	\includegraphics[width=0.24\linewidth]{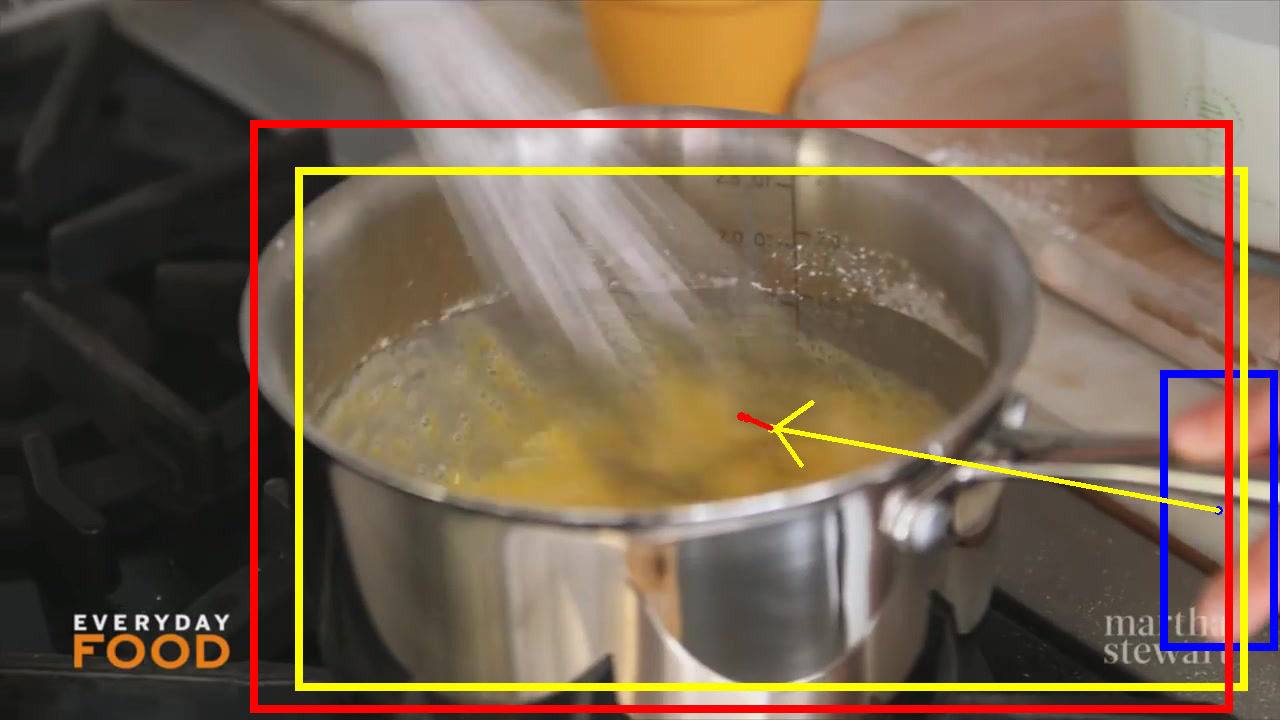}\\
	\centering
	\caption{\label{fig:iterative_100doh} Visualization of iterative refinement on the 100DOH dataset. We show the initial active object hypothesis (yellow bounding box) and the refined active object estimation (red bounding box) corresponding to the hand (blue bounding box).}
\end{figure*}

\begin{figure*}[t]
	\centering
	\includegraphics[width=0.24\linewidth]{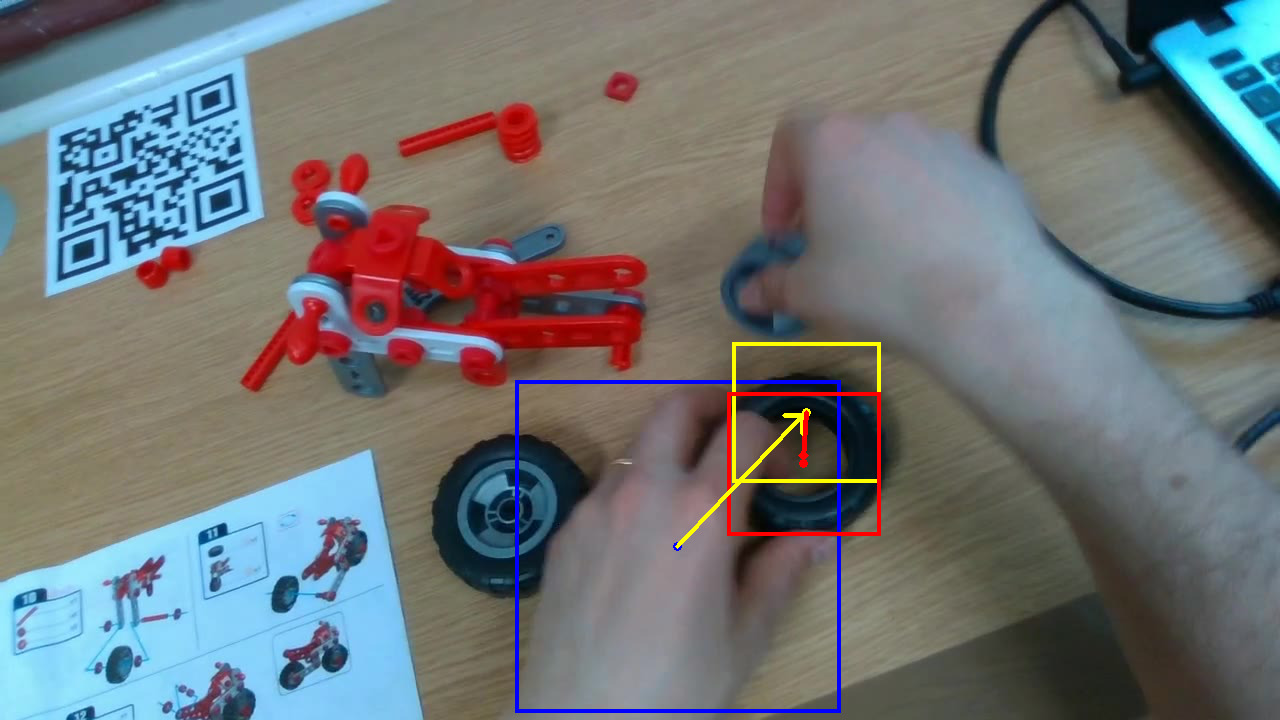}
	\includegraphics[width=0.24\linewidth]{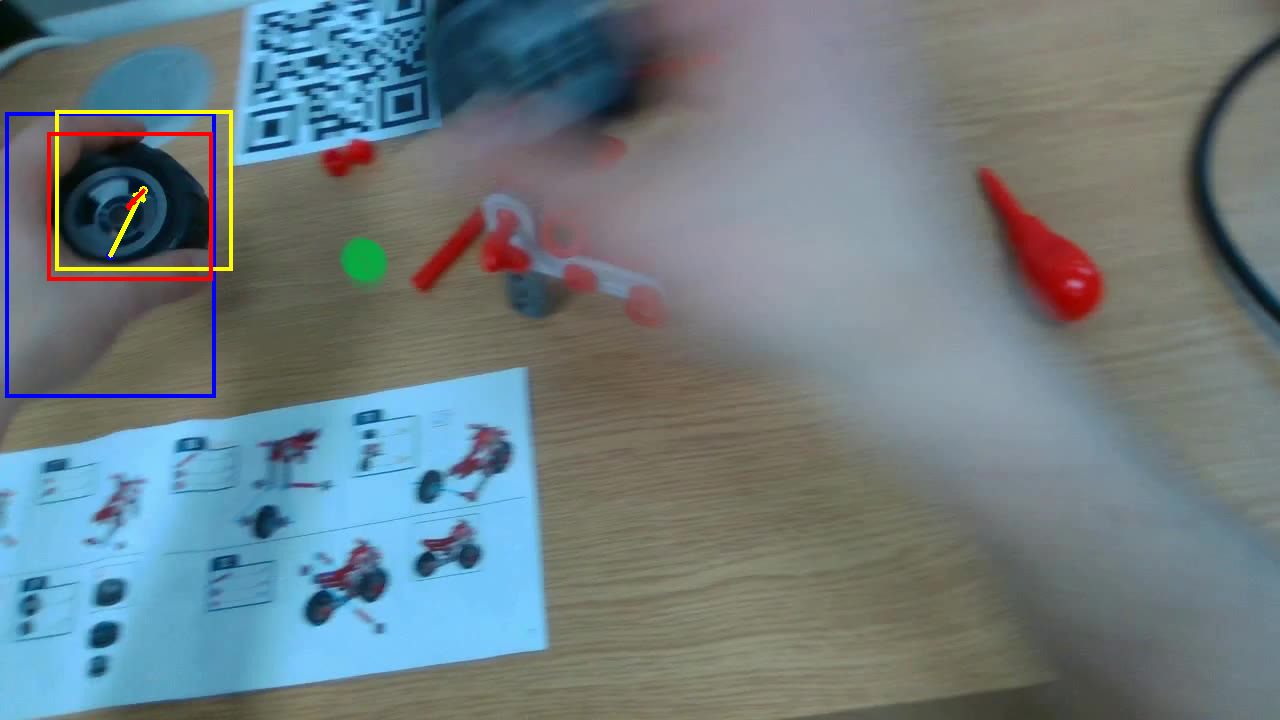}
	\includegraphics[width=0.24\linewidth]{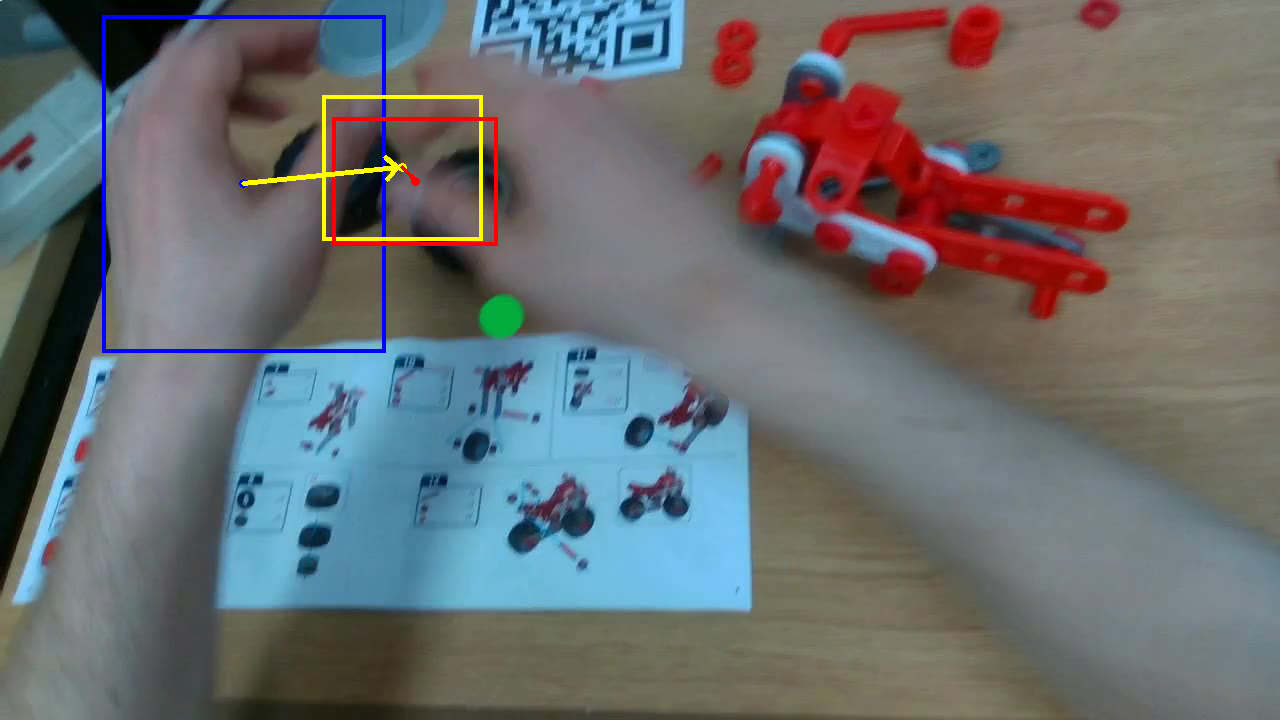}
	\includegraphics[width=0.24\linewidth]{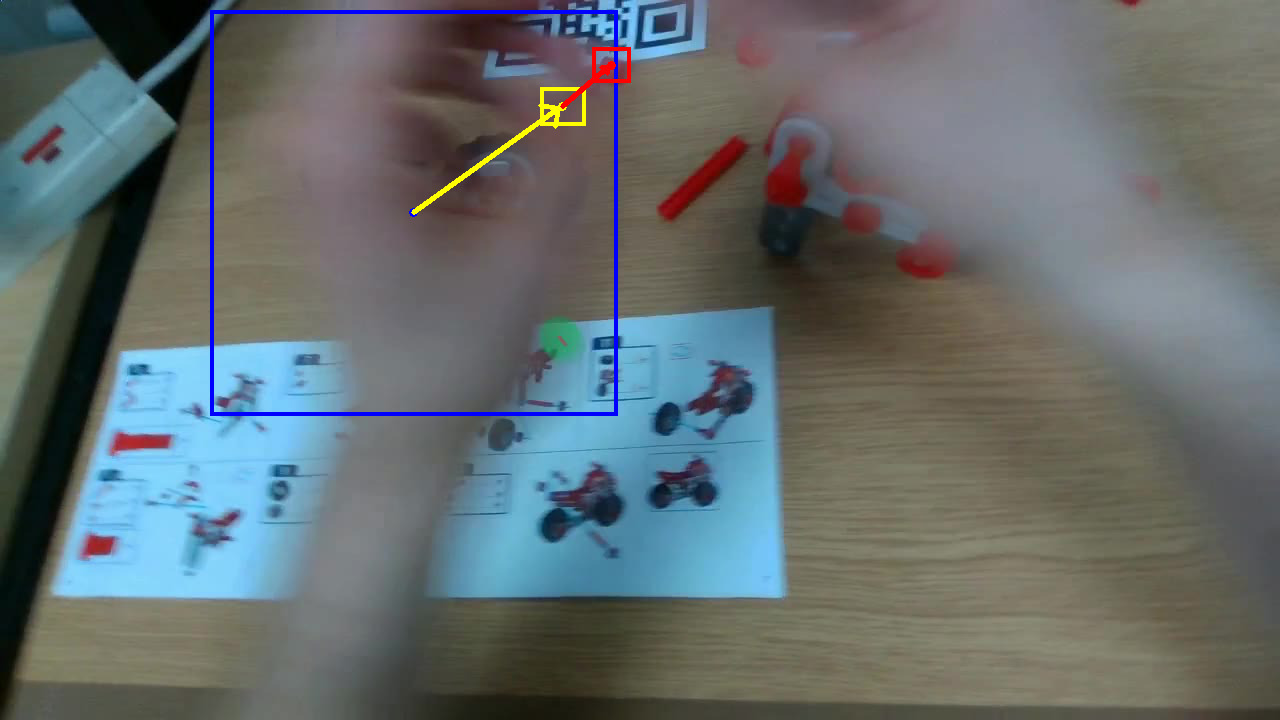}\\
	\includegraphics[width=0.24\linewidth]{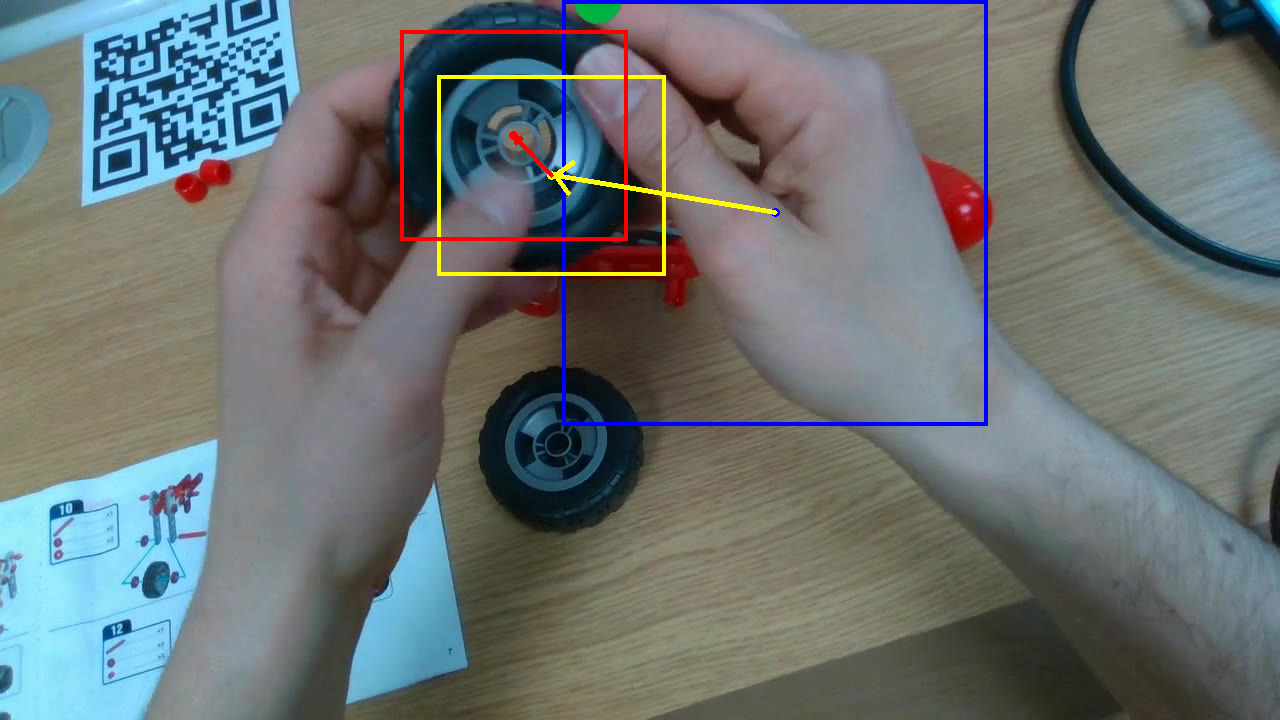}
	\includegraphics[width=0.24\linewidth]{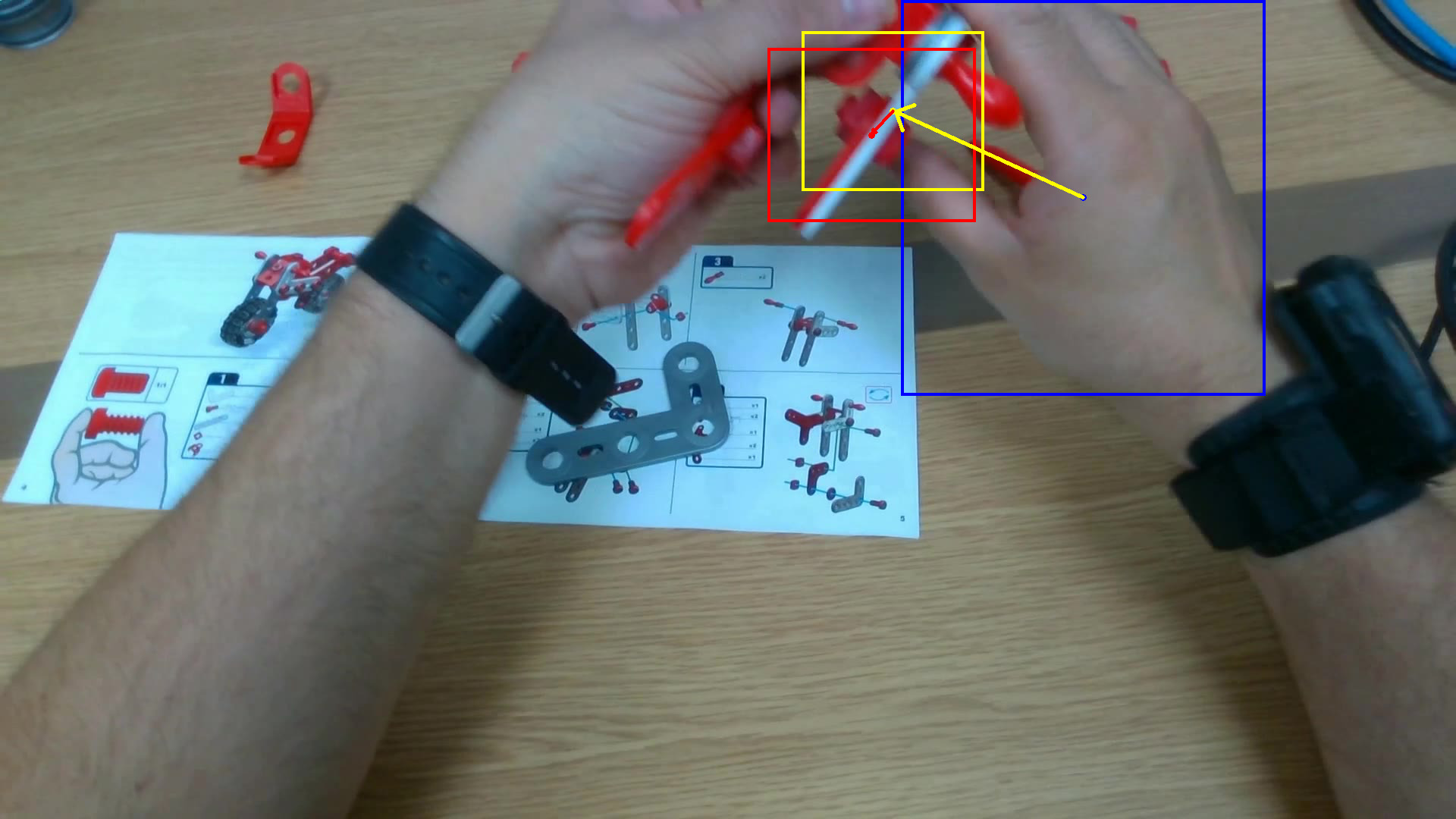}
	\includegraphics[width=0.24\linewidth]{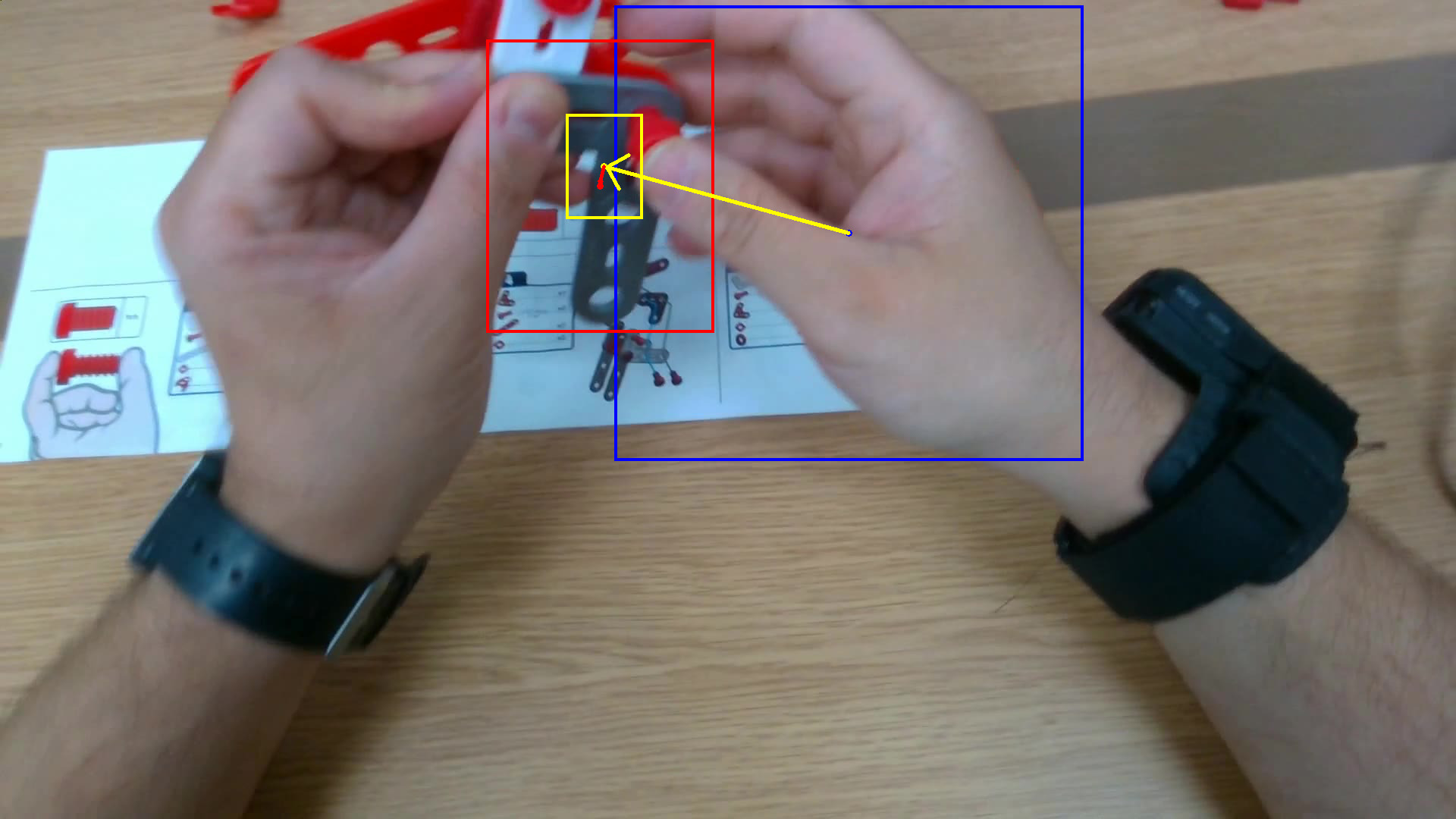}
	\includegraphics[width=0.24\linewidth]{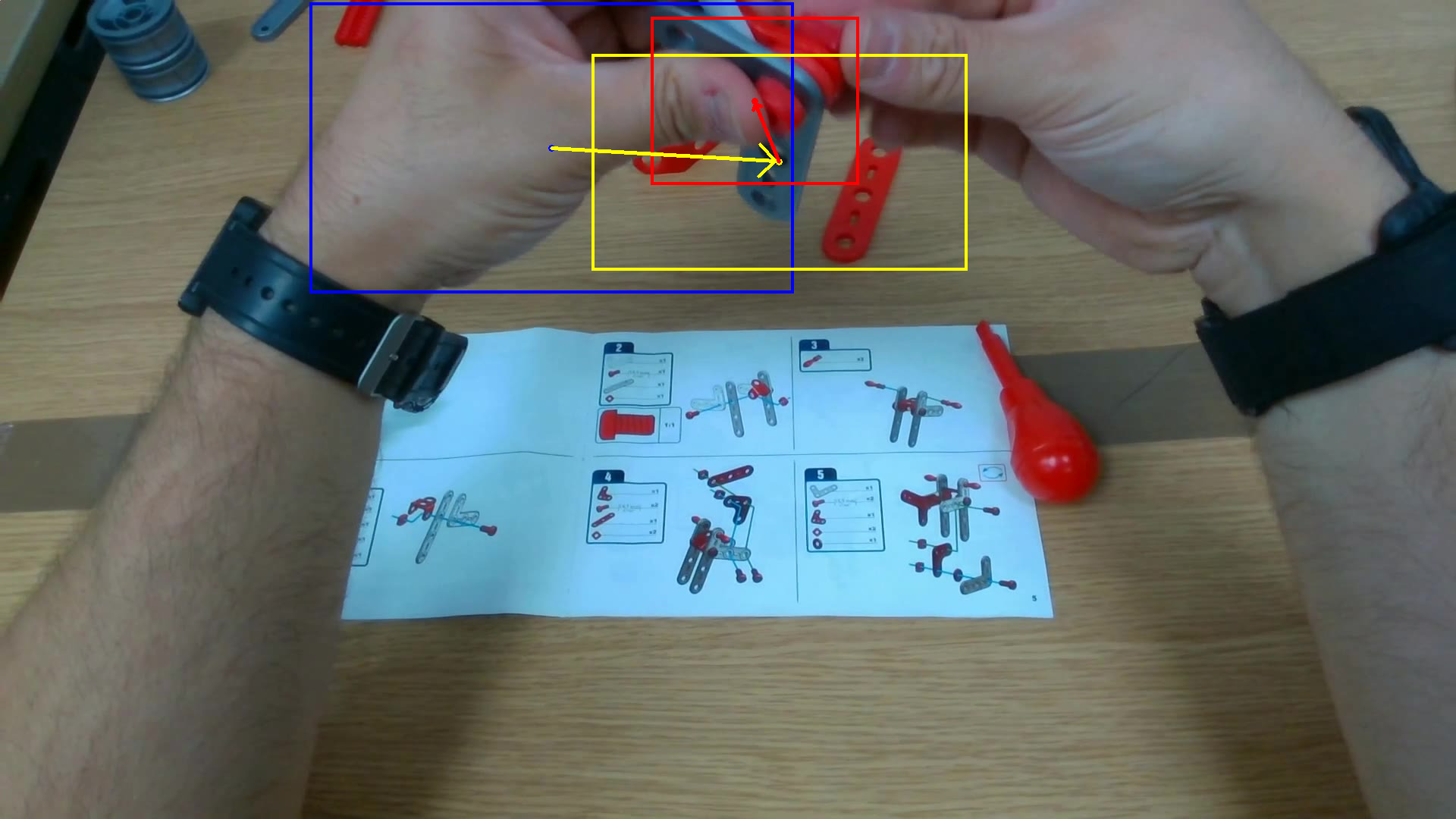}\\
	\centering
	\caption{\label{fig:iterative_meccano} Visualization of iterative refinement on the MECCANO dataset. We show the initial active object hypothesis (yellow bounding box) and the refined active object estimation (red bounding box) corresponding to the hand (blue bounding box).}
\end{figure*}

\paragraph{Visualization of Iterative Refinement} We further visualize the effect of iterative refinement. In this visualization, we show the initial active object hypothesis (yellow bounding box) and the refined active object estimation (red bounding box) on 100DOH dataset (in \cref{fig:iterative_100doh}) and MEECANO dataset (in \cref{fig:iterative_meccano}). All the examples show that the iterative refinement by applying the voting function multiple times could improve the active object bounding box estimation. For better visibility, every sample only shows one pair of hands and objects.

\begin{figure*}[t]
	\centering
	\includegraphics[width=0.19\linewidth]{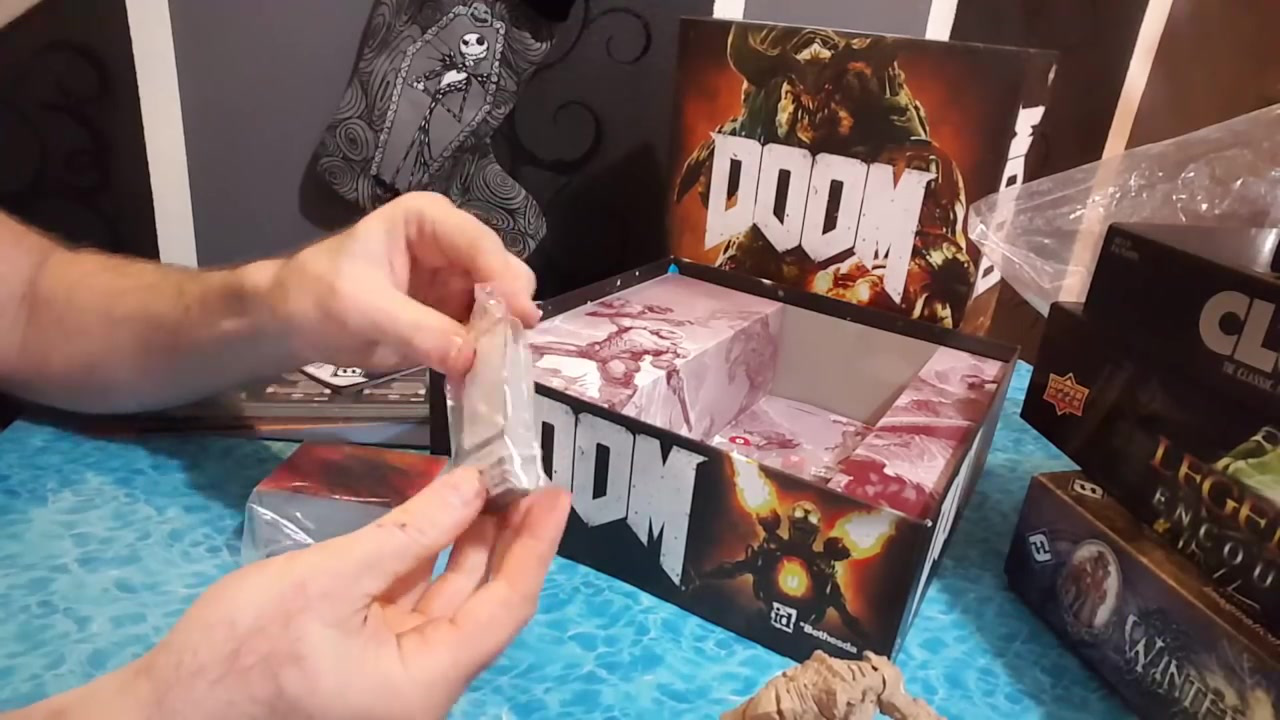}
	\includegraphics[width=0.19\linewidth]{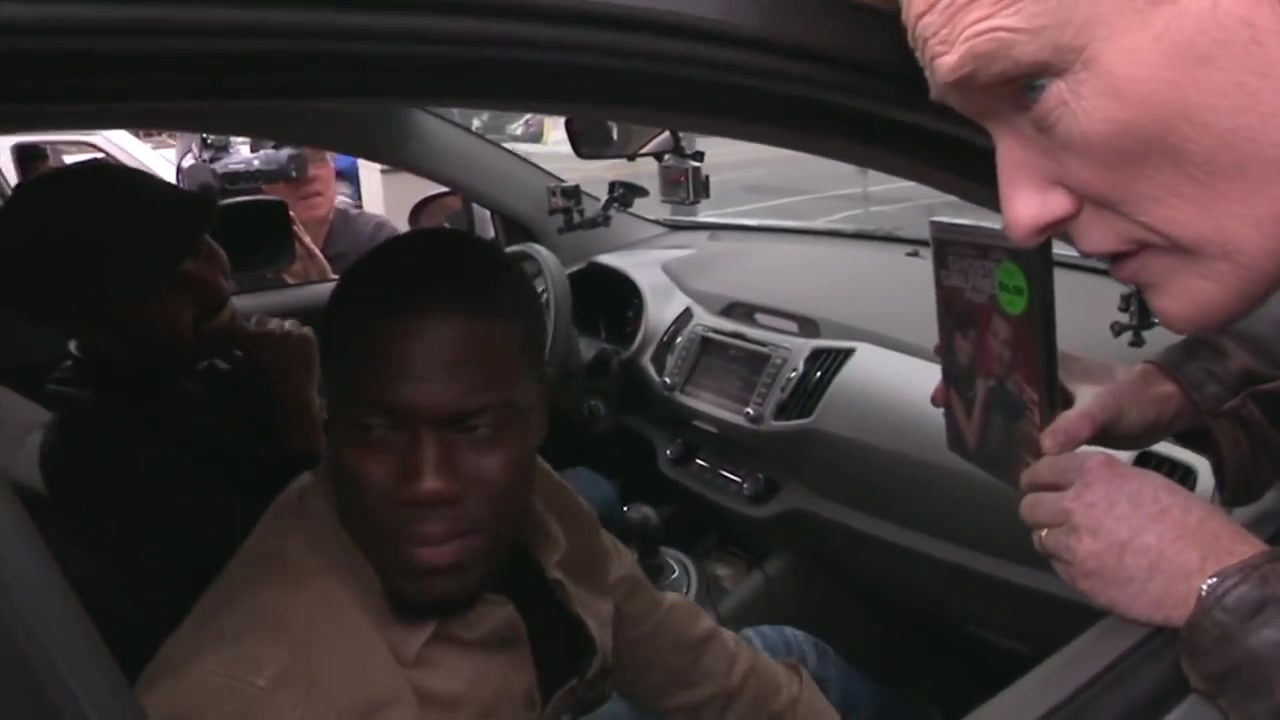}
	\includegraphics[width=0.19\linewidth]{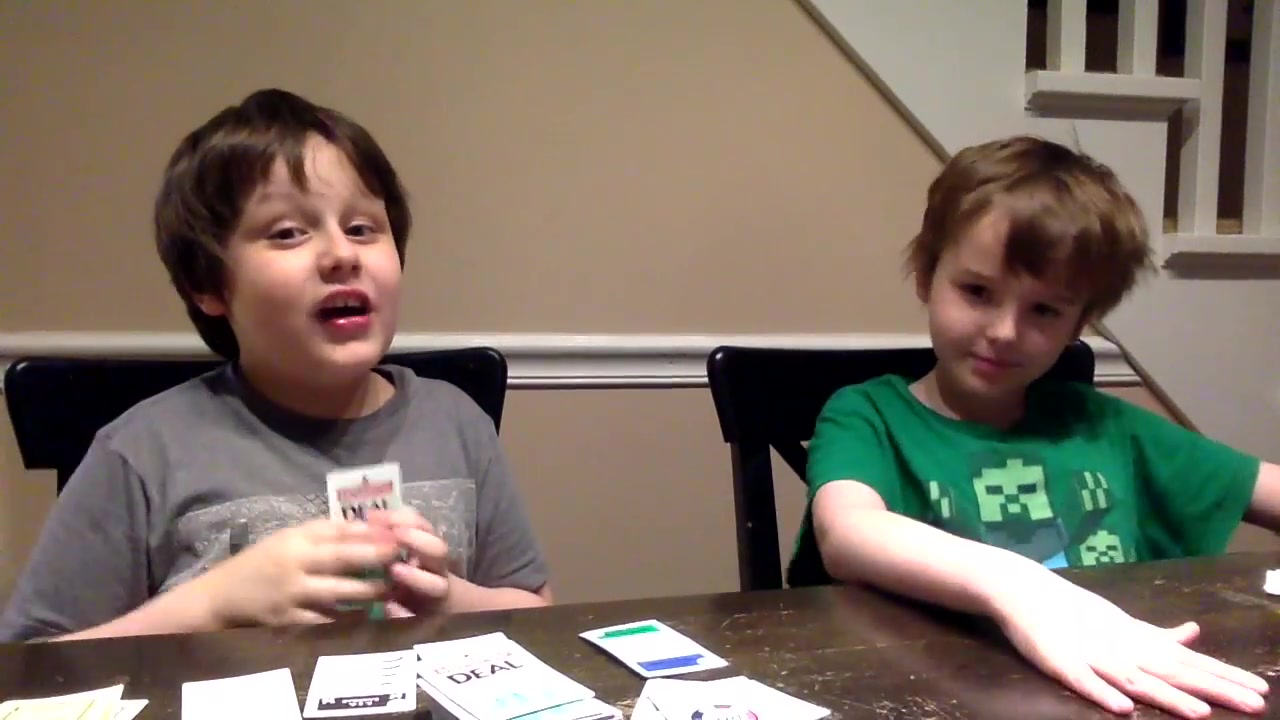}
	\includegraphics[width=0.19\linewidth]{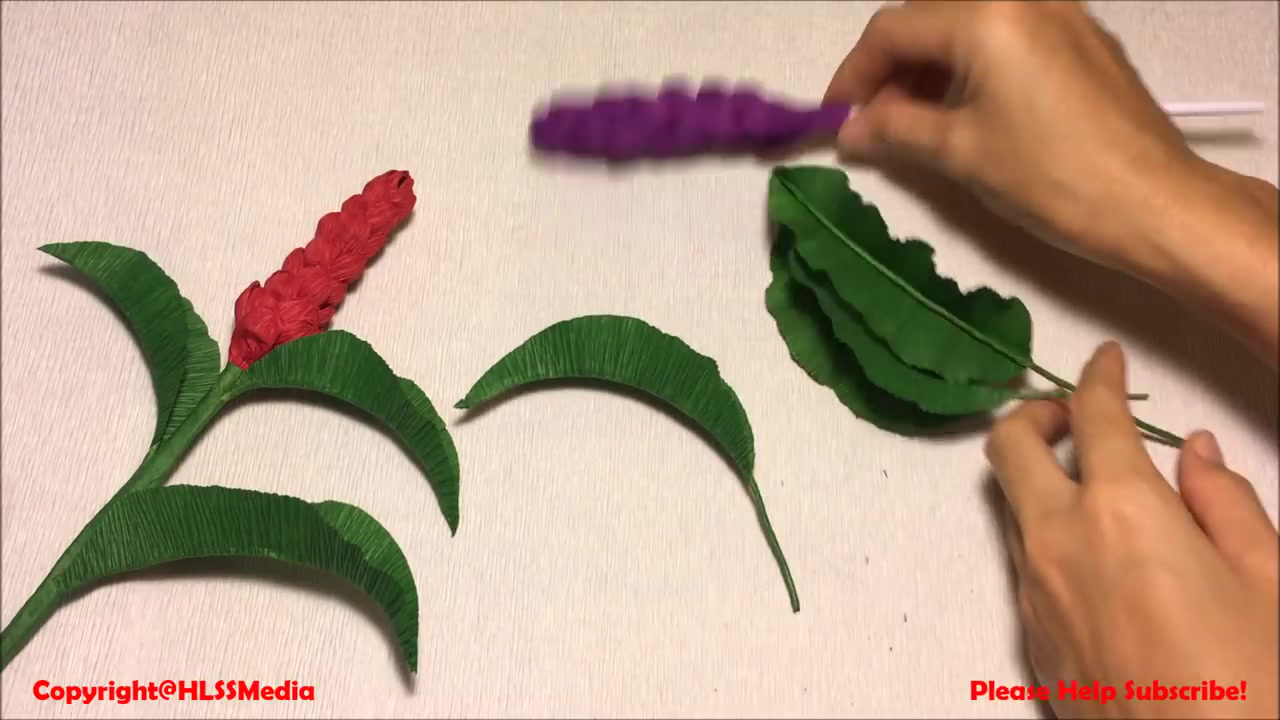}
	\includegraphics[width=0.19\linewidth]{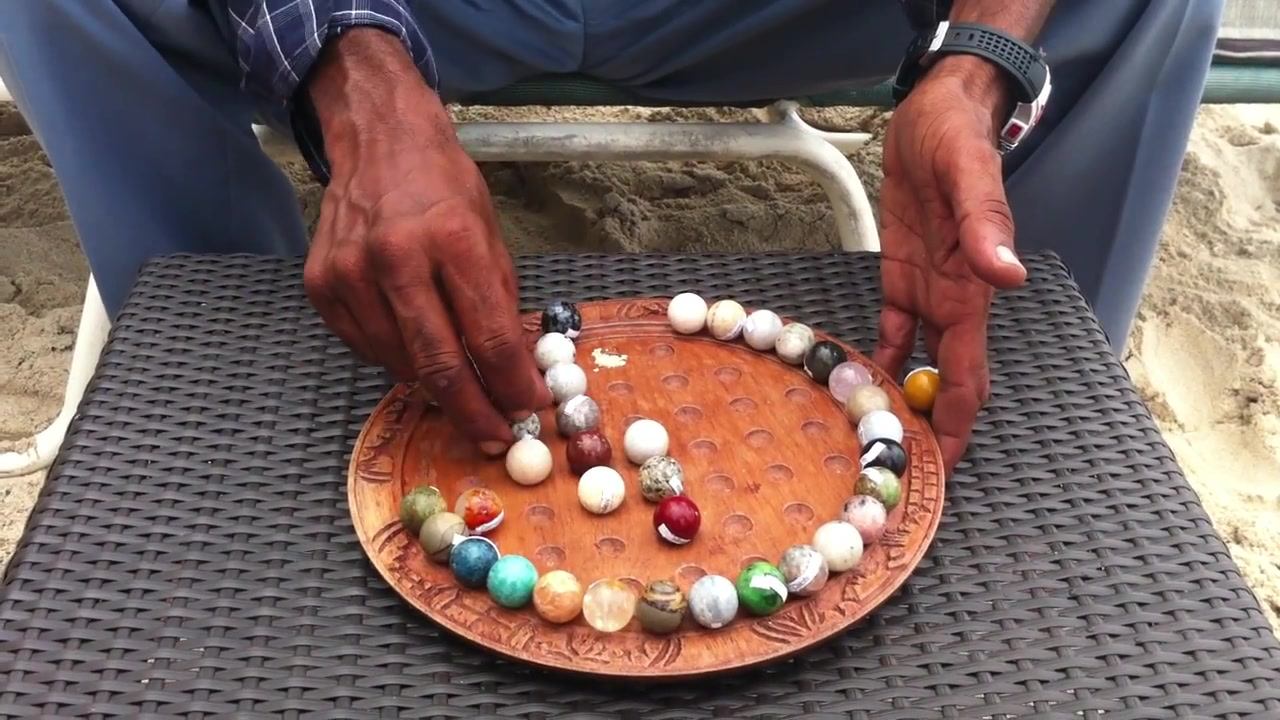}
	\\
	\includegraphics[width=0.19\linewidth]{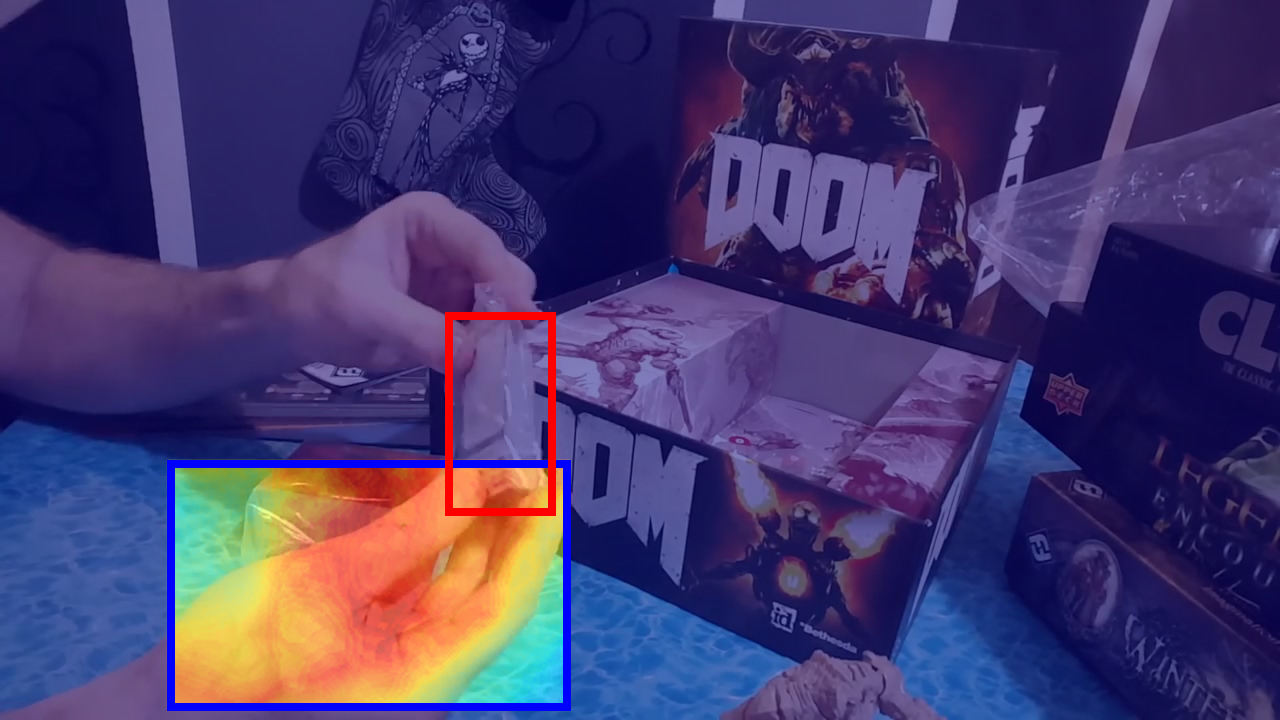}
	\includegraphics[width=0.19\linewidth]{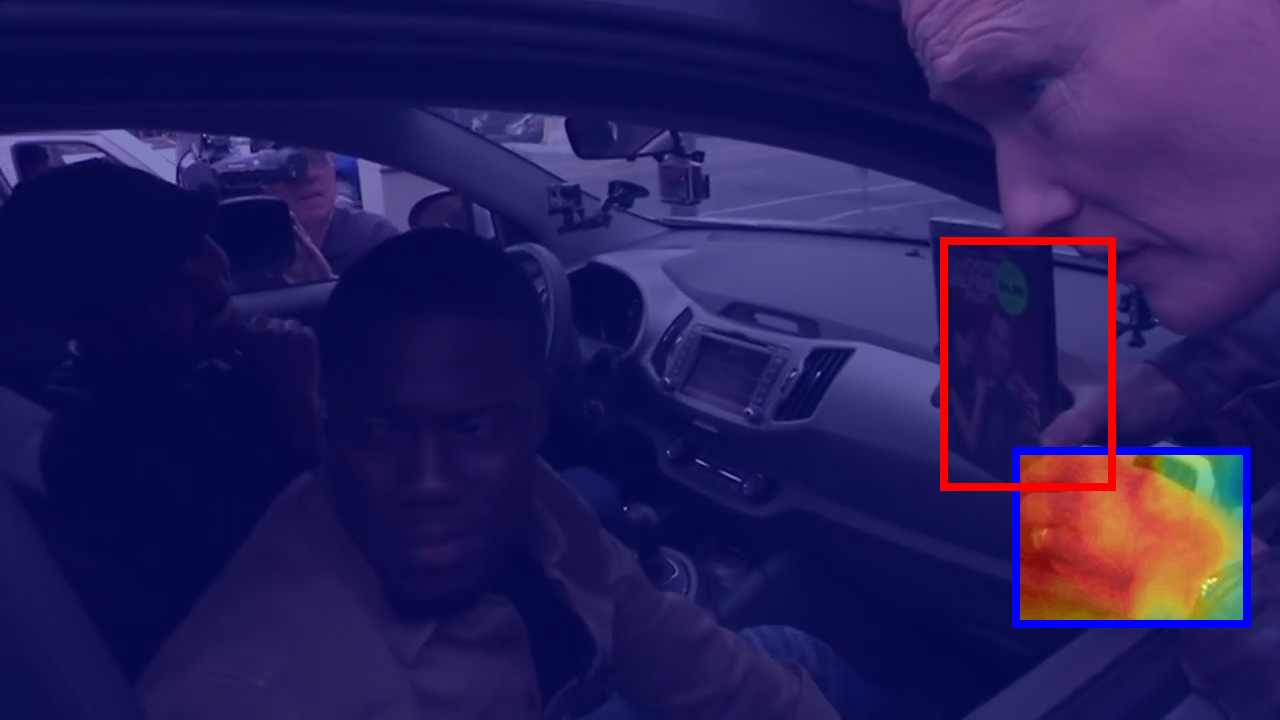}
	\includegraphics[width=0.19\linewidth]{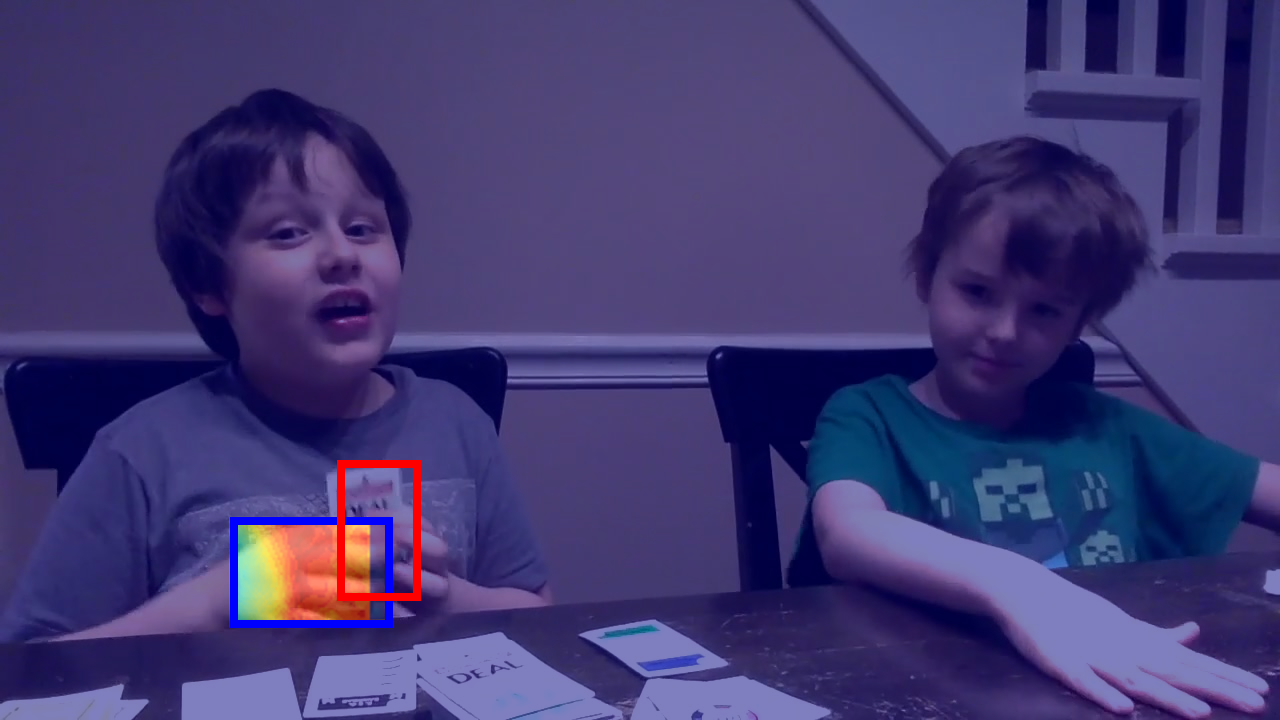}
	\includegraphics[width=0.19\linewidth]{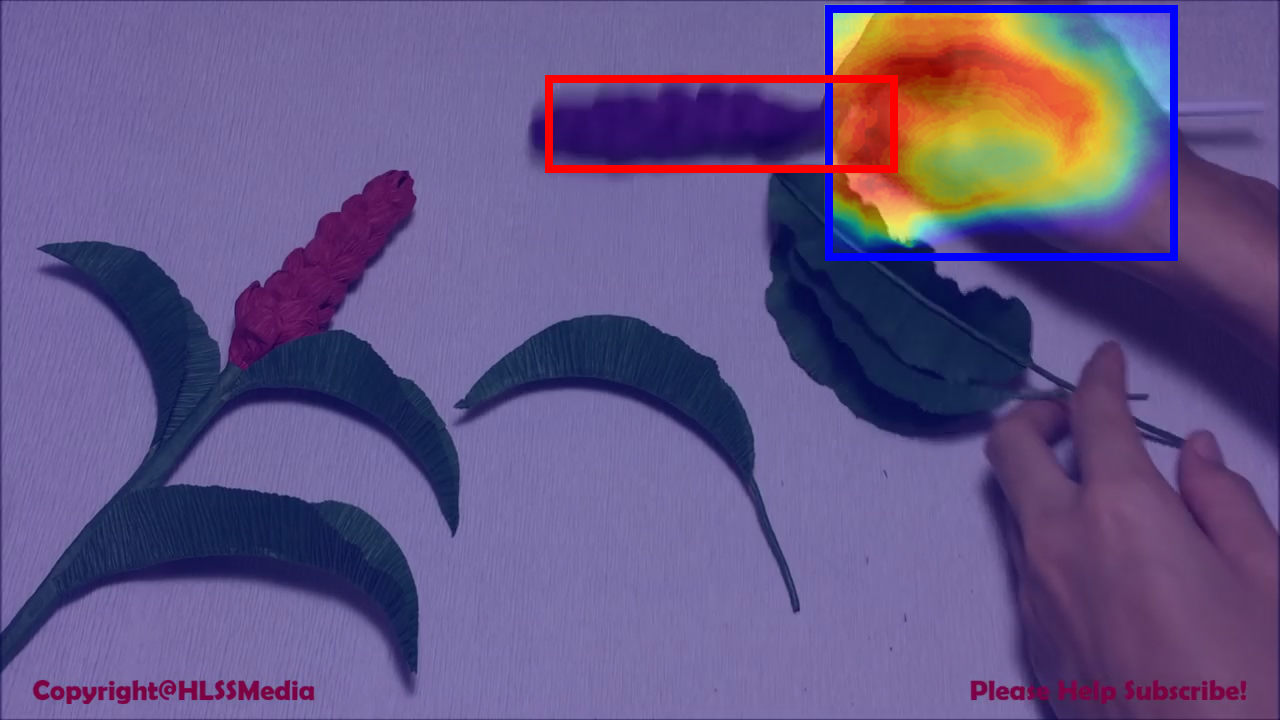}
	\includegraphics[width=0.19\linewidth]{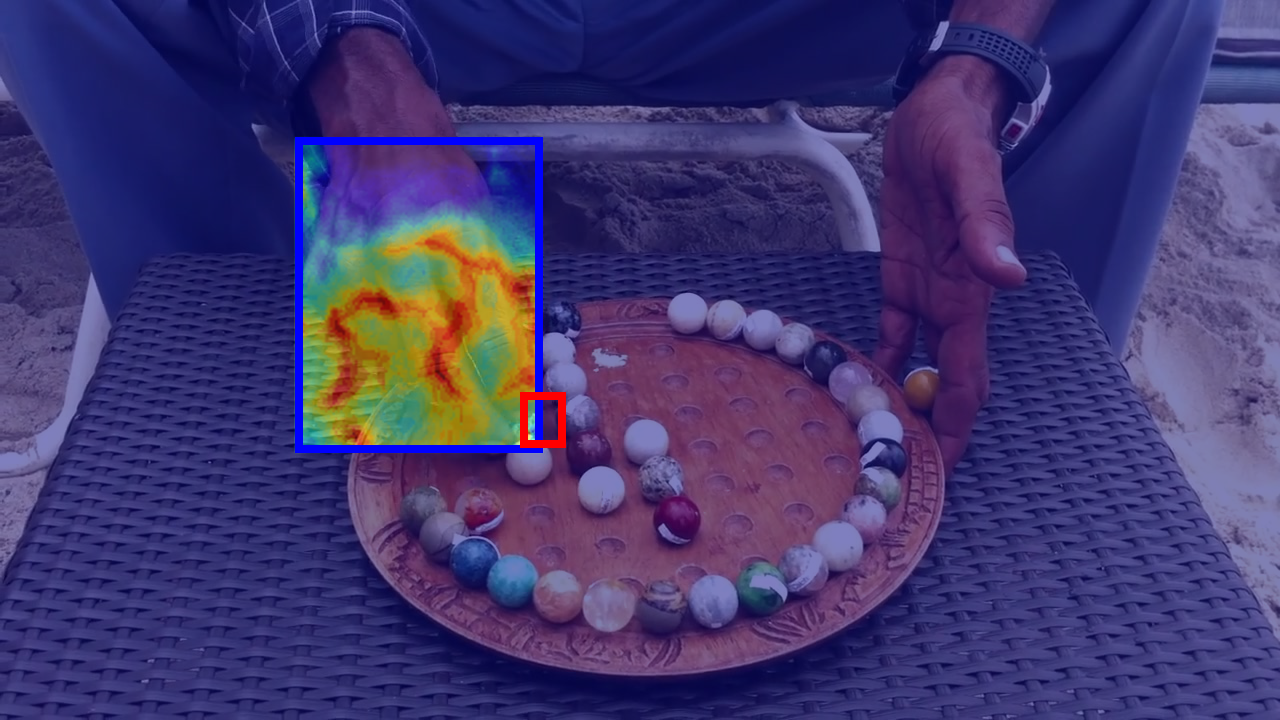}
	\\
	\vspace{10pt}
	
	\includegraphics[width=0.19\linewidth]{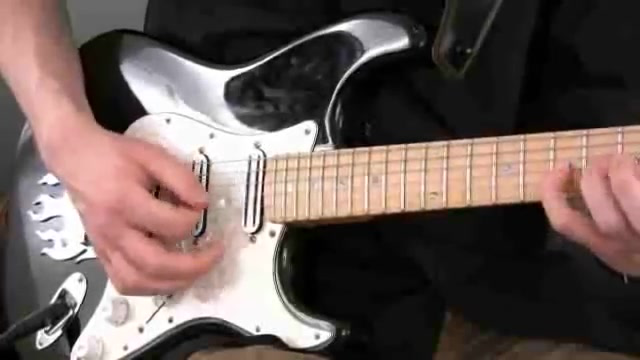}
	\includegraphics[width=0.19\linewidth]{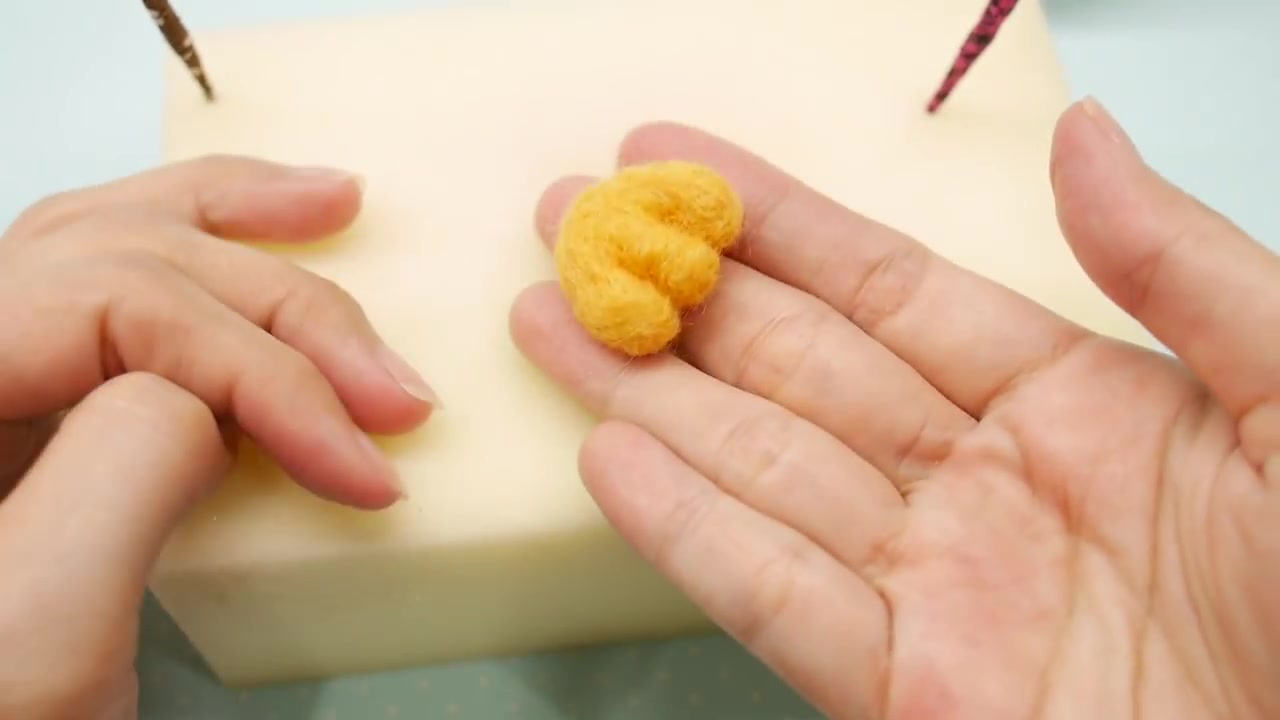}
	\includegraphics[width=0.19\linewidth]{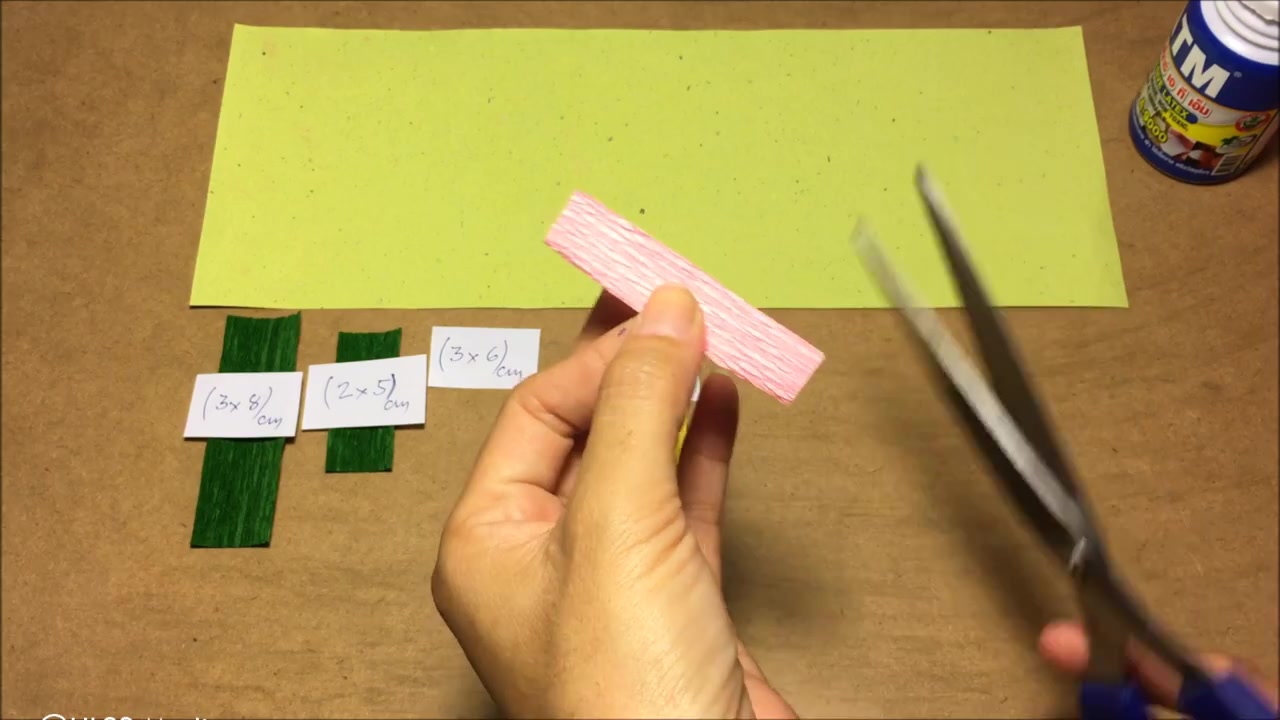}
	\includegraphics[width=0.19\linewidth]{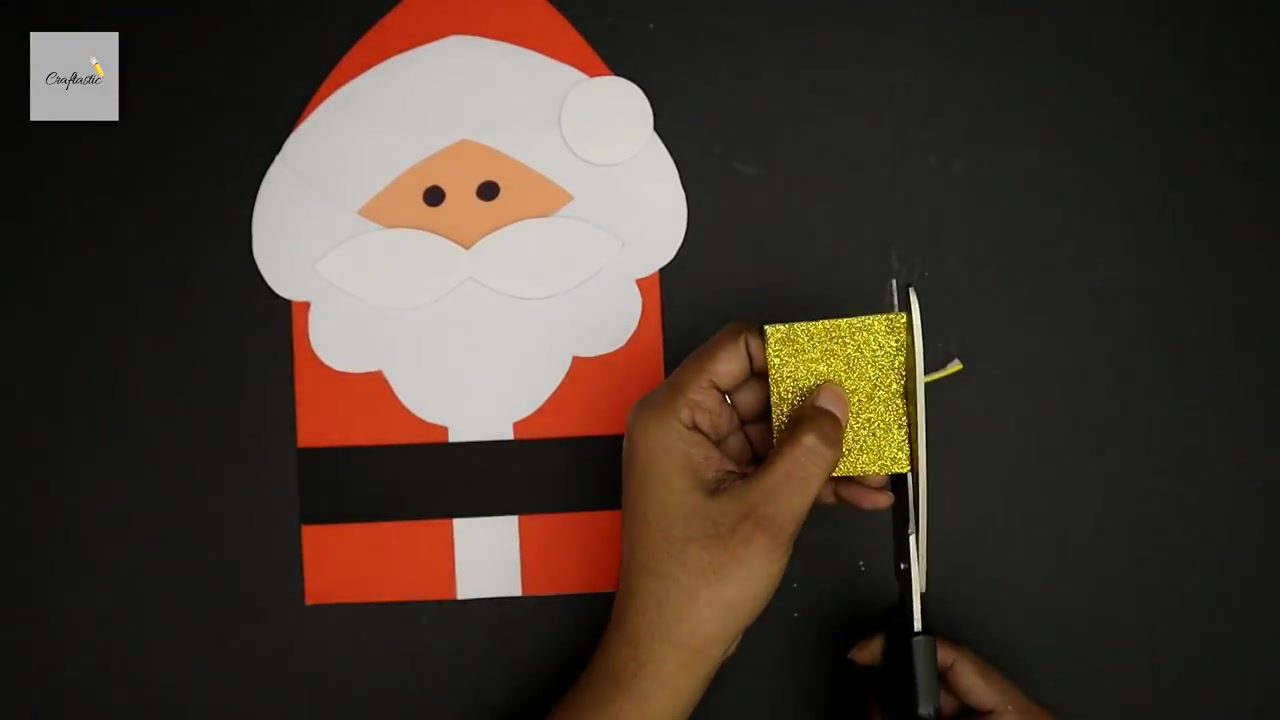}
	\includegraphics[width=0.19\linewidth]{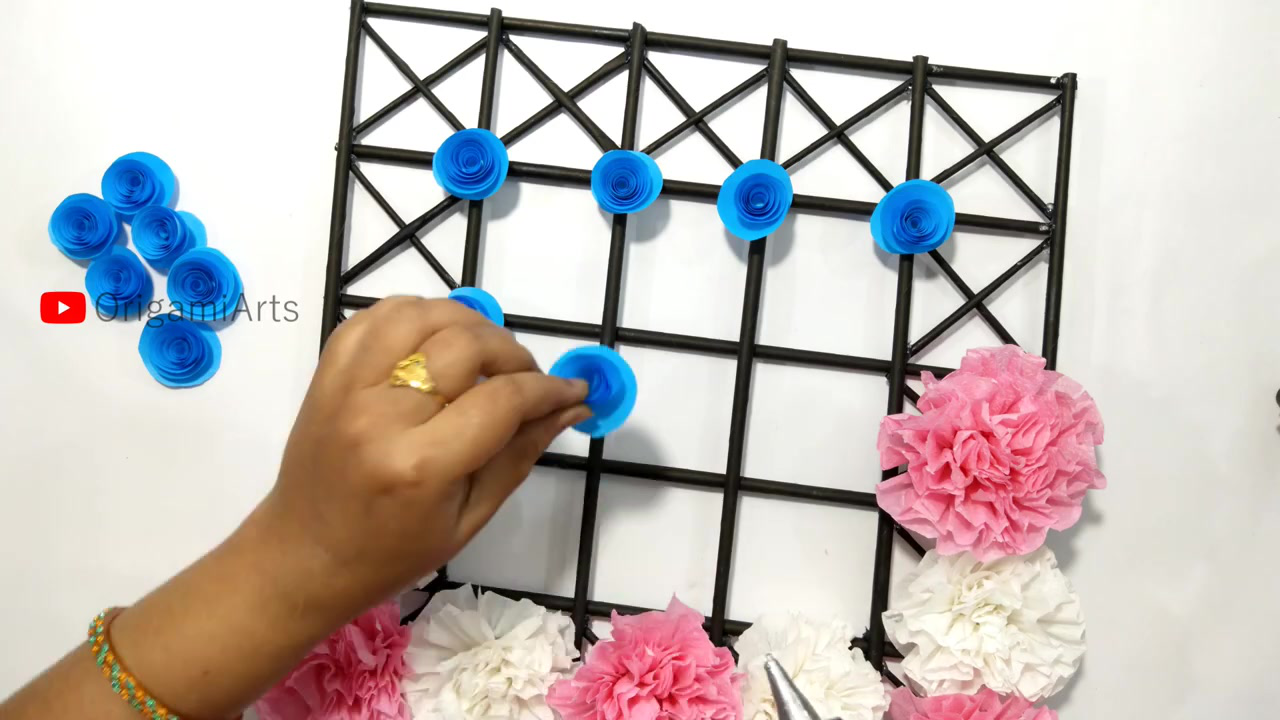}
	\\
	\includegraphics[width=0.19\linewidth]{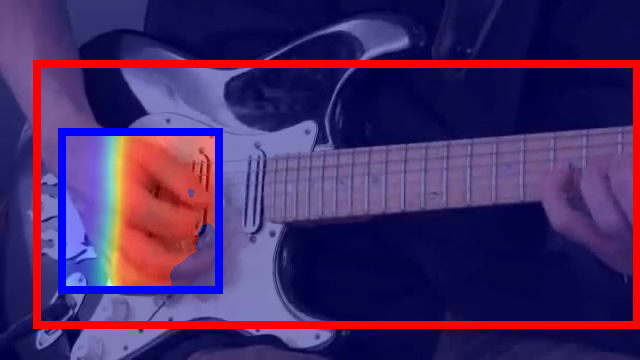}
	\includegraphics[width=0.19\linewidth]{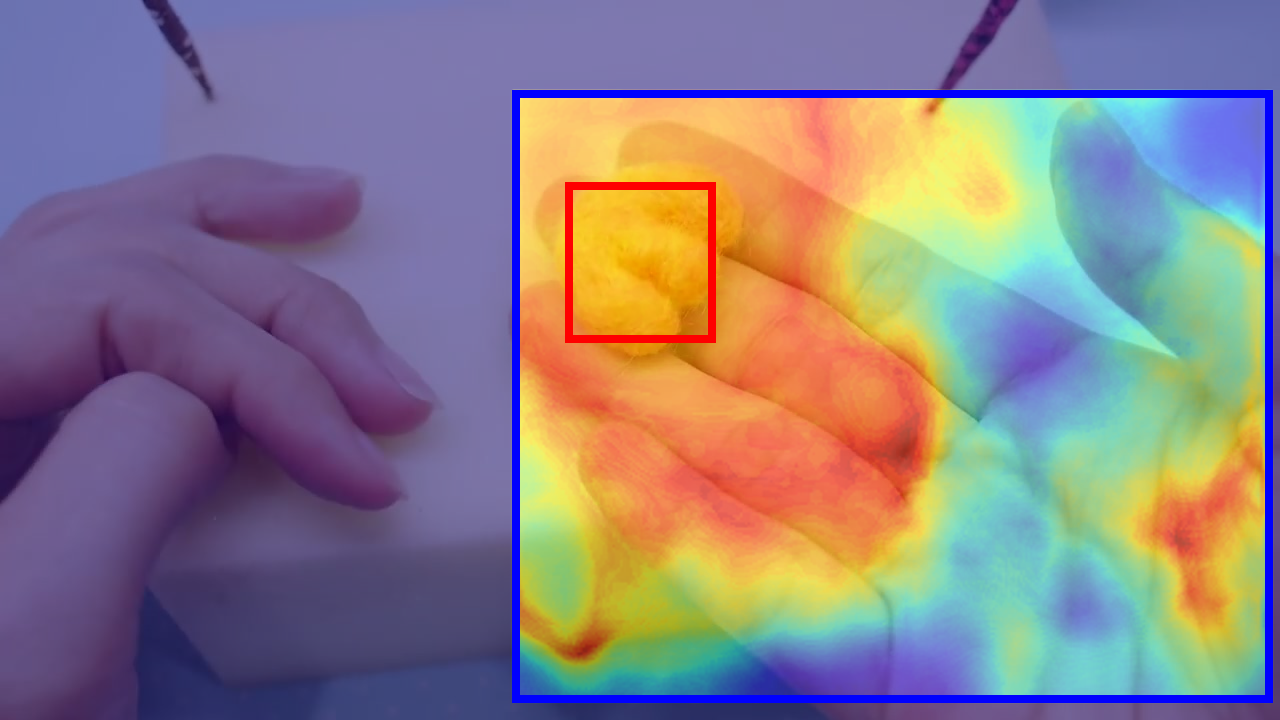}
	\includegraphics[width=0.19\linewidth]{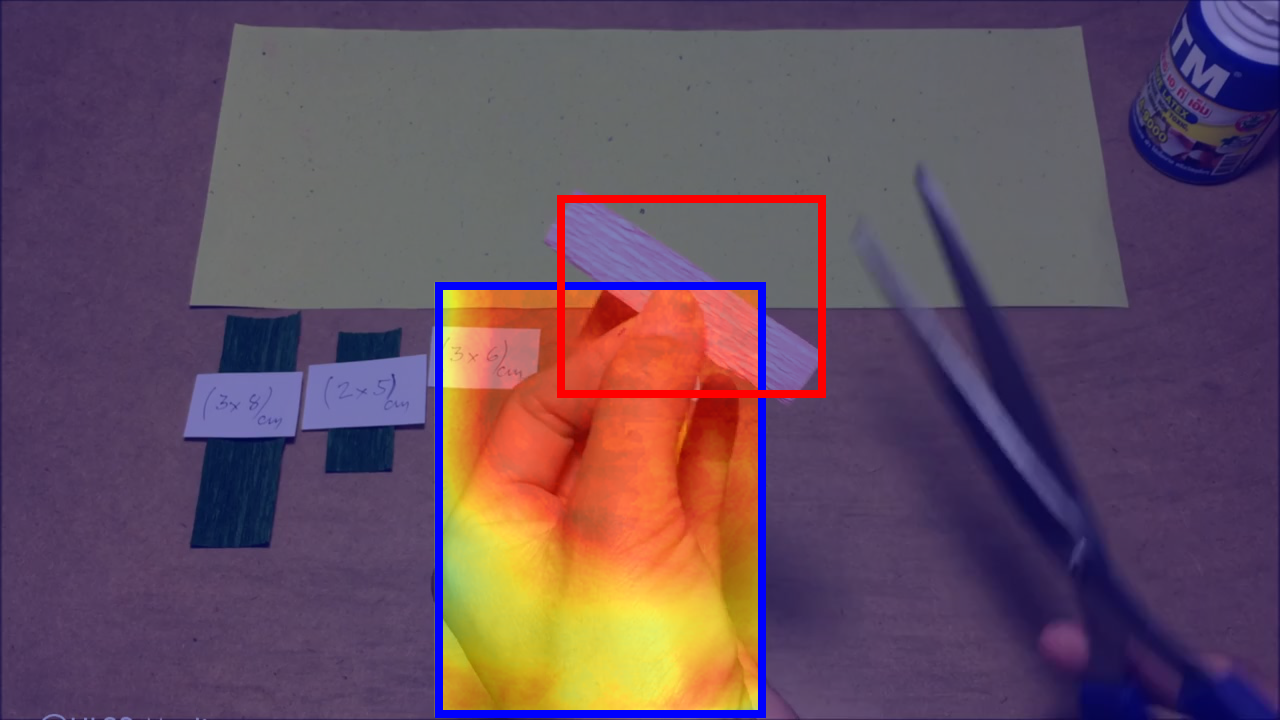}
	\includegraphics[width=0.19\linewidth]{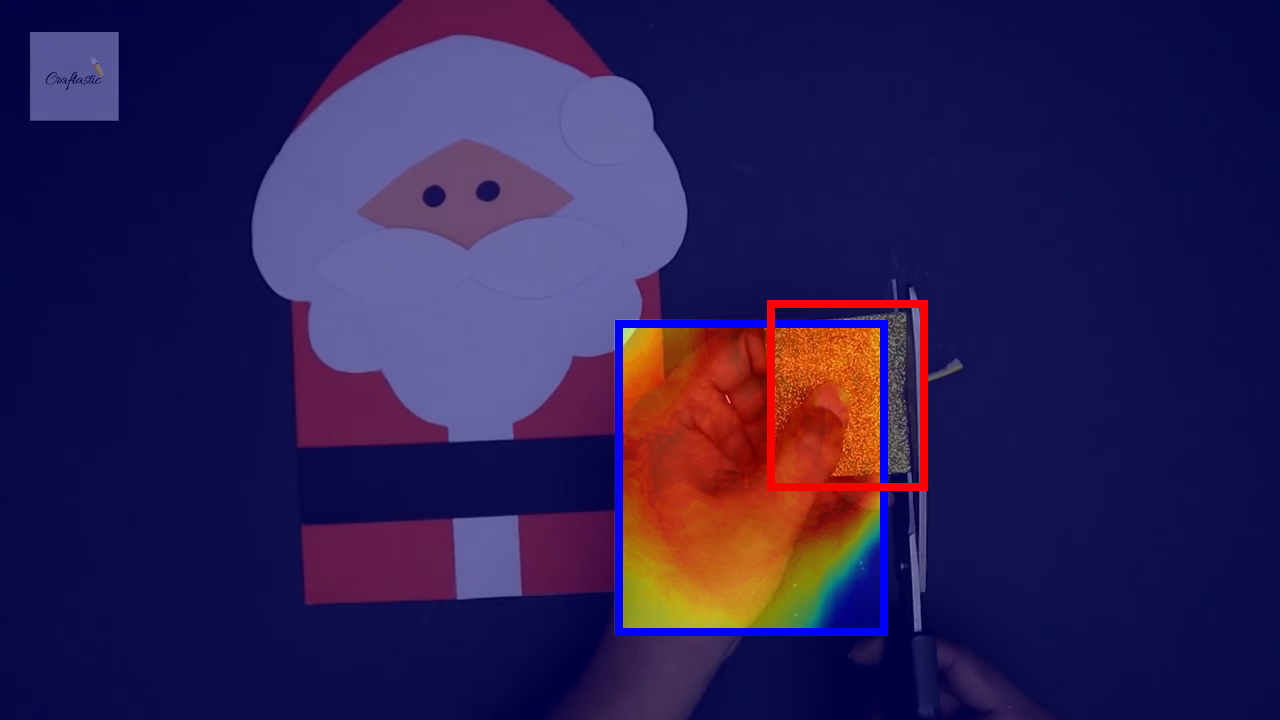}
	\includegraphics[width=0.19\linewidth]{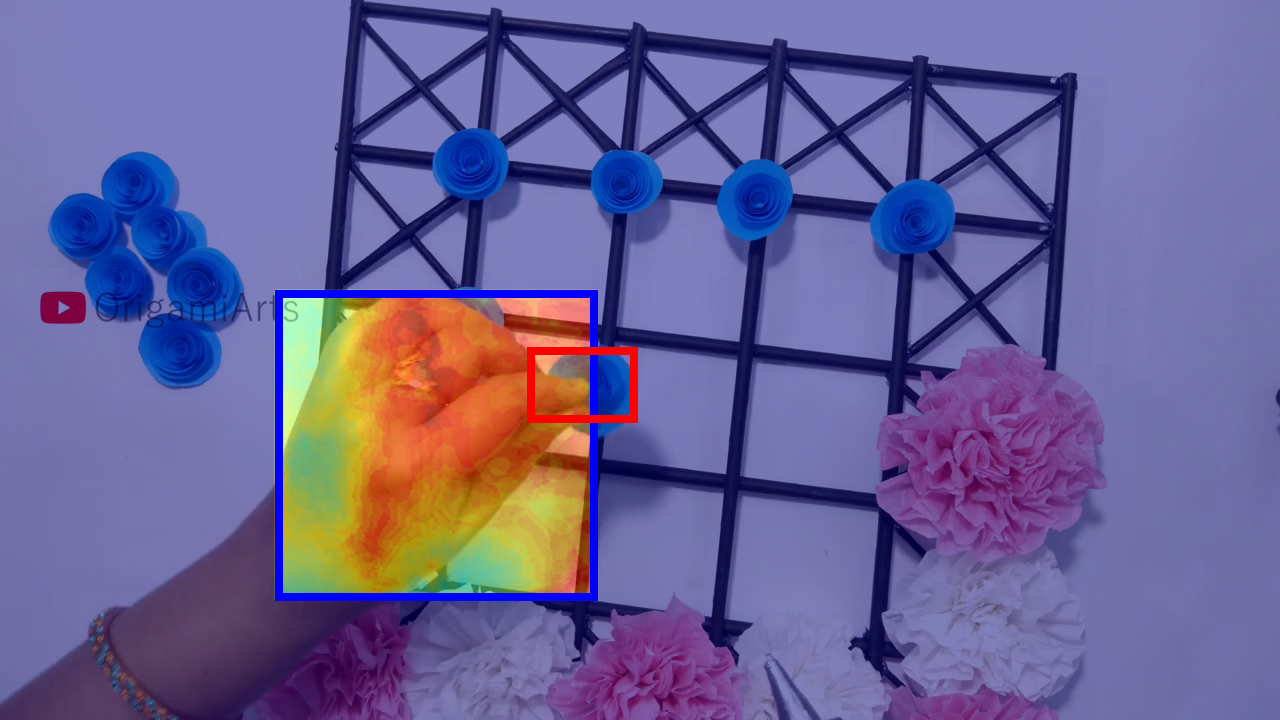}
	\\
	\vspace{10pt}
	
	\includegraphics[width=0.19\linewidth]{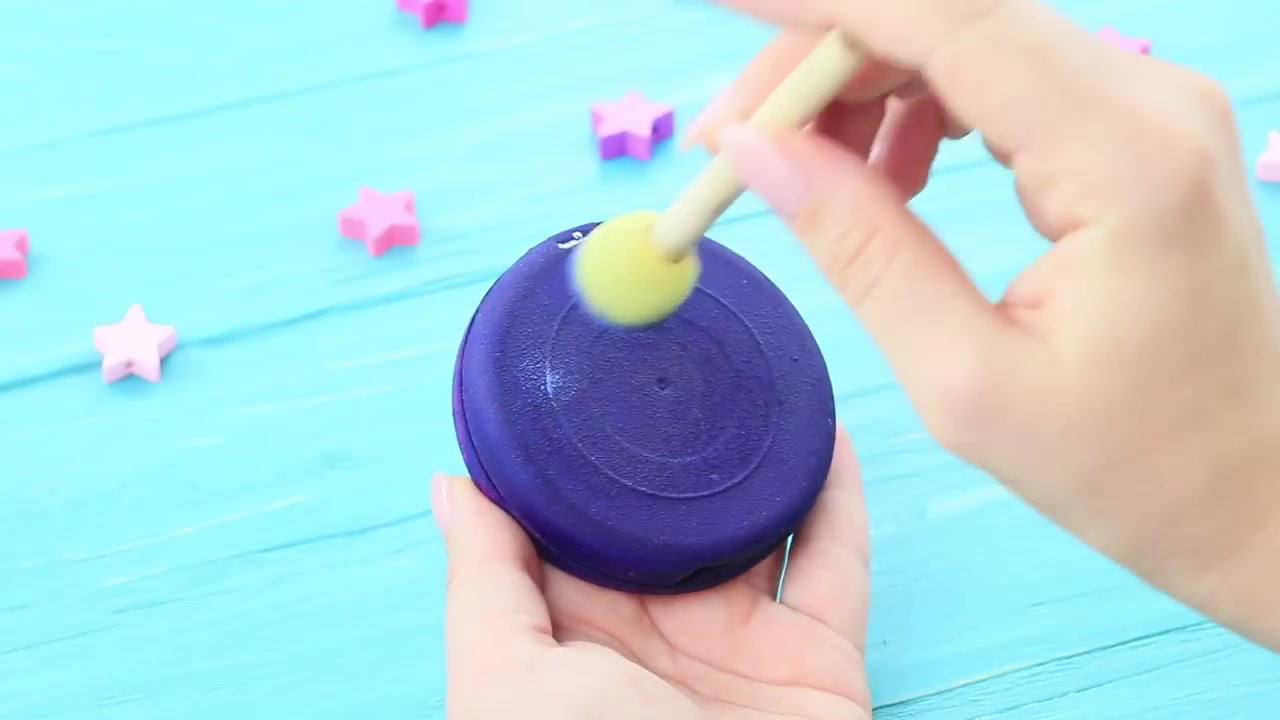}
	\includegraphics[width=0.19\linewidth]{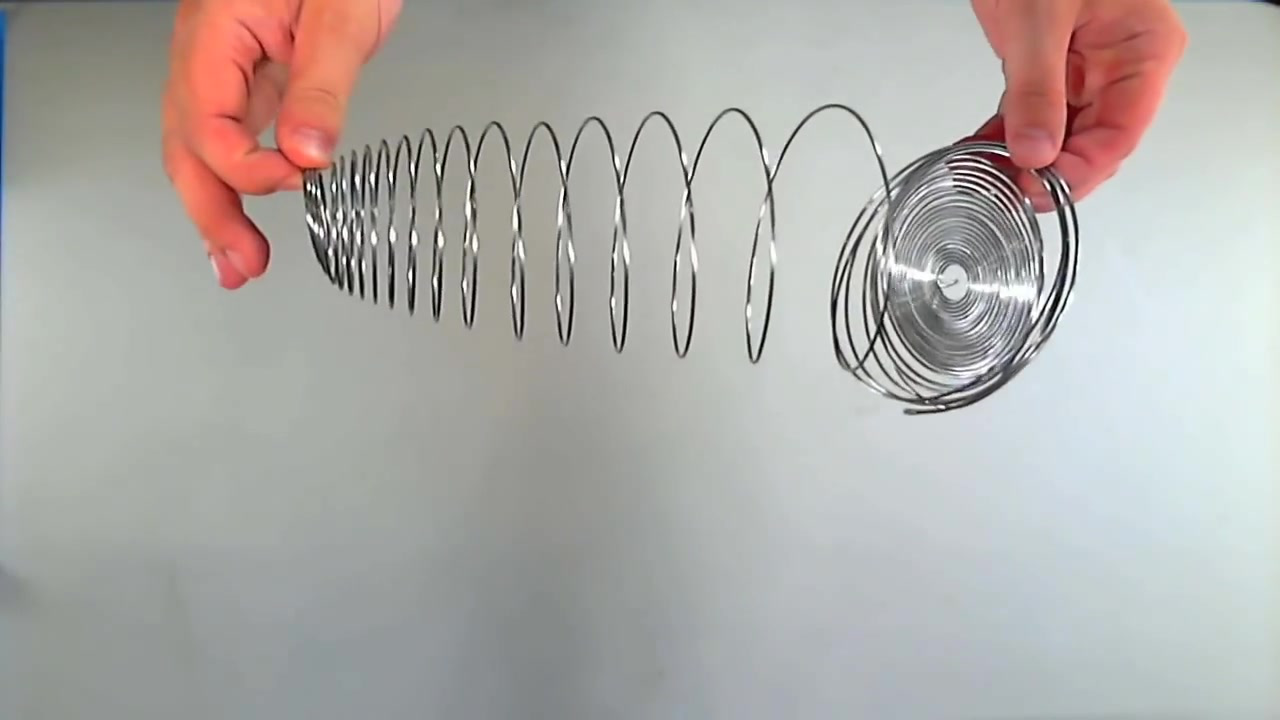}
	\includegraphics[width=0.19\linewidth]{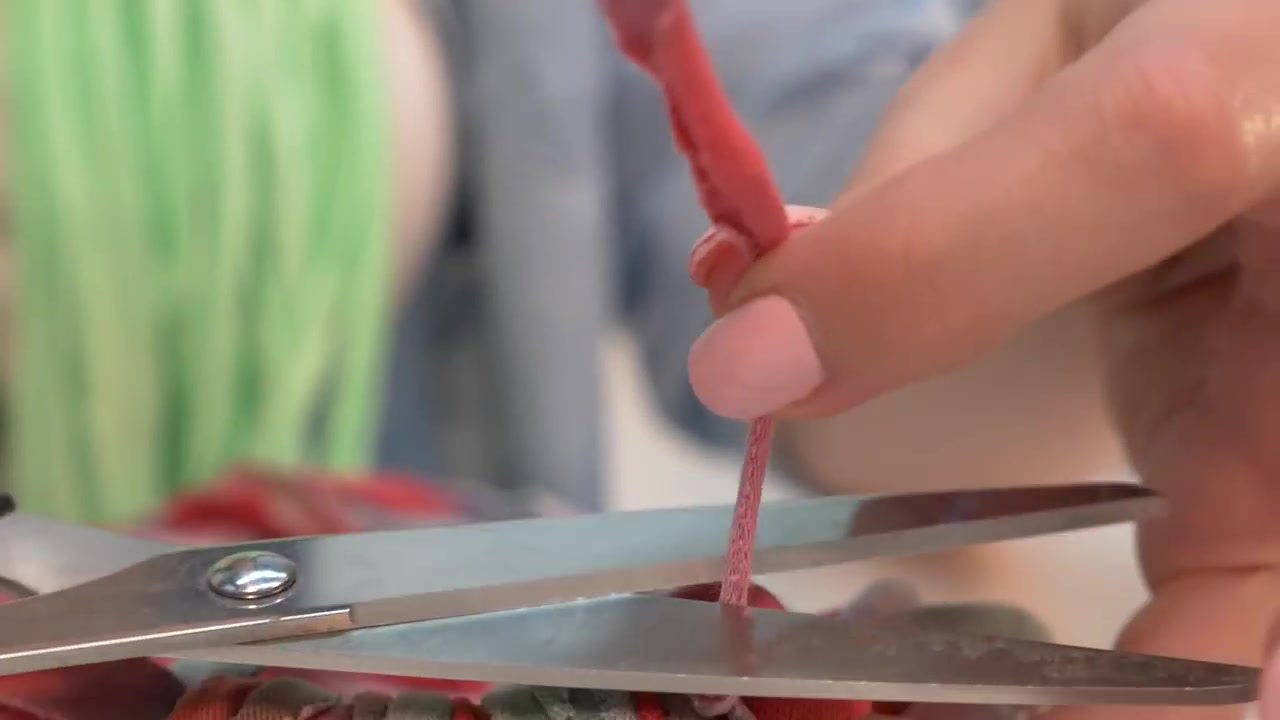}
	\includegraphics[width=0.19\linewidth]{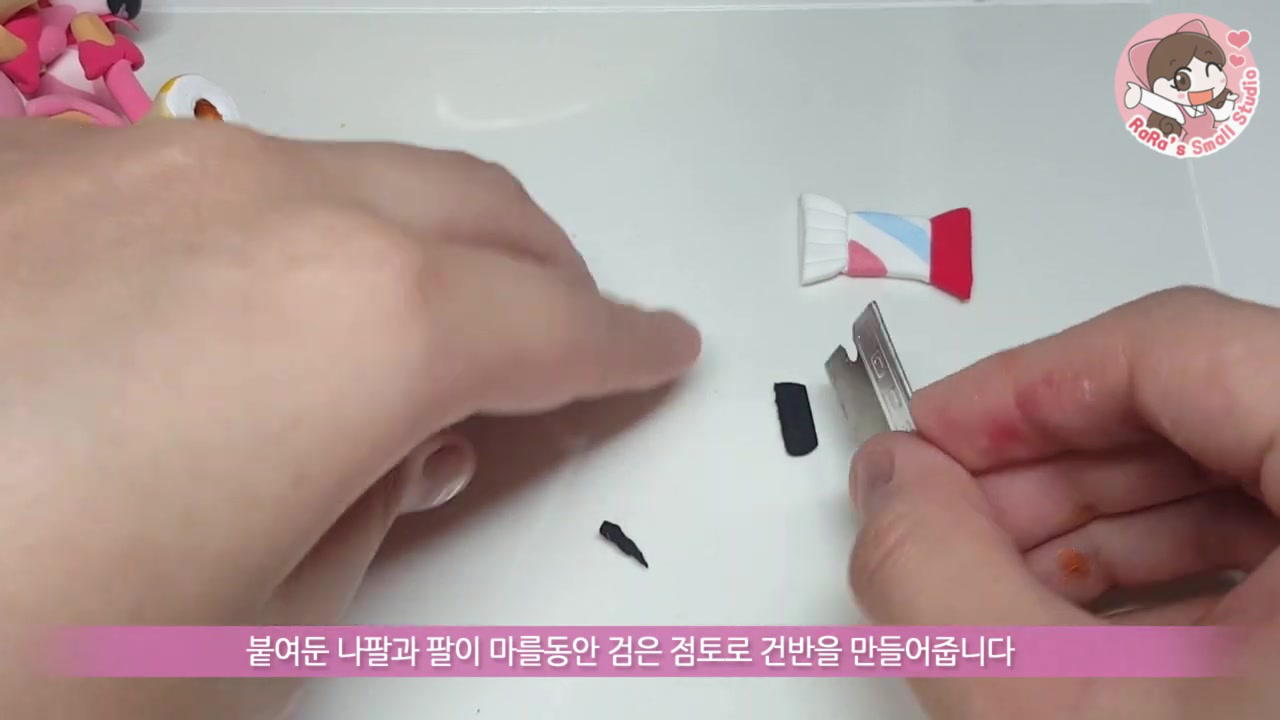}
	\includegraphics[width=0.19\linewidth]{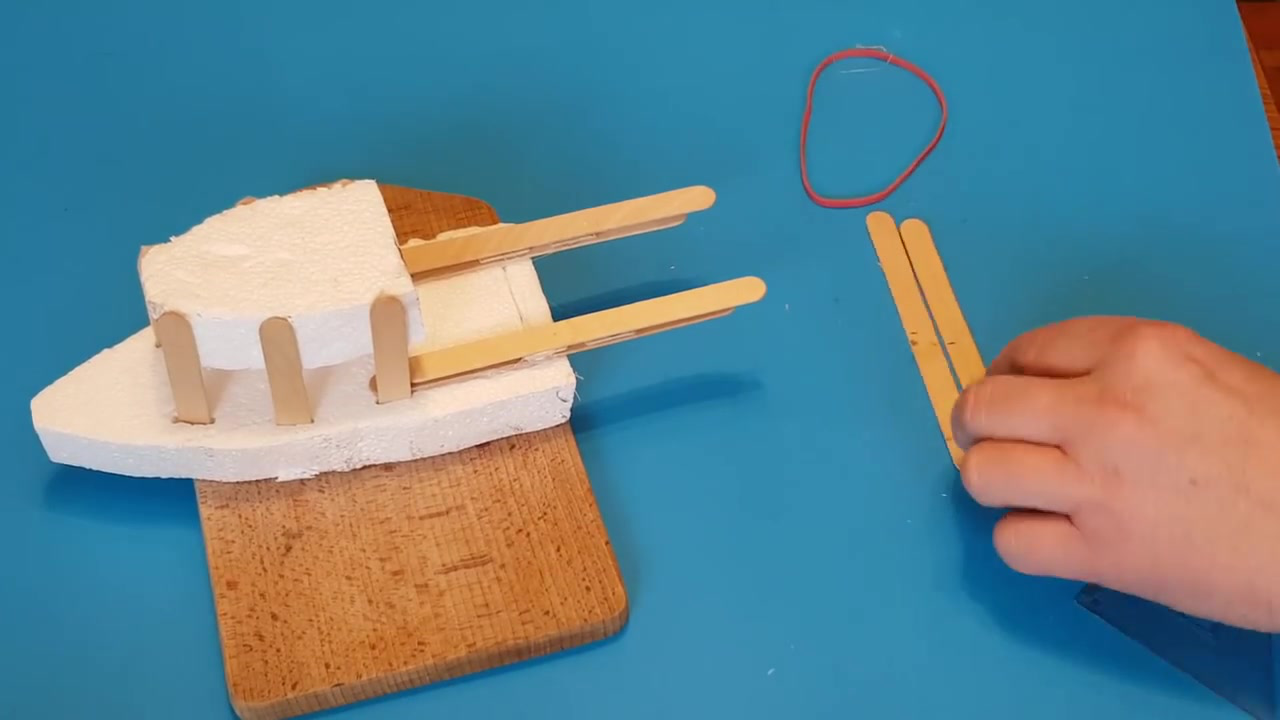}
	\\
	\includegraphics[width=0.19\linewidth]{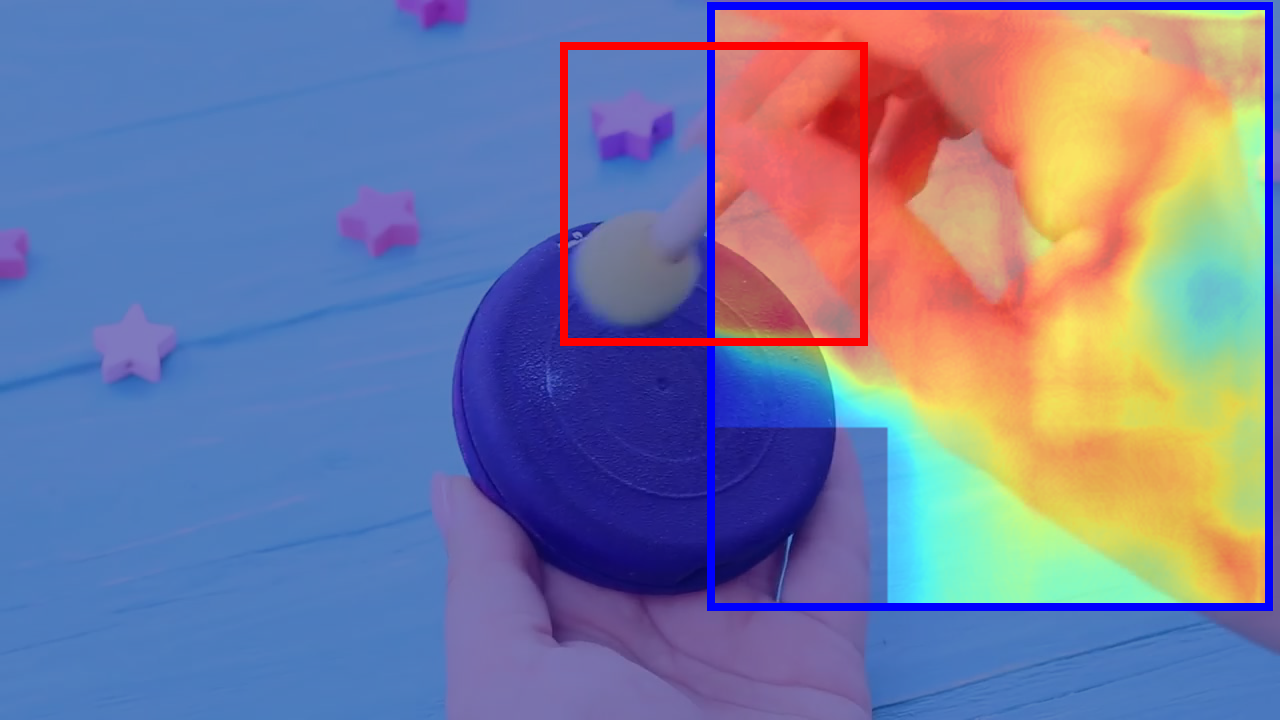}
	\includegraphics[width=0.19\linewidth]{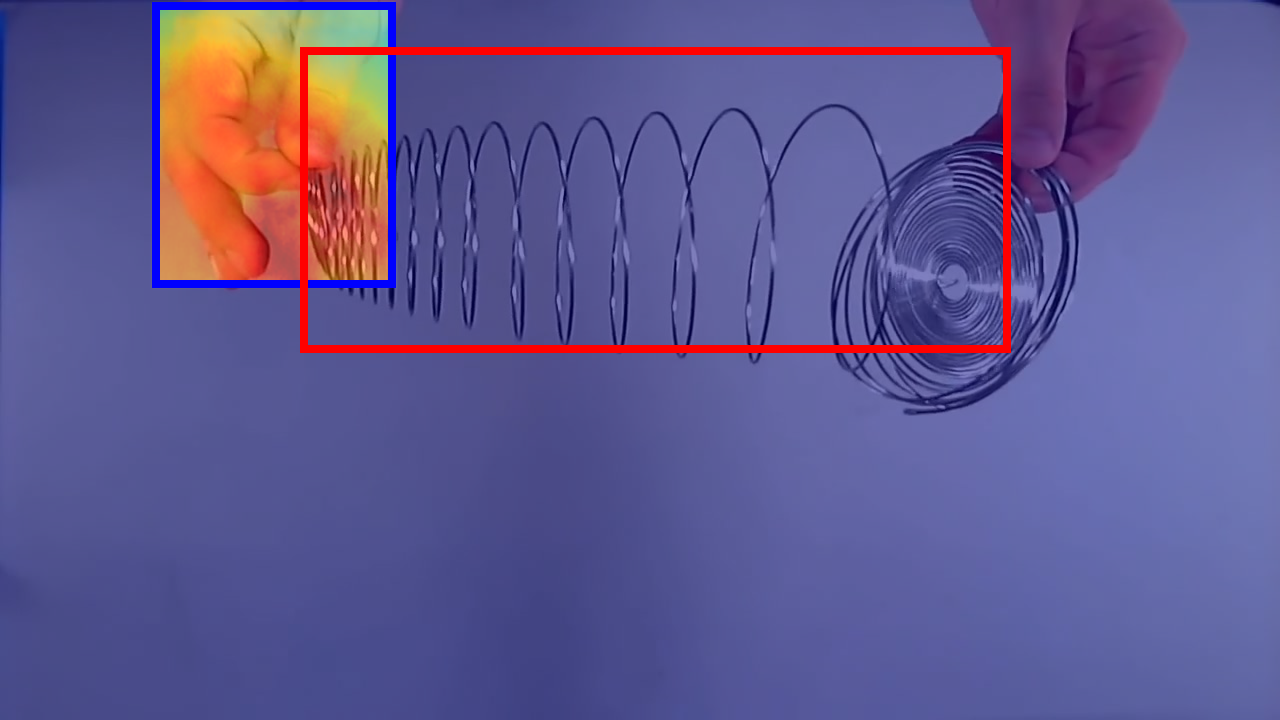}
	\includegraphics[width=0.19\linewidth]{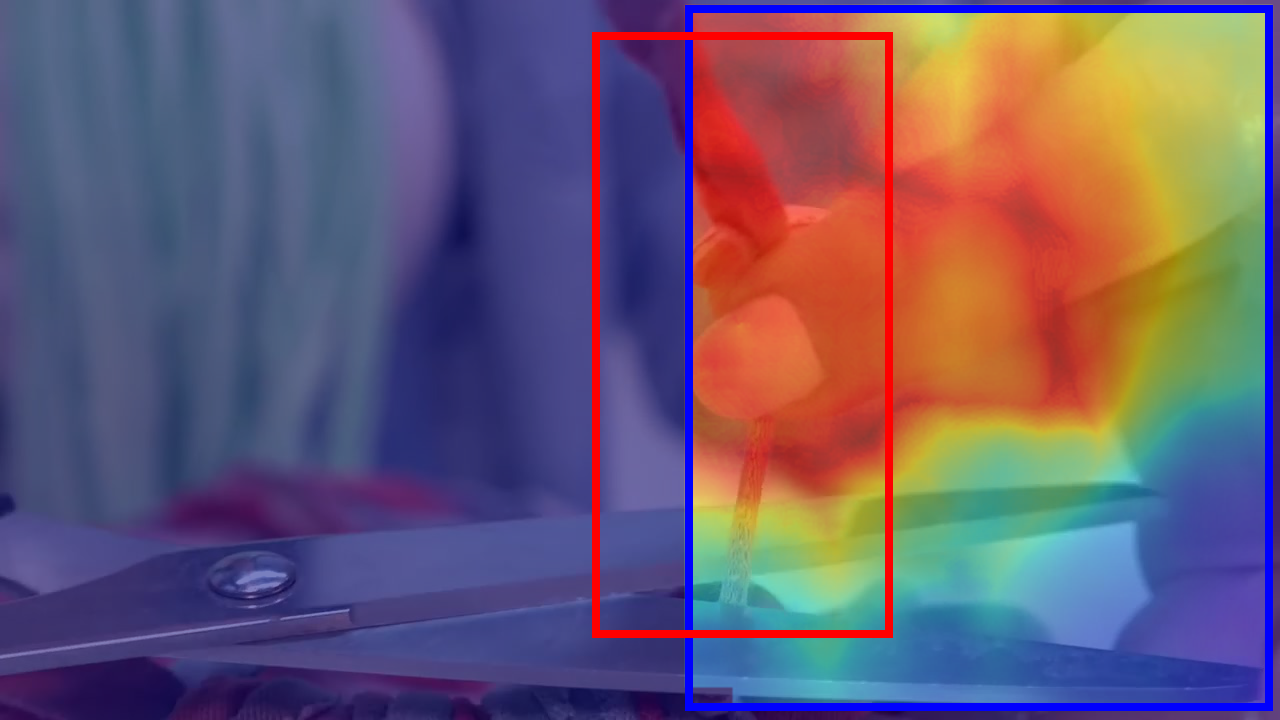}
	\includegraphics[width=0.19\linewidth]{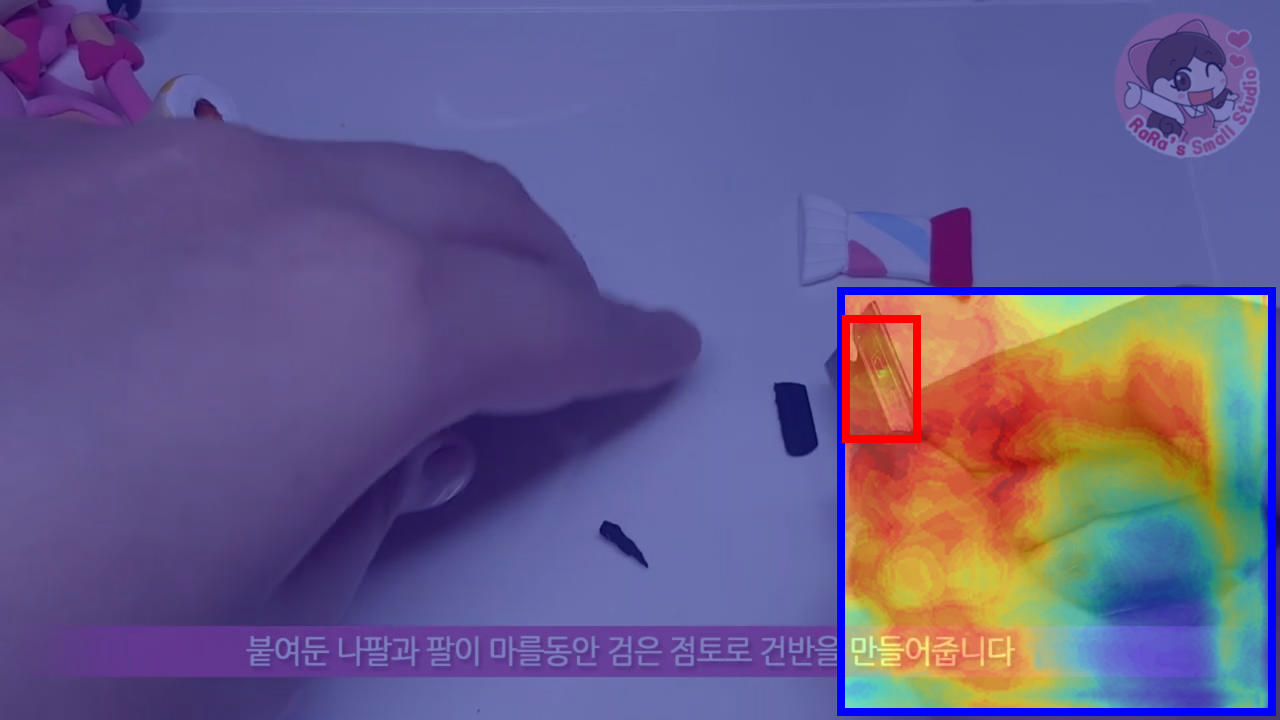}
	\includegraphics[width=0.19\linewidth]{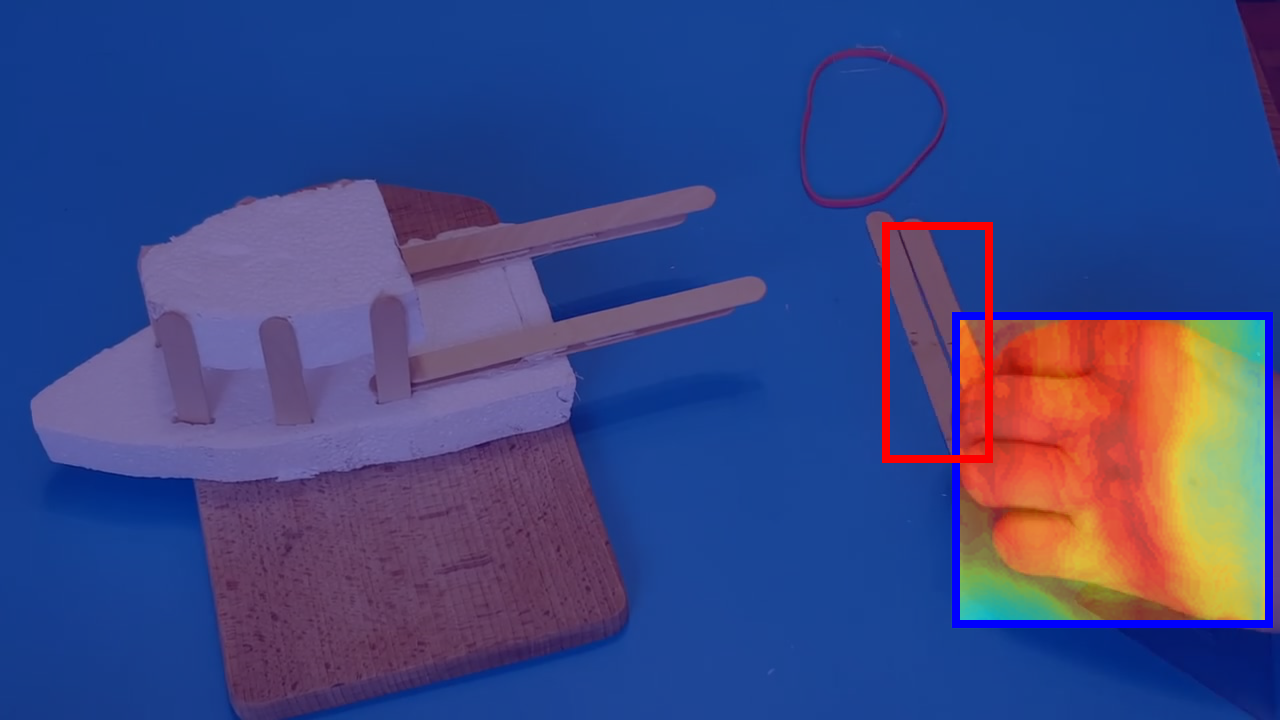}
	\\
	\centering
	\caption{\label{fig:vote_correctness_additional} We show more examples by visualizing the IoU (red indicates higher IoU) between the final active object box estimation (red) and the pixel-wise predictions inside the hand bounding box (blue). The final estimated bounding boxes picked by the voting are more closely related to the predictions in the regions of informative patterns such as fingers and objects as opposed to irrelevant information such as the background.}
\end{figure*}

\paragraph{Visualization of Pixel-wise Voting}
To validate the design of pixel-wise voting, we visualize more examples about the heatmap of the IoU between pixel-wise bounding box predictions and the final predicted bounding boxes after voting in \cref{fig:vote_correctness_additional}. In this visualization, we clearly observe that the final estimated bounding boxes picked by the voting are related more closely to the predictions in the regions of informative patterns such as fingers and objects as opposed to irrelevant information such as the background. For better visibility, every sample only shows one pair of hands and objects.

\section{Running Time}
\label{apdx:ssec:runtime}

We report the runtime on a desktop with a Ryzen 3900X CPU and an RTX 2080Ti GPU. For a $512 \times 512$ input image with $2$ hands on average, the proposed method runs at $18$ frames/second, with $13$ ms for network forward inference, and $42$ ms for active object localization with voting.

\end{document}